\documentclass[10pt,a4paper]{article}

\usepackage{graphicx}
\usepackage{amsmath}
\usepackage{verbatim}
\usepackage{amssymb}
\usepackage{linguex}
\usepackage[x11names,dvipsnames]{xcolor}
\usepackage{tikz}
\usetikzlibrary{arrows,automata}
\usepackage{booktabs}
\usepackage{multirow}
\usepackage{multicol}
\usepackage[mathscr]{euscript}
\usepackage[utf8x]{inputenc}
\usepackage[f]{esvect}
\usepackage{xspace}
\usepackage{soul}
\usepackage{centernot}
\usepackage[T1]{fontenc}
\usepackage{lineno}
\usepackage{comment}
\usepackage{lmodern}

\usepackage{hyperref}

\hypersetup{
  pdfstartview      = {FitH},
  pdfpagelayout     = {OneColumn},
  plainpages        = false,
  bookmarksnumbered = {true},
  naturalnames      = {true},
  colorlinks        = {true},
  citecolor         = ForestGreen
}

\usepackage[UKenglish]{babel}

\usepackage{natbib}

\setcitestyle{round,semicolon,authoyear,aysep={},yysep={,}}

\usepackage{enumitem}

\setlist[enumerate,1]{label=\textbf{\textit{(\roman*)}}, topsep=0.2em, itemsep=0.2em}
\setlist[enumerate,2]{label*=\textbf{\textit{(\arabic*)}}, topsep=0.2em, itemsep=0.2em}

\setlist[itemize,1]{label=\textbullet, topsep=0.35em, itemsep=0.2em}

\newlist{compactitemize}{itemize}{2}
\setlist[compactitemize,1]{label=\textbullet, leftmargin=1em, rightmargin=0em, topsep=0.25em, itemsep=0.25em}
\setlist[compactitemize,2]{label=--, leftmargin=1em, rightmargin=0em, topsep=0.25em, itemsep=0.25em}

\newlist{compactenumerate}{enumerate}{1}
\setlist[compactenumerate,1]{label=\textbf{\textit{(\roman*)}}, leftmargin=1.5em, rightmargin=0em, topsep=0.25em, itemsep=0.25em}

\newlist{inlineenum}{enumerate*}{1}
\setlist[inlineenum,1]{label=\textbf{\textit{(\roman*)}}}

\newlist{inductionproof}{description}{1}
\setlist[inductionproof,1]{topsep=0.4em, leftmargin=0em, itemsep=0.25em}

\usepackage[hyperref,thmmarks,framed]{ntheorem}

\theoremstyle{plain}
\theoremheaderfont{\normalfont\bfseries}
\theorembodyfont{\normalfont}
\theoremsymbol{\ensuremath{\blacktriangleleft}}
\newtheorem{exa}{Example}[section]
\newtheorem{definition}[exa]{Definition}

\theoremstyle{plain}
\theoremsymbol{\ensuremath{\blacksquare}}
\newtheorem{theorem}[exa]{Theorem}
\newtheorem{proposition}[exa]{Proposition}
\newtheorem{corollary}[exa]{Corollary}
\newtheorem{lemma}[exa]{Lemma}

\theoremheaderfont{\itshape}
\theorembodyfont{\normalfont}
\theoremstyle{nonumberplain}
\theoremsymbol{\ensuremath{\blacksquare}}
\newtheorem{proof}{Proof.}

\newcommand{\bs}[1]{\boldsymbol{#1}}
\newcommand{\set}[1]{\mathopen{\{} {#1} \mathclose{\}}}
\newcommand{\tuple}[1]{\langle #1 \rangle}
\newcommand{\card}[1]{\mathopen{\vert} #1 \mathclose{\vert}}
\newcommand{\intint}[2]{[#1 \mathop{..} #2]}
\newcommand{\prooflr}{\ensuremath{\bs{(\Rightarrow)}}\xspace}
\newcommand{\proofrl}{\ensuremath{\bs{(\Leftarrow)}}\xspace}
\newcommand{\qiffq}{\quad\text{if and only if}\quad}
\newcommand{\Nat}{\mathbb{N}}

\newcommand{\myparagraph}[1]{\medskip\noindent\textbf{#1}\,}
\newcommand{\myparagraphit}[1]{\medskip\noindent\textit{\textbf{#1}}\,}

\newenvironment{ctabular}[1]
{\begin{center}\begin{tabular}{#1}}
{\end{tabular}\end{center}}

\newenvironment{cbtabular}[1]
{\begin{center}\begin{tabular}[b]{#1}}
{\end{tabular} \end{center}}

\newenvironment{smallctabular}[1]
{\begin{center}\begin{small}\begin{tabular}[b]{#1}}
{\end{tabular}\end{small}\end{center}}

\newenvironment{smallltabular}[1]
{\begin{flushleft}\begin{small}\begin{tabular}{#1}}
{\end{tabular}\end{small}\end{flushleft}}

\renewcommand{\arraystretch}{1.3}

\newcommand{\keywords}[1]{\medskip\noindent\textbf{Keywords}: #1}

\newcommand{\Sig}{\bs{\mathscr{S}}}
\newcommand{\NV}{\bs{\mathcal{V}}}
\newcommand{\XV}{\bs{\mathcal{U}}}
\newcommand{\AV}{\bs{\mathcal{W}}}

\newcommand{\Ran}{\bs{\mathcal{R}}}
\renewcommand{\O}{\bs{\mathcal{O}}}
\newcommand{\allminus}[1]{\AV_{-#1}}
\newcommand{\F}{\mathcal{F}}
\newcommand{\fun}{F}
\newcommand{\A}{\mathcal{A}}
\newcommand{\B}{\mathcal{B}}
\newcommand{\C}{\mathcal{C}}
\newcommand{\R}{\mathcal{R}}

\renewcommand{\vec}[1]{\vv{#1}}
\newcommand{\ina}{{=}}
\newcommand{\opint}[2]{#1\ina#2}
\newcommand{\newf}[2]{{{#1}_{#2}}}
\newcommand{\newval}[3]{{{#1}^{#2}_{#3}}}
\renewcommand{\int}[1]{[#1]}
\newcommand{\syndirparof}{\rightsquigarrow}
\newcommand{\T}{\mathcal{T}}
\newcommand{\ev}{\operatorname{E}} 
\newcommand{\ec}{\ensuremath{\mathcal{E}}\xspace} 
\newcommand{\LAc}{\ensuremath{\mathcal{L}_{\mathit{C}}}\xspace}
\newcommand{\LAcres}{\ensuremath{\mathcal{L'}_{\mathit{C}}}\xspace}
\newcommand{\LAkc}{\ensuremath{\mathcal{L}_{\mathit{KC}}}\xspace}
\newcommand{\LAkcres}{\ensuremath{\mathcal{L'}_{\mathit{KC}}}\xspace}
\newcommand{\LAfull}{{\ensuremath{\mathcal{L}_{\mathit{PAKC}}}}\xspace}
\newcommand{\LAfulla}{{\ensuremath{\mathcal{L}_{\mathit{PAKC}}^{@}}}\xspace}
\newcommand{\LAfullres}{{\ensuremath{\mathcal{L'}_{\mathit{PAKC}}}}\xspace}

\newcommand{\ax}[1]{\ensuremath{\operatorname{\bs{\operatorname{#1}}}}}
\newcommand{\LOc}{\ensuremath{\mathsf{L}_{\mathit{C}}}\xspace}
\newcommand{\CO}{\mathcal{CO}}
\newcommand{\vdashloc}{\mathrel{\vdash_{\LOc}}}
\newcommand{\LOcres}{\ensuremath{\mathsf{L'}_{\mathit{C}}}\xspace}
\newcommand{\vdashlocres}{\mathrel{\vdash_{\LOcres}}}
\newcommand{\LOkc}{\ensuremath{\mathsf{L}_{\mathit{KC}}}\xspace}
\newcommand{\vdashlokc}{\mathrel{\vdash_{\LOkc}}}
\newcommand{\LOkcres}{\ensuremath{\mathsf{L'}_{\mathit{KC}}}\xspace}
\newcommand{\vdashlokcres}{\mathrel{\vdash_{\LOkcres}}}
\newcommand{\LOfull}{\ensuremath{\mathsf{L}_{\mathit{PAKC}}}\xspace}
\newcommand{\vdashlofull}{\mathrel{\vdash_{\LOfull}}}
\newcommand{\LOfullo}{\ensuremath{\mathsf{L}_{\mathit{PAKC}}^{\mathit{O}}}\xspace}
\newcommand{\LOkco}{\ensuremath{\mathsf{L}^O_{\mathit{KC}}}\xspace}

\newcommand{\sub}{\operatorname{sub}}
\newcommand{\com}{\operatorname{c}}

\newcommand{\tr}{\operatorname{tr}}
\newcommand{\tro}{\operatorname{tr_1}}
\newcommand{\trt}{\operatorname{tr_2}}
\newcommand{\trr}{\operatorname{tr_3}}
\newcommand{\trf}{\operatorname{tr_4}}
\newcommand{\mcss}{\mathbb{C}_{\mathit{C}}}
\newcommand{\mcssk}{\mathbb{C}_{\mathit{KC}}}
\newcommand{\mcssko}{\mathbb{C}^{O}_{\mathit{KC}}}
\newcommand{\rmcss}{\mathbb{D}}

\newcommand{\dirparof}[1]{\hookrightarrow_{#1}}
\newcommand{\parof}[1]{\hookrightarrow^{+}_{#1}}

\newcommand{\colornew}{Bittersweet}

\definecolor{bordeaux}{RGB}{192,0,64}

\newcommand{\rewritef}[1]{\textcolor{Blue}{\bf #1}}

\newcommand{\colorfer}{BrickRed}

\newcommand{\newKX}[1]{\textcolor{orange}{#1}}
\newcommand{\changeFV}[2]{\textcolor{lightgray}{\st{\protect#1}}\textcolor{\colorfer}{\protect#2}}
\newenvironment{newtextFV}
{\color{\colorfer}}
{\normalcolor}

\usepackage{authblk}

\includecomment{appendix:full}
\excludecomment{appendix:short}

\begin{document}


\title{Observing Interventions:\\ A logic for thinking about experiments\thanks{The authors are listed in alphabetical order.}\;\;\thanks{This is the extended version of a paper that will appear in a special issue of the Journal of Logic and Computation dedicated to the 3rd DaL{\'\i} Workshop on \emph{Dynamic Logic: New Trends and Applications}. Different from the journal version, here the reader can find the full technical appendix.}}

\author[1]{Fausto Barbero}
\author[2]{Katrin Schulz}
\author[3]{Fernando R. Vel{\'a}zquez-Quesada}
\author[4]{Kaibo Xie\thanks{Corresponding author}}

\affil[1]{University of Helsinki. \texttt{fausto.barbero@helsinki.fi}}
\affil[2]{ILLC, Universiteit van Amsterdam. \texttt{K.Schulz@uva.nl}}
\affil[3]{Department of Information Science and Media Studies, Universitetet i Bergen. \texttt{Fernando.VelazquezQuesada@uib.no}}
\affil[4]{Tsinghua University. \texttt{xkb@mail.tsinghua.edu.cn}}

\date{}

\maketitle

\begin{abstract}
  This paper makes a first step towards a logic of learning from experiments. For this, we investigate formal frameworks for modeling the interaction of causal and (qualitative) epistemic reasoning. Crucial for our approach is the idea that the notion of an {\it intervention} can be used as a formal expression of a (real or hypothetical) experiment \citep{Pearl09, Woo2003}. In a first step we extend a causal model \citep{Galles,halpern2000axiomatizing,Pearl09,briggs2012interventionist} with a simple Hintikka-style representation of the epistemic state of an agent. In the resulting setting, one can talk about the knowledge of an agent and information update. The resulting logic can model reasoning about thought experiments. However, it is unable to account for learning from experiments, which is clearly brought out by the fact that it validates the principle of {\it no learning} for interventions. Therefore, in a second step we implement a more complex notion of knowledge \citep{Nozick} that allows an agent to observe (measure) certain variables when an experiment is carried out. This extended system does allow for learning from experiments. For all the proposed logics, we provide a sound and complete axiomatization. 


\end{abstract}

\section{Introduction}\label{introduction}

In recent years a lot of effort has been put in the development of formal models of causal reasoning. A central motivation for this is the importance of causal reasoning for AI. Making computers take into account causal information is currently one of the central challenges of AI research \citep{PearlWhy,Bergstein}. There has also been tremendous progress in this direction after the earlier groundbreaking work in \citet{Pearl09} and \citet{Spirtes}. Advanced formal and computational tools have been developed for modelling causal reasoning and learning causal information, with applications 
in many different scientific areas. 

However, there is one aspect of causal reasoning that has not gotten enough attention yet: the interaction between causal and epistemic reasoning. Even though the standard logical approach to causal reasoning \citep{Pearl09,halpern2000axiomatizing,halpern2016actual} can model epistemic uncertainty\footnote{For this, add a probability distribution over the exogenous variables of the model. Uncertainty is then restricted to the value of variables. All causal dependencies are deterministic.}, it does not come with an object language that can make statements about the epistemic state of some agent. There are recent proposals adding probabilistic expressions to the object language (e.g., \citealt{ibeling2020probabilistic}), but very little has been done on combining causal and qualitative epistemic reasoning.\footnote{See \citet{barbero2019interventionist} for an exception, although the epistemic element is not made fully explicit in the language considered in that paper.} However, this kind of reasoning occurs frequently in our daily life. Consider, for instance, the following situation.

\begin{exa}\label{flashlight}
  Sarah got a new flashlight for her birthday. The flashlight emits light only if the button is pushed and the batteries are charged. At the moment the button is not pressed and the light is off. Sarah hasn't tried the flashlight yet. So, she doesn't know whether the battery is full or not. Let us assume that, in fact, the battery is empty.
\end{exa}

\noindent In such cases, we want to be able to infer that Sarah is not sure whether, had the button been pushed, the flashlight would have emitted light. A logic accounting for such inferences needs a language with statements involving epistemic attitudes towards causal inferences. This will be the topic of this paper.

We are interested here in one particular situation where the interaction between causal and epistemic is essential: experimentation. Our goal is to develop a logical framework that allows us to model and reason about experiments. Experiments play a central role for our survival, because they facilitate learning. We do not only learn by observing; we actively and purposefully interact with the world in order to understand our surroundings better. For instance, in the example given above, Sarah could experiment with the button of the flashlight (press it) in order to find out whether the battery is charged. Studying the laws and conditions that govern learning from experiments will help us to see the possibilities, but also the limits of the experimental method. In this paper we set out to make a first step towards a logical model for this type of reasoning.

The state of the art on causal and epistemic reasoning provides us with all the necessary means to embark on this project. On the one hand, we have well-developed systems of causal reasoning that center on the concept of intervention  -- which is meant to capture the notion of a (hypothetical) experiment. On the other hand, we have a rich literature on epistemic and dynamic epistemic logics designed explicitly to model an agent's knowledge and the way it changes. The only step missing is an integration of these two logical systems. This is what we set out to do here. More concretely, we will combine the standard approach to causal reasoning \citep{Pearl09,halpern2000axiomatizing,halpern2016actual} with tools from Epistemic Logic (\textit{EL}; \citealp{Hintikka1962,RAK1995})\footnote{Other options might be more powerful/expressive (e.g., probabilistic tools, as in \citealp{FaginH94}, and their dynamics extensions, as in \citealp{Kooi03,baltag2008probabilistic,MartinyM15}); yet, \textit{EL} is enough for our purposes.} and Dynamic Epistemic Logic (\textit{DEL}; \citealp{BaltagMossSolecki1998,vanBenthem2011ldii,vanDitmarschEtAl2007}).\footnote{Different from other options (e.g., epistemic/doxastic temporal logic; \citealp{ParikhR03}), \textit{DEL} allows us to represent different epistemic actions (interventions, observations), and also to study the way they interact with one another (see, e.g., the axioms in Table~\ref{tbl:rcm-LAka}).} 

We will start with building a very simple extension of the standard system of causal reasoning to a semantic framework that can interpret modal statements about the knowledge of an agent. 
This system will be extended with the dynamic epistemic operator of public announcement. In the context of experimentation, a public announcement can be naturally interpreted as an \emph{observation}.\footnote{Cf. the \emph{selective implication} of \citet{barbero2019interventionist,BarSan2020}.} The resulting logic will be able to model reasoning about hypothetical experiments (thought experiments). In particular, it will account for the first inference described in the context of Example~\ref{flashlight}: Sarah is not sure whether, had the button been pushed, the flashlight would have emitted light.

However, this system cannot model learning from interventions. This is reflected in the fact that  it validates the so called rule of {\it No-Learning}: if the agent knows $\phi$ after an intervention, then before the intervention she knew that intervening would make $\phi$ true; in symbols,
\begin{equation*}\label{nolearning}
   \ax{NL}: \ \ \
   [\vec{X}{=}\vec{x}]K\phi\rightarrow K[\vec{X}{=}\vec{x}]\phi.
\end{equation*}
For overcoming this limitation, we will introduce the concept of observables. Observables describe the variables the agent can observe (the experimenter measures). The extended system can account for the inference that, after testing the flashlight, Sarah would know that the batteries need to be charged.

We will proceed as follows. In Section~\ref{sec:causal_models} we introduce what we take here to be the basic system of causal reasoning: a slightly generalised form of the logic of causal reasoning introduced by Judea Pearl and collaborators and further developed in \citet{halpern2000axiomatizing,Pearl09,briggs2012interventionist}. In Section~\ref{sec:proposal} we introduce the notion of an {\it epistemic causal model} and extend the language with (i) a knowledge  operator and (ii) public announcement. Section~\ref{sec:no_learning} discusses the shortcomings of this first extension and motivates the new system that will be introduced in Section~\ref{sec:observables}. We finish the paper with the conclusions and an outlook on future work in Section~\ref{sec:conclusion}. For all systems discussed in the paper, we present a sound and strongly complete axiomatization. For the proofs of the main results stated in the paper the reader can consult the Appendix.

\section{Causal models}\label{sec:causal_models}

We start by introducing a basic system of causal reasoning. This logic goes back to \citet{Galles} (based on the work of structural causal models in \citealp{Pearl95}), and was further developed in, among others, \citet{halpern2000axiomatizing,Pearl09,briggs2012interventionist}. We slightly generalise the framework (along the lines of \citealp{briggs2012interventionist}) by allowing for interventions on exogenous variables.\footnote{This choice might be confusing for some readers. Exogenous variables are often seen as variables out of our control; theoretical entities that explain the (causal) behavior of those variables we actually want to model. In this paper we take the position that this is a conceptual choice that should not depend on formal restrictions. Thus, as far as possible exogenous variables will be treated the same way as endogenous variables.}\label{fn:exo}

\myparagraph{The structure.} The starting point is a formal representation of causal dependencies. This is done through \emph{causal models}, which represent the causal relationships between a finite set of variables. The variables are sorted into the set $\XV$ of \emph{exogenous} variables (those whose value is causally independent from the value of every other variable in the system) and the set $\NV$ of \emph{endogenous} variables (those whose value is completely determined by the value of other variables in the system). Note that $\XV\cap\NV=\emptyset$. Each variable $X \in \XV \cup \NV$ is assigned a non-empty and finite range $\Ran(X)$, which contains the possible values the variable can take. All this information is provided by a \emph{signature}.

\begin{definition}[Signature]\label{def:signature}
  Throughout this text, let $\Sig = \tuple{\XV, \NV, \Ran}$ be a \emph{finite} signature where
  \begin{compactitemize}
    \item $\XV = \set{U_1, \dots, U_m}$ is the finite set of \emph{exogenous} variables,
    \item $\NV = \set{V_1, \ldots, V_n}$ is the finite set of \emph{endogenous} variables, and
    \item $\Ran(X)$ is the finite non-empty range of $X$, for $X$ a variable in $\XV \cup \NV$.\footnote{Given $(X_1,\ldots,X_k) \in (\XV \cup \NV)^k$, abbreviate $\Ran(X_1) \times \cdots \times \Ran(X_k)$ as $\Ran(X_1, \ldots, X_k)$.}
  \end{compactitemize}
  Define the set $\AV := \XV \cup \NV$ and, for $X \in \AV$, the abbreviation $\allminus{X} := \AV \setminus \set{X}$. For simplicity, tuples of variables will be sometimes manipulated by set operations. To do this properly, we assume a canonical order over $\AV$ (say, $\tuple{U_1, \dots, U_m, V_1, \ldots, V_n}$), which yields a one-to-one correspondence between a subset of $\AV$ and the tuple containing its elements in the canonical order.
\end{definition}

\noindent A causal model is formally defined as follows.

\begin{definition}[Causal model]\label{def:causal_model}
  A \emph{causal model} is a triple $\tuple{\Sig, \F, \A}$ where
  \begin{compactitemize}
    \item $\Sig = \tuple{\XV,\NV,\Ran}$ is the model's signature,
    \item $\F = \set{ \fun_{V_j} \mid V_j \in \NV }$ contains, for each endogenous variable $V_j \in \NV$, a map
    \[
      \fun_{V_j}:\Ran(U_1, \ldots, U_m, V_1, \ldots, V_{j-1}, V_{j+1}, \ldots, V_n) \to \Ran(V_j).
    \]
    Each $\fun_V$ is $V$'s \emph{structural function}.

    \item $\A$ is a \emph{valuation} function, assigning to every $X \in \AV$ a value $\A(X) \in \Ran(X)$.\footnote{Given $(X_1,\ldots,X_k) \in (\XV \cup \NV)^k$, abbreviate $(\A(X_1), \ldots \A(X_k))$ as $\A(X_1, \ldots, X_k)$.} The valuation must comply with the exogenous variables' structural functions: for each endogenous variable $V_j \in \NV$ we have
    \[ \A(V_j) = \fun_{V_j} \left( \A(U_1, \ldots, U_m, V_1, \ldots, V_{j-1}, V_{j+1}, \ldots, V_n) \right). \]
  \end{compactitemize}
\end{definition}

\noindent In a causal model $\tuple{\Sig, \F, \A}$, the functions in $\F$ describe the causal relationship between the variables. Using these functional dependencies, we can define what it means for a variable to directly causally affect another variable.\footnote{This notion of a {\it direct cause}, adopted from \citet{Galles}, is related to the notion {\it direct effect} of a variable to another, as discussed in \citet{Pearl09} (for Causal Bayes Nets). The notion used here differs from Halpern's notion of {\it affect} \citep{halpern2000axiomatizing}. This difference shows up in the axiomatization: axiom \ax{A_6} (Table \ref{tbl:rcm-LA}) has the same function as \ax{C6} in \citet{halpern2000axiomatizing} (making the canonical model recursive), but does so in a slightly different way.}

\begin{definition}[Causal dependency]\label{def:parentof}
   Let $\F$ be a set of structural functions for $\NV$. Given $V_j \in \NV$, rename the  variables in $\allminus{V_j}$ as $X_1, \ldots, X_{m+n-1}$.

   Under $\F$, an endogenous variable $V_j \in \NV$ is \emph{directly causally affected} by a variable $X_i \in \allminus{V_j}$ (in symbols, $X_i \dirparof{\F} V_j$) if and only if there is a tuple
   \[
     (x_1, \ldots, x_{i-1}, x_{i+1}, \ldots, x_{m+n-1})
     \in
     \Ran(X_1, \ldots, X_{i-1}, X_{i+1}, \ldots, X_{m+n-1})
  \]
  and there are $x'_i \neq x''_i \in \Ran(X_i)$ such that
  \[ \fun_{V_j} (x_1, \ldots, x'_{i}, \ldots, x_{m+n-1}) \neq \fun_{V_j} (x_1, \ldots, x''_{i}, \ldots, x_{m+n-1}). \]
  When $X_i \dirparof{\F} V_j$, we will also say that $X_i$ is a \emph{causal parent} of $V_j$. The pair $\tuple{\AV, \dirparof{\F}}$ is called the \emph{causal graph} induced by $\F$. The relation $\parof{\F}$ is the transitive closure of $\dirparof{\F}$.
\end{definition}


\noindent As it is common in the literature, we restrict ourselves to causal models in which circular causal dependencies do not occur.\footnote{This is because only acyclic relations are thought to have a causal interpretation (see, e.g., \citealp{StrWol1960}). Counterfactuals behave differently if cyclic dependencies are allowed (see \citealp{halpern2000axiomatizing}).}

\begin{definition}[Recursive causal model]\label{def:recursive}
  A set of structural functions $\F$ is \emph{recursive} if and only if $\parof{\F}$ is a strict partial order (i.e., an asymmetric and transitive relation).\footnote{\label{ftn:recursive} Alternatively (but equivalently), $\F$ is \emph{recursive} 
  if and only if $\dirparof{\F}$ does not contain cycles
  .}   A causal model $\tuple{\Sig, \F, \A}$ is \emph{recursive} if and only if $\F$ is recursive. Here, a recursive causal model will be called simply a \emph{causal model}.
\end{definition}

\noindent An important feature of recursive causal models is that, once the value of all exogenous values is fixed, the value of all endogenous variables is uniquely determined (see, e.g., \citealp{halpern2000axiomatizing}).



\myparagraph{Interventions.} The most important concept in this approach to causal reasoning is that of \emph{intervention}: an action that changes the values of the system's variables. Before defining it formally, we introduce the notion of assignment.

\begin{definition}[Assignment, subassignment]
  Given $k \in \Nat$ and a set of variables $\set{X_1, \ldots, X_k} \subseteq \AV$, an \emph{assignment} on $\Sig$ is a function that allocates a value in $\Ran(X_i)$ to each variable $X_i$. For simplicity, an assignment will be denoted as $\opint{\vec{X}}{\vec{x}}$, with $\vec{X} = (X_1, \ldots, X_k)$ the assignment's domain (thus, with $X_i \neq X_j$ for $i \neq j$) and $\vec{x} \in \Ran(\vec{X})$ the tuple containing a value for each variable in $\vec{X}$.\footnote{Note how taking $k=0$ yields the \emph{empty assignment}.} An assignment $\opint{\vec{X'}}{\vec{x'}}$ is a \emph{subassignment} of $\opint{\vec{X}}{\vec{x}}$ if and only if $\vec{X'} \subseteq \vec{X}$ and $\vec{x'}$ is the restriction of $\vec{x}$ to the values indicated for variables in $\vec{X'}$.
\end{definition}

\noindent Given an assignment $\opint{\vec{X}}{\vec{x}}$, an intervention that sets each variable in $\vec{X}$ to its respective value in $\vec{x}$  can be defined as an operation that maps a given causal model $M$ to a new model $M_{\opint{\vec{X}}{\vec{x}}}$.

\begin{definition}[Intervention]\label{def:int}
  Let $M = \tuple{\Sig, \F, \A}$ be a causal model with $\opint{\vec{X}}{\vec{x}}$ an assignment on $\Sig$. In the causal model $M_{\opint{\vec{X}}{\vec{x}}} = \tuple{\Sig, \newf{\F}{\opint{\vec{X}}{\vec{x}}}, \newval{\A}{\F}{\opint{\vec{X}}{\vec{x}}}}$, which results from an intervention setting the values of variables in $\vec{X}$ to $\vec{x}$,
  \begin{compactitemize}
    \item the functions in $\newf{\F}{\opint{\vec{X}}{\vec{x}}} = \set{ \fun'_{V} \mid V \in \NV }$ are such that \begin{inlineenum} \item for each $V$ \emph{not in $\vec{X}$}, the function $\fun'_{V}$ is exactly as $\fun_{V}$, and \item for each $V = X_i \in \vec{X}$, the function $\fun'_{X_i}$ is a \emph{constant} function returning the value $x_i \in \vec{x}$ regardless of the values of all other variables\end{inlineenum}.

    \item $\newval{\A}{\F}{\opint{\vec{X}}{\vec{x}}}$ is the unique valuation (uniqueness will be proved below) where \begin{inlineenum} \item the value of each exogenous variable in $\vec{X}$ is the one in $\vec{x}$, \item the value of each exogenous variable not in $\vec{X}$ is as in $\A$, and \item each endogenous variable complies with its \emph{new} structural function (that in $\newf{\F}{\opint{\vec{X}}{\vec{x}}}$)\end{inlineenum}. In other words,  $\newval{\A}{\F}{\opint{\vec{X}}{\vec{x}}}(Y)
      $ is the unique valuation satisfying the following equations:
    \[
      \newval{\A}{\F}{\opint{\vec{X}}{\vec{x}}}(Y)
      =
      \left\{
        \begin{array}{ll}
          x_i                                                              & \text{if}\; Y \in \XV \;\text{and}\; Y = X_i \in \vec{X} \\
          \A(Y)                                                            & \text{if}\; Y \in \XV \;\text{and}\; Y \notin \vec{X} \\
          \fun'_Y(\newval{\A}{\F}{\opint{\vec{X}}{\vec{x}}}(\allminus{Y})) & \text{if}\; Y \in \NV. \\
        \end{array}
      \right.
    \]
  \end{compactitemize}
\end{definition}

\noindent Note that $\newval{\A}{\F}{\opint{\vec{X}}{\vec{x}}}$ can be equivalently defined as the (again, unique) valuation where \begin{inlineenum} \item the value of each $X_i \in \vec{X}$ is the indicated $x_i \in \vec{x}$, \item the value of each exogenous variable not in $\vec{X}$ is exactly as in $\A$, and \item the value of each endogenous variable not in $\vec{X}$ complies with its structural function (in $\F$ or in $\newf{\F}{\opint{\vec{X}}{\vec{x}}}$, as the functions for endogenous variables not in $\vec{X}$ remain the same)\end{inlineenum}. In this reformulation it becomes more transparent that an intervention with empty assignment does not affect the given causal model.

{\medskip}

An important observation with respect to Definition~\ref{def:int} is that intervention preserves (recursive) causal models.

\begin{proposition}\label{pro:int-preserves}
  Let $\opint{\vec{X}}{\vec{x}}$ be an assignment. If $M$ is a (recursive) causal model, then so is $M_{\opint{\vec{X}}{\vec{x}}}$.
  \begin{proof}
    For showing that $M_{\opint{\vec{X}}{\vec{x}}}$ is a causal model, one needs to show that its valuation $\newval{\A}{\F}{\opint{\vec{X}}{\vec{x}}}$ complies with the structural functions $\newf{\F}{\opint{\vec{X}}{\vec{x}}}$. This is given directly by the third case of the definition of $\newval{\A}{\F}{\opint{\vec{X}}{\vec{x}}}$. For showing that $M_{\opint{\vec{X}}{\vec{x}}}$ is recursive, simply notice that the intervention only removes causal dependencies, and thus no circular dependencies can be created.
  \end{proof}
\end{proposition}

\noindent Lastly, we prove that indeed there is a unique valuation satisfying the constraints imposed by the definition of intervention.

\begin{corollary}
  In $M_{\opint{\vec{X}}{\vec{x}}}$, the valuation $\newval{\A}{\F}{\opint{\vec{X}}{\vec{x}}}$ is uniquely determined.
  \begin{proof}
   In $M_{\opint{\vec{X}}{\vec{x}}}$, the value of every exogenous variable $U$ is fixed, either from $\vec{x}$ (if $U$ occurs in $\vec{X}$) or from $\A$ (otherwise). Then, by $\newf{\F}{\opint{\vec{X}}{\vec{x}}}$' recursiveness (Proposition \ref{pro:int-preserves}), each endogenous variable gets a unique value. 
  \end{proof}
\end{corollary}

\myparagraph{The language.} Now, everything is in place for introducing the object language.

\begin{definition}[Language \LAc]
  Formulas $\varphi$ of the language \LAc based on the signature $\Sig$ are given by
  \[ \varphi ::= Y{=}y \mid \lnot \varphi \mid \varphi \land \varphi \mid \int{\opint{\vec{X}}{\vec{x}}}{\varphi} \]
  for $Y \in \AV$, $y \in \Ran(Y)$ and $\opint{\vec{X}}{\vec{x}}$ an assignment on $\Sig$. The expression $\int{\opint{\vec{X}}{\vec{x}}}{\varphi}$ should be read as a counterfactual conditional: ``if the variables in $\vec{X}$ were set to the values $\vec{x}$, respectively, then $\varphi$ would be the case''.\footnote{Note: $\int{\;}{\varphi}$ expresses that $\varphi$ holds after an intervention with the empty assignment.} The assignment $\opint{\vec{X}}{\vec{x}}$ is called the counterfactual's \emph{antecedent}, and $\varphi$ is called its \emph{consequent}. Other Boolean operators ($\vee, \rightarrow, \leftrightarrow$) are defined as usual.

  For the semantic interpretation, let $\tuple{\Sig, \F, \A}$ be a causal model. Boolean operators are evaluated as usual; for the rest,
  \begin{ctabular}{l@{\quad{iff}\quad}l}
    $\tuple{\Sig, \F, \A} \models Y{=}y$                                & $\A(Y) = y$, \\
    $\tuple{\Sig, \F, \A} \models \int{\opint{\vec{X}}{\vec{x}}}{\varphi}$ & $\tuple{\Sig, \newf{\F}{\opint{\vec{X}}{\vec{x}}}, \newval{\A}{\F}{\opint{\vec{X}}{\vec{x}}}} \models \varphi$. \\
  \end{ctabular}
  A formula $\varphi \in \LAc$ is valid w.r.t \emph{recursive} causal models (notation: $\models \varphi$) if and only if $M \models \varphi$ for every recursive causal model $M$.
\end{definition}

\noindent The language \LAc has more restrictions than that in \citet{briggs2012interventionist}: while here a counterfactual's antecedent $\opint{\vec{X}}{\vec{x}}$ is effectively only a conjunction of requirements (set $X_1$ to $x_1$ \emph{and} so on), in \citet{briggs2012interventionist} the antecedent might contain disjunctions. Yet, \LAc has less restrictions than the languages in \citet{halpern2000axiomatizing}, as it allows complex formulas in a counterfactual's consequent.\footnote{Thus, \LAc has also less restrictions than the languages in \citet{GallesPearl1997,Galles}.} Note also how, in contrast to most literature on causal models, \LAc can talk about the values of exogenous variables, and also allows interventions on them.\footnote{Some interventionist approaches to counterfactuals, such as \citet{briggs2012interventionist} and \citet{BarSan2020}, also allow interventions on exogenous variables.}

Note also how \LAc is expressive enough to characterise syntactically the notion of direct causal effect $\dirparof{}$. Indeed, recall (Definition \ref{def:parentof}) that $X \dirparof{\F} V$ holds in a given $\tuple{\Sig, \F, \A}$ if and only if there are values $\vec{z}$ for the variables in $\vec{Z} = \AV \setminus \set{X, V}$ and two different values $x_1, x_2$ for $X$ such that the value $V$ gets from $\fun_V$ by setting $(\vec{Z},X)$ to $(\vec{z}, x_1)$ is different from the value it gets from $\fun_V$ by setting the same variables to $(\vec{z}, x_2)$. This is expressed by the formula
\[
  \bigvee_{
    \renewcommand{\arraystretch}{1}
    \begin{array}{l}
      \vec{z} \in \Ran(\AV \setminus \set{X,V}), \\
      \set{x_1,x_2} \subseteq \Ran(X),\, x_1 \neq x_2, \\
      \set{v_1, v_2} \subseteq \Ran(V),\, v_1 \neq v_2
    \end{array}
  }
  \int{\opint{\vec{Z}}{\vec{z}},\opint{X}{x_1}}{(V{=}v_1)} \;\land\; \int{\opint{\vec{Z}}{\vec{z}},\opint{X}{x_2}}{(V{=}v_2)},
\]
which is abbreviated as $X \syndirparof V$ (cf. with the syntactic definition of causal dependency in \citealp{halpern2000axiomatizing}). Thus, for any causal model $\tuple{\Sig, \F, \A}$ and any variables $X \in \AV$ and $V \in \NV$,
\[
  \tuple{\Sig, \F, \A} \models X \syndirparof V
  \qquad\text{if and only if}\qquad
  X \dirparof{\F} V
\]

\myparagraph{Axiom system.} As stated, \LAc is different from previous languages used for causal models. Thus, it is worthwhile to provide its axiom system. The system, whose axioms and rules appear in Table \ref{tbl:rcm-LA}, uses as a parameter the signature $\Sig$ of both the language and the intended class of models (in some cases relying on the signature's finiteness). Yet, note that no axiom refers to a particular model.

Note the role played by the signature's \emph{finiteness}, i.e., by the fact that there are only a finite number of variables, each one of them having a finite range. Just as in \citet{halpern2000axiomatizing}, axiom \ax{A_2} makes use of the second condition to indicate that a variable gets assigned a value in its range, and axiom \ax{A_6} makes use of both to characterise recursive models (via the abbreviation $X \syndirparof V$).\footnote{Still, \citet[Page 323]{halpern2000axiomatizing} also mentions how to deal with infinite variables with infinite range: in such cases, axiom \ax{A_2} is required only for variables with finite range, and axiom \ax{A_6} can be replaced \emph{``by all its instances''}. (This somehow vague suggestion is turned into a concrete rule in \citet{briggs2012interventionist}, although that paper only addresses finite signatures.) The rest of the setting remains unchanged. An alternative approach for dealing with infinite variables and infinite ranges is discussed in  \citet{IbelingI19}.}

\begin{table}[ht!]
  \begin{center}
   \begin{footnotesize}
     \renewcommand{\arraystretch}{1.7}
     \begin{tabular}{l@{\;}l}
       \toprule
       \ax{P}        & $\vdashloc \varphi$ \hfill for $\varphi$ an instance of a propositional tautology \\
       \ax{MP}       & From $\varphi_1$ and $\varphi_1 \rightarrow \varphi_2$ infer $\varphi_2$ \\
       \midrule
       \ax{A_1}      & $\vdashloc \int{\opint{\vec{X}}{\vec{x}}}{Y{=}y} \rightarrow \lnot \int{\opint{\vec{X}}{\vec{x}}}{Y{=}y'}$ \hfill for $y,y' \in \Ran(Y)$ with $y \neq y'$ \\
       \ax{A_2}      & $\vdashloc \bigvee_{y \in \Ran(Y)} \int{\opint{\vec{X}}{\vec{x}}}{Y{=}y}$ \\
       \ax{A_3}      & $\vdashloc \left( \int{\opint{\vec{X}}{\vec{x}}}{(Y{=}y)} \land \int{\opint{\vec{X}}{\vec{x}}}{(Z{=}z)} \right) \rightarrow \int{\opint{\vec{X}}{\vec{x}}, \opint{Y}{y}}{(Z{=}z)}$ \\
       \ax{A_4}      & $\vdashloc \int{\opint{\vec{X}}{\vec{x}}, \opint{Y}{y}}{(Y{=}y)}$ \\
       \ax{A_5}      & $\vdashloc \left( \int{\opint{\vec{X}}{\vec{x}}, \opint{Y}{y}}{(Z{=}z)} \land \int{\opint{\vec{X}}{\vec{x}}, \opint{Z}{z}}{(Y{=}y)} \right) \rightarrow \int{\opint{\vec{X}}{\vec{x}}}{(Z{=}z)}$ \hfill for $Y \neq Z$ \\
       \ax{A_6}      & $\vdashloc (X_0 \syndirparof X_1\wedge \cdots \wedge X_{k-1} \syndirparof X_k) \rightarrow \lnot (X_k \syndirparof X_0)$ \\

       \ax{A_7}      & $\vdashloc \int{\;}{U{=}u} \leftrightarrow \int{\opint{\vec{X}}{\vec{x}}}{U{=}u}$ \hfill for $U \in \XV$ with $U \notin\vec{X}$ \\
       \midrule
       \ax{A_{[]}}   & $\vdashloc Y{=}y \leftrightarrow \int{\;}{Y{=}y}$ \\
       \ax{A_\lnot}  & $\vdashloc \int{\opint{\vec{X}}{\vec{x}}}{\lnot \varphi} \leftrightarrow \lnot \int{\opint{\vec{X}}{\vec{x}}}{\varphi}$ \\
       \ax{A_\land}  & $\vdashloc \int{\opint{\vec{X}}{\vec{x}}}{(\varphi_1\wedge\varphi_2)} \leftrightarrow (\int{\opint{\vec{X}}{\vec{x}}}{\varphi_1} \land \int{\opint{\vec{X}}{\vec{x}}}{\varphi_2})$ \\
       \ax{A_{[][]}} & $\vdashloc \int{\opint{\vec{X}}{\vec{x}}}{\int{\opint{\vec{Y}}{\vec{y}}}{\varphi}} \leftrightarrow \int{\opint{\vec{X'}}{\vec{x'}}, \opint{\vec{Y}}{\vec{y}}}{\varphi}$ \qquad \hfill \begin{minipage}[t]{0.35\textwidth}\begin{flushright} with $\opint{\vec{X'}}{\vec{x'}}$ the subassignment of $\opint{\vec{X}}{\vec{x}}$ for $\vec{X'} := \vec{X} \setminus \vec{Y}$ \smallskip \end{flushright}\end{minipage} \\
       \bottomrule
     \end{tabular}
   \end{footnotesize}
  \end{center}
  \caption{Axiom system \LOc, for \LAc-formulas valid in recursive causal models.}
  \label{tbl:rcm-LA}
\end{table}

\begin{theorem}[Axiom system for \LAc]\label{thm:rcm-LA}
  Recall that $\Sig = \tuple{\XV, \NV, \Ran}$ is a finite signature. The axiom system \LOc, whose axioms and rules are shown in Table \ref{tbl:rcm-LA}, is sound and strongly complete for the language \LAc based on $\Sig$ with respect to recursive causal models for $\Sig$.
  \begin{proof}
    See Appendix \ref{app:rcm-LA}.
  \end{proof}
\end{theorem}

\section{Epistemic causal models}\label{sec:proposal}

Section~\ref{sec:causal_models} introduced the logic of causal reasoning that we will work with. To this logic we will now add an epistemic component. We first extend the causal model with a representation of the epistemic state of an agent. It is assumed that, while the agent knows the causal laws, she might not know the value of some variables. This uncertainty can be represented in the form of a set of valuations $\T$, which represents the alternatives the agent considers possible.\footnote{The choice of the name $\T$ is made for analogy with causal team semantics \citep{barbero2019interventionist,BarSan2020}, where a set of valuations is called a ``team''. The way we use the ``team'' here is purely modal: the formulas take truth values at single valuations, and $\T$ is the set of ``worlds'' accessible from this valuation. In team semantics this local perspective is absent.} To model the assumption that the agent knows the causal laws we require that all possible valuations comply with the same underlying set of structural functions.

\begin{definition}[Epistemic causal model]
  An \emph{epistemic (note: recursive) causal model} $\ec$ is a tuple $\tuple{\Sig, \F, \T}$ where $\Sig = \tuple{\XV, \NV, \Ran}$ is a signature, $\F$ is a (note: recursive) set of structural functions for $\NV$, and $\T$ is a non-empty set of valuation functions for $\XV \cup \NV$, each one of them complying with $\F$.\footnote{A remark. In standard \textit{EL}, an epistemic possibility is a world, and each world gets assigned an atomic valuation; hence, the same atomic valuation might be assigned to two different worlds in the agent's epistemic range. In our setting, an epistemic possibility is a valuation; hence, the same valuation cannot appear twice in the set $\T$. Because of this, and given the finiteness of $\AV$, each epistemic possibility (valuation) $\A$ can be characterised by the conjunction $\bigwedge_{X \in \AV} X=\A(X)$; still, not every such formula characterises an epistemic possibility, as not all valuations need to be considered possible. (Cf. the \emph{nominals} in hybrid logic [\citealp{ArecestenCate2006,sep-logic-hybrid}]: in named models, each epistemic possibility (world) is characterised by a nominal, and every nominal names a unique epistemic possibility.)}
\end{definition}

\noindent Using this new notion of an epistemic causal model we can now formalise the context of Example \ref{flashlight} from the introduction. We define an epistemic causal model $\ev =\tuple{\Sig, \F, \T}$ whose signature $\Sig$ has three variables: the exogenous variables $P$ for pushing the button and $B$ for the batteries being full, and the endogenous variable $L$ for the flashlight emitting light. All three variables can take two values, $0$ or $1$. The set of functions $\F$ contains only  one element: the function mapping $L$ to $1$ iff $P=1$ and $B=1$. It seems natural to assume that the agent is aware of the state of the button (not pressed) and the lamp (off), so the set $\T$ contains the valuation $\A_1$ that maps $B$ to $0$, $P$ to $0$ and $L$ to $0$, and the valuation $\A_2$\label{A2} that maps $B$ to $1$, $P$ to $0$ and $L$ to $0$. Note how $\T$ cannot contain the valuation $B=0$, $P=1$ and $L=1$, for instance, because this valuation does not comply  with the causal law in $\F$. This observation highlights an important assumption we made above: there is no uncertainty about the causal dependencies. Investigating the consequences of lifting this restriction is left for future research.

{\smallskip}

The next step is to extend the notion of intervention to epistemic causal models, see Definition~\ref{def:int:epis}.

\begin{definition}[Intervention on epistemic causal models]\label{def:int:epis}
  Let $\ec = \langle \Sig$, $\F, \T \rangle$ be an epistemic causal model; let $\opint{\vec{X}}{\vec{x}}$ be an assignment on $\Sig$. The epistemic causal model $\ec_{\opint{\vec{X}}{\vec{x}}} = \tuple{\Sig, \F_{\opint{\vec{X}}{\vec{x}}}, \newval{\T}{\F}{\opint{\vec{X}}{\vec{x}}}}$, resulting from an intervention setting the values of variables in $\vec{X}$ to $\vec{x}$, is such that
  \begin{itemize}
    \item $\F_{\opint{\vec{X}}{\vec{x}}}$ is defined from $\F$ just as in Definition \ref{def:int},

    \item $\newval{\T}{\F}{\opint{\vec{X}}{\vec{x}}} := \set{ \newval{\B}{\F}{\opint{\vec{X}}{\vec{x}}} \mid \B \in \T}$ (see Definition \ref{def:int}).
  \end{itemize}
\end{definition}

\noindent In this definition, $\tuple{\Sig, \F_{\opint{\vec{X}}{\vec{x}}}, \newval{\T}{\F}{\opint{\vec{X}}{\vec{x}}}}$ is indeed an epistemic causal model: $\F_{\opint{\vec{X}}{\vec{x}}}$ is recursive and all valuations in $\newval{\T}{\F}{\opint{\vec{X}}{\vec{x}}}$ comply with it. With this we can now calculate the effects of an intervention $P=1$ on the epistemic causal model $\ev$ defined above. According to Definition~\ref{def:int:epis}, an intervention on an epistemic causal model amounts to intervening on each of the valuations contained in the epistemic state. Thus, for our concrete example, we need to calculate the effects of an intervention with $P=1$ on the valuations $\A_1$ and $\A_2$ that make up the epistemic state $\T$. It is not difficult to see that the new epistemic state $\T^{\F}_{P=1}$ will contain the valuation $\A^{\F}_{1, P=1}$ that maps $B$ to $0$, $P$ to $1$ and $L$ to $0$ and the valuation $\A^{\F}_{2,P=1}$ that maps $B$ to $1$, $P$ to $1$ and $L$ to $1$.

An important assumption underlying this definition of intervention is that the agent has full epistemic access to the effect of the intervention on the model. In particular, she knows that the intervention takes place (in the counterfactual scenario considered). This is reasonable if one thinks of the agent whose epistemic state is being modelled as the one engaging in counterfactual thinking. It is less plausible in connection to counterfactual thinking about the knowledge states of other agents. But this is something that we can leave for now, as we are not considering epistemic causal models for multiple agents in this paper.

{\smallskip}

Based on these changes on the semantic side, we can extend the object language with a modality for talking about the epistemic state of the agent.

\begin{definition}[Language \LAkc]
  Formulas $\xi$ of the language \LAkc based on $\Sig$ are given by
  \[ \xi ::= Y{=}y \mid \lnot \xi \mid \xi \land \xi \mid K\xi \mid \int{\opint{\vec{X}}{\vec{x}}}{\xi}  \]
  for $Y \in \AV$, $y \in \Ran(Y)$ and $\opint{\vec{X}}{\vec{x}}$ an assignment on $\Sig$. Formulas of the form $K\xi$ are read as ``the agent knows $\xi$''.
\end{definition}

\noindent The semantics for this epistemic language is straightforward.

\begin{definition}\label{def:LAk:semint}
  Formulas in \LAkc are evaluated in a pair $(\ec, \A)$ with $\ec = \tuple{\Sig, \F, \T}$ an epistemic causal model and $\A \in \T$. The semantic interpretation for Boolean operators is the usual; for the rest,
  \begin{cbtabular}{l@{\qquad{iff}\qquad}l}
    $(\ec, \A) \models Y{=}y$                             & $\A(Y) = y$, \\
    $(\ec, \A) \models \int{\opint{\vec{X}}{\vec{x}}}\xi$ & $(\ec_{\opint{\vec{X}}{\vec{x}}}, \newval{\A}{\F}{\opint{\vec{X}}{\vec{x}}}) \models \xi$, \\
    $(\ec, \A) \models K\xi$                              & $(\ec, \B) \models \xi$ for every $\B \in \T$. \\
  \end{cbtabular}
\end{definition}

\noindent To illustrate this definition, we go back to the epistemic model $\ev$ introduced for Example~\ref{flashlight}. For evaluating a concrete formula with respect to this model we need to select, next to $\ev$, a valuation representing the actual world. In the example this is valuation $\A_1$: in the actual world, the batteries are empty. We can calculate that the counterfactual $[P{=}1]L{=}0$ comes out as true given $\ev$ and $\A_1$, just as in the non-epistemic approach (Section~\ref{sec:causal_models}). But because we now also have a representation of the epistemic state of some agent, we can additionally consider epistemic attitudes the agent has towards this counterfactual. For instance, we can check that $K([P{=}1]L{=}0)$ is not true given $\ev$ and $\A_1$: the agent cannot predict the outcome of pressing the button in this situation -- this was one of the inferences we discussed in the introduction. This sentence is only true in case the formula $[P{=}1]L{=}0$ holds in both $(\ev, \A_1)$ and $(\ev, \A_2)$ (where $\A_1,\A_2$ are the two elements of $\T$). 
However, while in $\A^\F_{2, P{=}1}$ the flashlight is emitting light (in this possibility the battery is charged), this is not the case in  $\A^\F_{1, P{=}1}$. Thus, $(\ev_{P{=}1}, \A^{\F}_{2, P{=}1}) \not \models L{=}0$ and, hence, $(\ev, \A_1) \not \models K([P{=}1]L{=}0)$. Thus, the agent cannot predict whether pressing the button of the flashlight will turn on the light, just as intended in this case.

\myparagraph{Axiom system.} To get a clearer idea of the laws governing the interaction between knowledge and causality in the proposed system, we can look for an axiom system characterising the formulas in \LAkc that are valid in epistemic (recursive) causal models. A sound and complete axiom system for the proposed logic is given by the axiom system \LOkc, which extends \LOc (Table \ref{tbl:rcm-LA}) with the axioms and rules from Table \ref{tbl:rcm-LAk}; thus, once again, it is relative to a given signature $\Sig$. Among the new axioms, those in the \textit{epistemic} part form the standard modal S5 axiomatization for truthful knowledge with positive and negative introspection (see, e.g., \citealp{RAK1995}). Then, axiom \ax{CM} indicates that what the agent will know after an intervention ($\int{\opint{\vec{X}}{\vec{x}}}K\xi$) is exactly what she knows now about the effects of the intervention ($K\int{\opint{\vec{X}}{\vec{x}}}\xi$). Although maybe novel in the literature on causal models, the axiom is simply an instance of the more general pattern of interaction between knowledge and a deterministic action without precondition. Axiom \ax{KL} indicates that the agent knows how each endogenous variable $Y \in \NV$ is affected when \emph{all other variables} are intervened; it can be understood as stating that the agent knows the causal laws.

\begin{table}[ht!]
  \begin{center}
   \begin{footnotesize}
     \renewcommand{\arraystretch}{1.5}
     \begin{tabular}{l@{\;}l@{\quad}l@{\;}l}
       \toprule
       \multicolumn{4}{l}{\textit{Epistemic}:} \\
       \ax{K}   & $\vdashlokc K(\xi_1\rightarrow\xi_2)\rightarrow(K\xi_1\rightarrow K\xi_2)$ &
       \ax{T}   & $\vdashlokc K\xi\rightarrow\xi$ \\
       \ax{N_K} & From $\vdashlokc \xi$ derive $\vdashlokc K\xi$ &
       \ax{4}   & $\vdashlokc K\xi\rightarrow KK\xi$ \\
       \multicolumn{2}{l}{} &
       \ax{5}   & $\vdashlokc \neg K\xi\rightarrow K\neg K\xi$ \\
       \midrule
       \multicolumn{4}{l}{\textit{Epistemic+Intervention}:} \\
       \ax{CM}  & \multicolumn{3}{@{}l}{$\vdashlokc \int{\opint{\vec{X}}{\vec{x}}}K\xi \,\leftrightarrow\, K\int{\opint{\vec{X}}{\vec{x}}}\xi$} \\
       \ax{KL}  & \multicolumn{3}{@{}l}{$\vdashlokc \int{\opint{\vec{X}}{\vec{x}}}Y{=}y \,\rightarrow\, K\int{\opint{\vec{X}}{\vec{x}}}Y{=}y$ \qquad for $Y\in \NV$ and $\vec{X} = \AV \setminus \set{Y}$} \\
       \bottomrule
     \end{tabular}
   \end{footnotesize}
  \end{center}
  \caption{Additional axioms and rules for axiom system \LOkc, for \LAkc-formulas valid in epistemic (recursive) causal models.}
  \label{tbl:rcm-LAk}
\end{table}

\begin{theorem}[Axiom system for \LAkc]\label{thm:rcm-LAk}
  Recall that $\Sig = \tuple{\XV, \NV, \Ran}$ is a finite signature. The axiom system \LOkc, extending \LOc (Table \ref{tbl:rcm-LA}) with the axioms and rules in Table \ref{tbl:rcm-LAk}, is sound and strongly complete for the language \LAkc based on $\Sig$ with respect to epistemic (recursive) causal models for $\Sig$.
 \begin{proof}
    See Appendix \ref{app:rcm-LAk}.
  \end{proof}
\end{theorem}

\paragraph{Adding public announcement}

So far the extension of the logic of causal inference that we introduced is completely static from the epistemic point of view: there is no action that can induce a change in the information state of an agent. But this can be easily amended. For instance, we can `import' the well-known action of \emph{(truthful) public announcements} \citep{Plaza89,GerbrandyG97}, which in this single-agent setting can be understood as an act of observing/learning. This action corresponds, semantically, to a model operation that removes the epistemic possibilities in which the announced/observed/learnt formula does not hold. 
We first add public announcements to the syntax of the language and then provide a semantics for this extended language.

\begin{definition}[Language \LAfull]\label{def:ann:epis}
  Formulas $\chi$ of the language \LAfull based on $\Sig$ are given by
  \[ \chi ::= Y{=}y \mid \neg \chi \mid \chi \land \chi \mid K\chi \mid [\chi!]\chi \mid \int{\opint{\vec{X}}{\vec{x}}}{\chi} \]
  for $Y \in \AV$, $y \in \Ran(Y)$ and $\opint{\vec{X}}{\vec{x}}$ an assignment on $\Sig$. Formulas of the form $[\alpha!]\chi$ are read as ``after $\alpha$ is observed, $\chi$ is the case''.
\end{definition}

\begin{definition}[Semantics with announcements]\label{def:LAka:semint}
  Formulas in \LAfull are evaluated in a pair $(\ec, \A)$ with $\ec = \tuple{\Sig, \F, \T}$ an epistemic causal model and $\A \in \T$. Operators already in \LAkc are evaluated as before (Definition \ref{def:LAk:semint}); for announcements,
  \begin{cbtabular}{l@{\qquad{iff}\qquad}l}
    $(\ec, \A) \models [\alpha!]\chi$ & $(\ec, \A) \models \alpha$ implies $(\ec^{\alpha}, \A) \models \chi,$ \\
  \end{cbtabular}
  where $\ec^{\alpha} = \tuple{\Sig, \F, \T^{\alpha}}$ is such that $\T^{\alpha} := \set{\B \in \T \mid (\ec, \B) \models \alpha}$.
\end{definition}

\noindent Note how $\ec^{\alpha}$ is indeed an epistemic causal model: since $\ec$ is an epistemic causal model, $\F$ is recursive and all valuations in $\T^{\alpha}$ comply with it.

{\medskip}

Let us go back to Example \ref{flashlight} to illustrate the working of these definitions. 
Above we modelled the scenario using an epistemic causal model $\ev$ and a designated assignment $\A_1$ representing the actual world.
Let us assume that in the given context Sarah is told that the counterfactual $[P{=}1]L{=}0$ is true (public announcement). In this case the model correctly predicts that with this new information Sarah can conclude that the batteries are empty, i.e. $(\ev,\A_1)\models[[P{=}1]L=0!]K(B{=}0)$. Notice first that $(\ev,\A_1)\models [P{=}1]L{=}0$. So, we only need to check if $(\ev^{[P{=}1]L{=}0},\A_1)\models K(B{=}0)$, i.e. whether for all assignments $\A \in \T^{[P{=}1]L{=}0}$, $\A(B)=0$. Recall that $\T=\{\A_1,\A_2\}$ where $\A_1$ describes the scenario in which the battery is empty, the button not pressed and the light is off, and  $\A_2$ describes the scenario where the battery is charged, the button not pressed and the light is off. Only with respect to the first assignment $\A_1$ the counterfactual $[P{=}1]L{=}0$ is true. This means, following Definition~\ref{def:ann:epis}, that $\ev^{[P=1]E=0}=\{\A_1\}$. Furthermore, we have $\A_1(B)=0$. Therefore, following Definition~\ref{def:LAk:semint},  $(\ev,\A_1)\models[[P{=}1]L{=}0!]K(B{=}0)$, just as intended.\footnote{This example illustrates an interesting aspect of the system introduced in this section: it allows us to model a restricted form of update with counterfactual information. 
The only kind of information an agent can learn from counterfactuals is information about the value of so far unknown variables. In case the counterfactual that is announced is inconsistent with the given causal dependencies and the setting of independent variables, the update breaks down. 
}

\medskip

The languages \LAkc and \LAfull introduced in this section have many similarities to the $\CO$ languages proposed in \citet{barbero2019interventionist}; these also extend the basic causal language  with an operator (``selective implication'') which plays the role of an observation or announcement operator. However, the $\CO$ languages are not given a modal interpretation: they are interpreted in a variant of \emph{team semantics} \citep{Hod1997,Vaa2007}, which makes a comparison of the two languages a difficult task. In \citet{Dali_Proceedings} we have shown that, in the special case of \emph{recursive} models of \emph{finite} signature, there is a truth-preserving translation from $\CO$ to $\LAfull$.\footnote{The translation works also if $\CO$ is extended with \emph{dependence atoms} \citep{Vaa2007}.} The translation nicely brings out the point that team semantics interprets many formulas as preceded by a $K$ operator, even though this operator is not explicitly present in $\CO$.

\myparagraph{Axiom system.} We also provide a sound and complete axiomatization for this extension with public announcements. The axiom system \LOfull (relative to the signature $\Sig$) extends \LOkc (Tables \ref{tbl:rcm-LA} and \ref{tbl:rcm-LAk}) with the axioms and rules from Table \ref{tbl:rcm-LAka}. In the latter table, the rule in the first block is simply the necessitation rule for interventions; this rule was admissible in \LOkc but it is not anymore in the presence of announcement operators. The axioms and rules in the second block are well-known from public announcement logic; they essentially define a translation that eliminates the observation operator $[\alpha!]$. (See \citealp{wang2013axiomatizations} for a discussion on their role in the completeness proof.) The lone axiom in the third block, indicating a form of commutativity between interventions and observations, is what makes the translation work for our language, which makes free use of the intervention operators.\footnote{
Recall that, in contrast with the early literature on interventionist counterfactuals \citep{Galles,halpern2000axiomatizing}, here we allow nesting in the consequents of counterfactuals.}

\begin{table}[ht!]
  \begin{center}
   \begin{footnotesize}
     \renewcommand{\arraystretch}{1.5}
     \begin{tabular}{l@{\,}l@{\;\;\;}l@{\,}l}
       \toprule
        \ax{N_{=}}   & From $\vdashlofull \chi$ derive $\vdashlofull \int{\opint{\vec{X}}{\vec{x}}}\chi$ \\
        \midrule
        \ax{!_{=}}   & \multicolumn{3}{@{}l}{$\vdashlofull [\alpha!]\int{\opint{\vec{X}}{\vec{x}}}Y{=}y \,\leftrightarrow\, (\alpha \rightarrow \int{\opint{\vec{X}}{\vec{x}}}Y{=}y)$} \\
        \ax{!_\lnot} & $\vdashlofull [\alpha!]\neg \chi \,\leftrightarrow\, (\alpha \rightarrow \neg[\alpha!]\chi)$ &
        \ax{!_\land} & $\vdashlofull [\alpha!](\chi \wedge \chi) \,\leftrightarrow\, ([\alpha!]\chi \wedge [\alpha!]\chi)$ \\
        \ax{!_K}     & $\vdashlofull [\alpha!]K\chi \,\leftrightarrow\, (\alpha \rightarrow K(\alpha\rightarrow [\alpha!]\chi))$ &
        \ax{!_!}     & $\vdashlofull [\alpha_1!][\alpha_2!]\chi \leftrightarrow [\alpha_1 \land [\alpha_1!]\alpha_2!]\chi$ \\
        \ax{K_!}     & $\vdashlofull [\alpha!](\chi_1\rightarrow\chi_2)\rightarrow([\alpha!]\chi_1\rightarrow [\alpha!]\chi_2)$ &
        \ax{N_!}     & From $\vdashlofull \chi$ derive $\vdashlofull [\alpha!]\chi$ \\
        \ax{_!RE}    & \multicolumn{3}{@{}l}{From $\vdashlofull \alpha_1 \leftrightarrow \alpha_2$ derive $\vdashlofull [\alpha_1!]\chi \leftrightarrow [\alpha_2!]\chi$} \\
        \midrule
        \ax{=_!}     & \multicolumn{3}{@{}l}{$\vdashlofull \int{\opint{\vec{X}}{\vec{x}}}[\alpha!]\chi \,\leftrightarrow\, [\int{\opint{\vec{X}}{\vec{x}}}\alpha!]\int{\opint{\vec{X}}{\vec{x}}}\chi$} \\
       \bottomrule
     \end{tabular}
   \end{footnotesize}
  \end{center}
  \caption{Additional axioms and rules for axiom system \LOfull, for \LAfull-formulas valid in epistemic (recursive) causal models.}
  \label{tbl:rcm-LAka}
\end{table}

\begin{theorem}[Axiom system for \LAfull]\label{thm:rcm-LAka}
  Recall that $\Sig = \tuple{\XV, \NV, \Ran}$ is a finite signature. The axiom system \LOfull, extending \LOkc Tables \ref{tbl:rcm-LA} and \ref{tbl:rcm-LAk} with the axioms and rules on Table \ref{tbl:rcm-LAka}, is sound and strongly complete for the language \LAfull based on $\Sig$ with respect to epistemic (recursive) causal models for $\Sig$.
  \begin{proof}
    See Appendix \ref{app:rcm-LAka}.
  \end{proof}
\end{theorem}

\section{Limitations of the system: The ``no learning'' assumption}\label{sec:no_learning}

Section~\ref{sec:proposal} introduced an epistemic extension for the standard structural equational model of causal inference. We also saw that this extension can successfully account for some intuitive valid inferences concerning the interaction of causal and epistemic reasoning. However, when introducing Example~\ref{flashlight} we also discussed inferences that this extension cannot yet account for. Consider, for instance, the consequences of pressing the button. Such an action would increase Sarah's knowledge about the world: given that we assumed that she has epistemic access to the button and the light, she would be able to observe that the light stays off and conclude that the battery is empty.  This is an example for reasoning about
actual experiments, the kind of reasoning that this paper attempts to model. We can see clearly here how this experiment can increase the knowledge of our agent. 

However, the logic introduced in Section~\ref{sec:proposal} cannot account for this type of reasoning; in other words, the formal counterpart of the inference we just described, $[P{=}1]K(B{=}1)$, is not valid in the model $\ev$ from the example. The underlying reason for this is axiom \ax{CM} of $\LOkc \slash \LOfull$ (see Table~\ref{tbl:rcm-LAk} on Page~\pageref{tbl:rcm-LAk}). More specifically the left-to-right direction of this rule is causing trouble here.
\begin{equation*}\label{nolearning2}
   \ax{NL} \ \ \  [\vec{X}{=}\vec{x}]K\phi\rightarrow K[\vec{X}{=}\vec{x}]\phi.
\end{equation*}
In the literature, the principle \ax{NL} is known as the ``no learning'' property. In the language of dynamic epistemic logic, the property of ``no learning'' can be characterized by the formula $\bigcirc_{\pi}K\phi\rightarrow K\bigcirc_{\pi}\phi$, where $\bigcirc_{\pi}\phi$ stands for ``after executing an action $\pi$, $\phi$ will be the case''. In our case the relevant action is an intervention. Now, notice that by contraposition and by using axiom \ax{A_\lnot} we can derive from this sentence the principle $\neg K[\vec{X}{=}\vec{x}]\phi\rightarrow  [\vec{X}{=}\vec{x}]\neg K\phi$: if an agent cannot predict whether an intervention results in a certain outcome, then the agent will also not know this outcome after the intervention is actually performed. Intuitively, this means that an agent is unable to gain any new knowledge from doing an intervention.\footnote{See also \citet{HalpernMV04}: ``no learning'' expresses that uncertainty cannot be reduced when an action is executed.} The example above points out that \ax{NL}  should not be valid for the type of action that we are considering here. The entire point of experimentation is to provide us with new information. Thus, we need to look for a variation of the proposed logic that doesn't validate \ax{NL}.



How can we enable learning from experimentation in our logic? For this we need to change how an intervention affects the knowledge state of an agent.  The way the system is set up at the moment will warrant that the agent knows about the action itself (the intervention), but not necessarily about its consequences. These are computed on the epistemic state of the agent completely independently from what happens in the actual world. The intervention affects the knowledge state of the agent as if she {\it hypothetically considers} the intervention at the same time it is taking place. This is why there is no learning. Just {\it thinking} about an experiment will teach you nothing about the outside world. You need to {\it observe} the effects.

To make learning possible, we need to make sure that the value of those variables the agent has epistemic access to (the variables measured in the experiment) is updated when change occurs. We will call these variables {\it observables}. Thus, we need to make sure that, when an intervention occurs, the agent learns the value the observables take in the changed world. This, together with other knowledge the agent might have, determines the inferences she can draw.

Another way to put it is that we need to implement {\it knowing the value of a variable} differently. In the system introduced in Section~\ref{sec:proposal} we implement a basic Hintikka-style notion of knowledge: knowledge is simply true belief, which is known to be problematic (e.g., the Gettier cases). One very attractive alternative comes from Nozick \citep{Nozick}. He proposes that an agent knows $\psi$ if the sentence is true and if one's belief in it was acquired by a method such that if the sentence $\psi$ were not true, the method would not lead one to believe $\psi$, and such that if $\psi$ were true, one would believe $\psi$. In our case this method is observation (or whatever method the experimenter uses to measure the value of the observable variables). Reliance on such a method ensures that change in the world will translate directly into change in the epistemic state of the agent, which is exactly what we are after here.

\section{Semantics with observables}\label{sec:observables}


In the previous section we saw that the framework introduced in Section~\ref{sec:proposal} does not represent correctly the increase of knowledge produced by actual experiments. 
As explained above, in order to overcome this limitation we will extend our notion of model with a set of observables: variables to which the agent has direct epistemic access. The semantics we propose will enforce knowledge of the values of the observables by guaranteeing that the observables take a constant value in a model. Furthermore, we will adapt the notion of intervention to ensure that it preserves this property. But first we need to enrich the notion of a signature so that it singles out a set of observables.\footnote{As explained in Footnote \ref{fn:exo}, we look for a formal system that is as general (and thus as flexible) as possible, without introducing restrictions (on exogenous variables) that are not necessary. Thus, Definition~\ref{def:sig_with_obs} treats all variables as potentially observable, even though that might not be how exogenous variables are normally conceptualized.}

\begin{definition}[Signature and observables]\label{def:sig_with_obs}
  A {\it signature} 
  with observables is a quadruple $\tuple{\XV,\NV,\O,\Ran}$ where $\tuple{\XV,\NV,\Ran}$ is as in Definition \ref{def:signature} and $\O\subseteq \XV\cup\NV$. The set $\O$ is called the set of {\it observables}.
\end{definition}

\noindent Next we introduce the notion of an {\it epistemic causal model with observables}. This is an epistemic causal model with one extra property: the agent knows at least the value of all observables. This is warranted by the final condition of the following definition.

\begin{definition}[Epistemic causal model with observables] Consider \, a signature $\Sig= \langle \XV,\NV,\O,\Ran\rangle$. An {\it epistemic causal model with observables} is a triple $\langle\Sig,\F,\T\rangle$ such that $\langle\langle\XV,\NV,\Ran\rangle,\F,\T\rangle$ is an epistemic causal model and, for each $\A,\A'\in\T$, $\A(\O)=\A'(\O)$.
\end{definition}

\noindent Now we modify the definition of intervention. As mentioned, the important difference in the new notion of intervention is that now the agent has epistemic access to the  effects of the intervention on the observable variables in the actual world. This is stated by the last condition in Definition~\ref{def:intervention with ob}. Based on this update of our basic definitions we define a new semantics ``with observables'' for the language \LAfull in Definition~\ref{def: semantics with ob}.

\begin{definition}[Interventions on models with observables]\label{def:intervention with ob}
  Let $\ec$ be an epistemic causal model with observables $\langle\Sig, \F,\T\rangle$ for $\Sig=\langle \XV,\NV,\O,\Ran\rangle$; take $\A\in\T$. The epistemic causal model $\ec^\A_{\vec X = \vec x} = \langle\Sig,\F_{\vec X=\vec{x}},\T^{\F,\A}_{\vec X = \vec x}\rangle$, resulting from an intervention setting the values of variables in $\vec X$ to $\vec x$, is such that
\begin{itemize}
    \item $\F_{\vec X=\vec{x}}$ is as in Definition~\ref{def:int:epis}, and
    \item $\B\in \T^{\F,\A}_{\vec X = \vec x}$ iff $\B\in \T^{\F}_{\vec X = \vec x}$ and, for all $O\in \O$, $\B(O) = \A_{\vec X = \vec x}(O)$.
\end{itemize}
\end{definition}

\begin{definition}[Semantics with observables]\label{def: semantics with ob}
The formulas of \LAfull are evaluated over pairs $(\ec, \A)$, where $\ec = \tuple{\Sig, \F, \T}$ is an epistemic causal model with observables and $\A \in \T$. The satisfaction relation $(\ec, \A)\models\varphi$ is defined by the inductive clauses given in Definition \ref{def:LAk:semint}, with the following exception:
  \begin{cbtabular}{l@{\qquad{iff}\qquad}l}
    $(\ec, \A) \models \int{\opint{\vec{X}}{\vec{x}}}\gamma$ & $(\ec^\A_{\opint{\vec{X}}{\vec{x}}}, \newval{\A}{\F}{\opint{\vec{X}}{\vec{x}}}) \models \gamma$. \\
  \end{cbtabular}
\end{definition}






\noindent Wherever there is need to distinguish between this semantics with observables and the one in Section \ref{sec:proposal}, we will write $\models^O$ for the satisfaction relation in the semantics with observables and $\models^W$ for the earlier semantics without observables. We point out, however, that the semantics $\models^W$ can be seen a special case of $\models^O$: the case when the set $\O$ of observables is empty.

Let us illustrate how this extended system works by going back to the inference discussed in Section~\ref{sec:no_learning} in the context of Example~\ref{flashlight}: we will show that in the extended system pushing the button will tell the agent that the battery of the flashlight is empty. In other words, we show that $[P{=}1]K(B{=}0)$ holds in the epistemic causal model $\ev=\tuple{\Sig,\F,\T}$ extended with the set of observables $\{P, L\}$ (i.e. we assume that the agent can observe the button and the light, but not whether the batteries of the flashlight are charged).

$\tuple{\Sig,\F,\T},\A_1\models^O [P{=}1]K(B{=}0)$ is the case if for each assignment $\B$ in the resulting knowledge state after pushing the button ($\T^{\F,\A_1}_{P=1}$) the battery is empty ($B=0$). So, let's calculate the resulting knowledge state. We know that $\T = \{\A_1, \A_2\}$, where $\A_1$ is the possibility that the battery is empty, the button not pressed and the light is off, and $\A_2$ is the same possibility, except with a full battery. After the intervention we get $\T^\F_{P=1}=\{\A^{\F}_{1,P=1},\A^{\F}_{2,P=1}\}$ where in $\A^{\F}_{1,P=1}$ the light is still off (because the battery is empty), but in $\A^{\F}_{2,P=1}$ the light is on. Now, we have to check which of these possibilities matches what the agent actually observes. In this case the agent can see the button and the light, and in the actual world $\A_1$ pressing the button will not turn on the light. Then, the possibility $\A^{\F}_{2,P=0}$ will be removed from her knowledge state; only the possibility $\A^{F}_{1, P=0}$ remains. In this world the battery is empty. Thus after the experiment described by $[P=0]$ the agent knows that the battery is empty.

This example also immediately illustrates that this new semantics does not validate the rule \ax{NL} ({\it no learning}). We have just seen that $\tuple{\Sig,\F,\T},\A_1\models^O [P{=}1]K(B{=}0)$. However,  $\tuple{\Sig,\F,\T}\not \models^O K[P{=}1](B{=}0)$ (i.e. hypothetically considering the intervention of pushing the button will not make the agent believe that the battery is empty). For this sentence to be true in the given model we need that, in all the possibilities in the agent's knowledge state $\T$ after pressing the button, the battery is empty. However, this is not true for the possibility $\A_2 \in \T$. In this assignment the battery is charged, the button not pushed and the lamp is off. After pressing the button the battery will still be charged, i.e. $B=0$ will not hold. 
Thus, the left-to-right direction of the axiom \ax{CM} from the deduction system \LOfull is not valid in the semantics with observables.

However, the opposite direction, \ax{PR}, is still valid.
\begin{equation*}\label{perfectrecall}
   \ax{PR} \ \   K[\vec{X}=\vec{x}]\phi\rightarrow [\vec{X}=\vec{x}]K\phi.
\end{equation*}
\ax{PR} is an instance of the principle known as \emph{perfect recall} in the context of dynamic logics. In its general formulation, this principle asserts that if the outcome of an action (in our case, an intervention) is known before the action is performed, this piece of information is not forgotten after the action takes place. The reason why this principle still holds is that throughout this paper we have always assumed that all interventions are public knowledge. In other words, the value that intervened variables take after an intervention is always known by an agent.\footnote{We will come back to this assumption in Section~\ref{sec:conclusion}.} Therefore, any changes a hypothetical intervention brings to the epistemic state of the agent (left side of the implication) will also be there in case the intervention takes actually place (right side of the implication). As a consequence, whatever the agent can infer from a hypothetical intervention she will also infer from the actual intervention.  Hence, \ax{PR} is valid.  However, in the case of an actual intervention the agent can infer more. She might observe consequences of the action on variables that she can observe, which will add to the knowledge that she has. This additional information an agent can gain in case of an actual intervention renders the right-to-left side of the implication ({\it no learning}) invalid.


\begin{proposition}[Perfect recall]
  Let $\ec = \tuple{\Sig,\F,\T}$ be an epistemic causal model with observables, $\A\in\T$, and suppose $(\ec,\A)\models K[\vec X {=} \vec x]\psi$. Then $(\ec,\A)\models [\vec X {=} \vec x]K\psi$.
  \begin{proof}
    If $(\ec,\A)\models K[\vec X {=} \vec x]\psi$, then for all $\B\in \T$, $(\ec,\B)\models[\vec X {=}\vec x]\psi$.  In particular, this holds for all $\B\in \T$ such that $\B^\F_{\vec X =\vec x}\in \T^{\F,\A}_{\vec X {=}\vec x}$. So, for all such $\B$,  $(\ec^{\B}_{\vec X =\vec x},\B^\F_{\vec X {=}\vec x})\models \psi$. But now observe that, for all such $\B$,  $T^{\B}_{\vec X =\vec x} = \{\C^\F_{\vec X =\vec x} \mid \C\in \T \text { and } \C^\F_{\vec X =\vec x}(O)  = \B^\F_{\vec X =\vec x}(O) \text{ for all } O\in \mathcal O_T \} = \{ \C^\F_{\vec X =\vec x} \mid \C\in \T \text { and } \C^\F_{\vec X =\vec x}(O)= \A^\F_{\vec X =\vec x}(O)   \text{ for all } O\in \mathcal O_T \} = \T^{\F,\A}_{\vec X =\vec x}$. So $(\ec^\A_{\vec X =\vec x},\B_{\vec X =\vec x})\models \psi$ for all $\B_{\vec X =\vec x}\in \T^{\F,\A}_{\vec X =\vec x}$; thus $(\ec^\A_{\vec X =\vec x},\A_{\vec X =\vec x})\models K\psi$. So $(\ec,\A)\models[\vec X {=} \vec x]K\psi$.
\end{proof}
\end{proposition}




\myparagraph{Axiom system}
Our considerations on the failure of the ``no learning'' principle highlight the fact that the proof theory of \LAfull in the semantics with observables will differ from that in the case without. In particular, we cannot rely on the axiom \ax{CM}, which played an important role in the elimination of public announcements in \LOfull.

We then propose the following axiom system $\LOfullo$ (once more, parametrized by a signature $\Sig$). \LOfullo is defined from $\LOfull$ by removing the axiom \ax{CM} and adding the principles in Table \ref{tbl:LOfullo}.  As for \LOfull, each axiom of \LOfullo is taken also in the special case that the set of intervened variables $\vec X$ is empty.
\begin{table}[h!]
  \begin{center}
   \begin{footnotesize}
     \renewcommand{\arraystretch}{1.7}
     \begin{tabular}{l@{\quad}l}
       \toprule
       \ax{OC} & $[\vec X = \vec x] \bigvee_{\vec o\in \Ran(\O)}K\O=\vec o$, for each $O\in\O$ \\
       \ax{PD} & $[\vec X = \vec x]K\psi \leftrightarrow \bigvee_{\vec o\in \Ran(\O)}\Big( [\vec X = \vec x] \mathcal O = \vec o \land \big[[\vec X = \vec x] \mathcal O = \vec o ! \big]K[\vec X = \vec x]\psi\Big)$ \\
       \bottomrule
     \end{tabular}
    \end{footnotesize}
   \end{center}
   \caption{Additional rules for \LOfullo.}
   \label{tbl:LOfullo}
\end{table}







The axiom \ax{OC} expresses the fact that the observables always take a constant value in the set of valuations, i.e. their value is always known to the experimenter - before and after the experiment. 
Axiom \ax{PD} (\emph{prediction axiom}) describes how the knowledge of the agent after an intervention is characterised by the knowledge she may have before any intervention occurs, just as axiom \ax{CM} does in the case without observables.
However, while in the case without observables the knowledge the agent needs to have prior to the intervention is full knowledge of the outcome (\emph{no learning}), now she may also make use of her knowledge of the effect that the intervention would have on the observables. Indeed, \ax{PD} essentially states that after an intervention the agent knows $\psi$ if and only if, \emph{after learning how  the values of the observables are affected by the intervention}, she knows that the intervention makes $\psi$ true. By characterizing knowledge after an intervention in terms of knowledge before the intervention, axiom \ax{PD} plays a crucial role in the procedure of elimination of the announcement operators, by  allowing us to extract occurrences of $K$ from the scope of an intervention (see Proposition \ref{prop_reduction} in the Appendix). A similar role was played by axiom \ax{CM} in the completeness proof for \LOfull.





\begin{theorem}[Axiom system for \LAfull over models w/\! observables]\label{thm: soundness completeness LOfullo}
  Let $\Sig = \langle \XV, \NV$, $\O,\Ran \rangle$ be a finite signature with observables. The axiom system \LOfullo
  is sound and strongly complete for the language \LAfull based on $\Sig$ with respect to epistemic (recursive) causal models with observables (and of signature $\Sig$).
  \begin{proof}
    See Appendix \ref{app:LOfullo}.
  \end{proof}
\end{theorem}

\noindent One may wonder what kind of relationship subsists between the two interpretations of the language \LAfull (i.e., with and without observables). 
It can be proved that, as long as only finite signatures are considered - or more generally, only finite sets of observables -  \LAfull, as interpreted with observables, can be embedded into its version without observables. So, in principle reasoning about real experiments could also be conducted within the semantics without observables; but this comes at the cost of an increase in the size and complexity of formulas. The involved details of this argument will be presented in a forthcoming manuscript.

\section{Conclusions}\label{sec:conclusion}

This paper has given some steps towards the integration of causal and epistemic reasoning, providing an adequate semantics, a language combining interventionist counterfactuals with (dynamic) epistemic operators and a sound and complete system of inference. Our deductive system models the thought of an agent reasoning about the consequences of hypothetical and real experiments. It describes what an agent may deduce from her/his \emph{a priori} pool of knowledge when considering a hypothetical intervention, but also what she could learn from performing this experiment in the actual world. For this later part the concept of observables was crucial. Observables allow us to model measuring the value of variables when performing experiment. Through the addition of observables to our logic we are able to model actual learning from experiments. In this respect the system proposed here goes substantially beyond \cite{Dali_Proceedings}.

Still, the framework developed here is just a first step towards providing a logic of reasoning about experiments. There are aspects of reasoning about experiments that our proposal cannot yet account for. Some of these limitations are due to assumptions made by the models of epistemic and causal reasoning that we combined. For instance, the particular model of causal information we have worked with, structural causal models, assume that the value of a variable is fully determined by the value of its causal parents. But most causal relationships science deals with are of a probabilistic/statistical nature: if the values of the causal parents are such and such, then there is a probability $p$ that the value of $X$ is $x$. This could be modelled better using causal Bayes nets as a starting point. In such case, one might want to use a probabilistic setting for modelling \emph{uncertainty} (see, e.g., \citealp{Halpern2017}) and a matching extension for dealing with change in uncertainty (e.g., the already mentioned \citealp{Kooi03,baltag2008probabilistic,MartinyM15}).

There are also some assumptions about the interaction between epistemic and causal reasoning that could be reconsidered in future work. One of them, for instance, concerns the way we implemented interventions. The logic proposed here treats an intervention as public knowledge: an epistemic state is automatically updated with any intervention that takes place -- all experiments are observed. Whether the agent can also see the effects of the intervention depends on which variables she can observe. But she will certainly be aware of the change in the value of variable(s) that are in the scope of the intervention. We have chosen this implementation of intervention because we wanted to push here the perspective of intervention as just another action in dynamic logic. For this reason we wanted to make it as similar as possible to the operation of public announcement. Furthermore, in the single agent setting it seems natural to assume that this agent is reasoning about experiments she is performing. However, in a multi-agent setting, this assumption can be given up.

There is one assumption we have made here that deserves particular attention. As explained in the introduction, the goal of the formal work reported in this paper is to come up with a qualitative logic of reasoning about experiments. One important function of experiments is to provide us information about causal relationships. However, this cannot be modelled by the system proposed here. This system takes the causal dependencies to be given to the agent. The only information an agent can gain from doing experiments is information about the actual value of variables. One way to solve this problem is to introduce uncertainty about the right causal network, for instance (as in \citealp{BarYan2020}) by letting in the epistemic state not only the assignment, but also the causal model vary. Making this change is not hard, but it raises the interesting and difficult question what kind of regularities we want to implement concerning learning causal relationships. This is something we want to study in future work, together with the question how such a framework would relate to the algorithms for learning causal networks that are used in various scientific applications, such as \cite{Spirtes, Peters}.

A third assumption concerns the structure of the causal laws. As is standard in the literature, we assumed here that causal structures are not circular. Models without this restriction are mostly used to represent complex feedback systems rather than everyday scenarios such as we considered in this paper; and their general formal treatment would lead us away from the safe methods of modal logic, since on acyclic structures interventions may not guarantee the existence and uniqueness of an actual world. There is however a ``safe'' class of causal models, not necessarily acyclic, on which interventions uniquely identify one actual world: the so-called \emph{unique-solution} systems \citep{GallesPearl1997}. It might be interesting to extend our work in this direction.

As mentioned before, our framework has many points in common with causal team semantics. In \cite{Dali_Proceedings} we provide a translation between the two approaches. Our semantics has the advantage of encoding explicitly the actual state of the world (in particular, the actual value of variables).  For our purposes having access to the actual value of variables was crucial, because it enabled us to account for learning from experiments. Actual values play also an important role in many definitions of \emph{token causation}, i.e. causation between events \citep{Hit2001b,Woo2003,halpern2016actual}. Our framework might help with formalising these definitions and comparing them to each other. In future work we plan to consider richer languages with hybrid features that will make even more use of the available information about the actual world. For example, one could distinguish in the language the actual action (intervention) from observing the action
; and operators for \emph{fixing the actual value} of a variable can help to spell out definitions of token causation. 

Another family of logics that our proposal is related to are the logics of dependence and independence. As demonstrated in \citep{Bal2016,EijGatYan2017}, the notion of functional dependence can be reduced in terms of knowledge operators. This raised the question whether other forms of correlational dependence may be described in terms of epistemic operators. The literature on team semantics considered a large variety of such (in)dependence notions, among which the most studied are the \emph{independence atoms} \citep{GraVaa2013} and \emph{inclusion atoms} \citep{Gal2012}. 
More generally, in the kinds of frameworks developed in the present paper, we can investigate how a notion of causal dependence can be decomposed into epistemic and purely causal aspects. We believe this might contribute to the everlasting problem of separating clearly the epistemology and the ontology of causation.

Finally, in future work we plan to extend the setting to a multi-agent system. This involves considering not only different agents with potentially different knowledge and different variables they can observe, but also epistemic attitudes for groups (e.g., distributed and common knowledge) and the effect of inter-agent communication. One advantage this will bring is the potential to contribute to the discussion about causal agency and the role of causation in the study of responsibility within AI (see, for instance, \citealp{paper_with_Ilaria}).

\appendix

\section{Appendix}

\subsection{Proof of Theorem \ref{thm:rcm-LA}}\label{app:rcm-LA}

\begin{appendix:full}

\myparagraph{Soundness.} The soundness of \ax{P} and \ax{MP} is straightforward; that of \ax{A_1-A_5} and \ax{A_\land} has been proved in  \citet{halpern2000axiomatizing}. Axioms \ax{A_6} and \ax{A_\lnot} are sound because we work on \emph{recursive} causal models. For the first, a recursive set of structural functions $\F$ produces a relation $\dirparof{\F}$ without cycles (Footnote~\ref{ftn:recursive}), which is syntactically characterised by $\syndirparof$. For the second observe that, in a recursive causal model, the value of each variable is uniquely determined.

For \ax{A_7} note how, for any assignment $\vec{X}{=}\vec{x}$, the valuations $\A$ and $\newval{\A}{\F}{\opint{\vec{X}}{\vec{x}}}$ coincide in the value of \emph{exogenous} variables not occurring in $\vec{X}$ (Definition \ref{def:int}). For \ax{A_{[]}}, an intervention with the empty assignment does not affect the given causal model. Finally, \ax{A_{[][]}} states that, when two interventions are performed in a row, the second overrides the first in the variables they both act upon.

\myparagraph{Completeness.} For completeness, it will be shown that axioms \ax{A_{[]}}, \ax{A_\lnot}, \ax{A_\land} and \ax{A_{[][]}} define a translation from \LAc to a language \LAcres for which axioms \ax{A_1}-\ax{A_7}, \ax{P} and rule \ax{MP} are complete. Here are the details.

\myparagraphit{Completeness, step 1: from \LAc to \LAcres.} The first part deals with a translation to the following language, consisting of Boolean combinations of counterfactuals with an atomic consequent.

\begin{definition}[Language \LAcres \citep{halpern2000axiomatizing}]
  Formulas $\gamma$ of the language \LAcres based on the signature $\Sig$ are given by
  \[ \gamma ::= \int{\opint{\vec{X}}{\vec{x}}}{Y{=}y} \mid \lnot \gamma \mid \gamma \land \gamma \]
  for $Y \in \AV$, $y \in \Ran(Y)$ and $\opint{\vec{X}}{\vec{x}}$ an assignment on $\Sig$. Given a causal model $\tuple{\Sig, \F, \A}$, formulas are interpreted as before.
\end{definition}

For terminology, let expressions of the form $Y{=}y$ be called \LAc-atoms, and expressions of them form $\int{\opint{\vec{X}}{\vec{x}}}{Y{=}y}$ be called \LAcres-atoms. Note how formulas in \LAcres are simply Boolean combinations of \LAcres-atoms.

{\smallskip}

Here is the translation.

\begin{definition}[Translation $\tro$]\label{def:tro}
  Define $\tro:\LAc \to \LAcres$ as
  \begingroup
    \small
    \[
      \renewcommand{\arraystretch}{1.6}
      \begin{array}{@{}c@{\quad}c@{}}
        \begin{array}{@{}r@{\,:=\,}l@{}}
          \tro(Y{=}y)               & \int{\;}{Y{=}y} \\
          \tro(\lnot\varphi)           & \lnot \tro(\varphi) \\
          \tro(\varphi_1 \land \varphi_2) & \tro(\varphi_1) \land \tro(\varphi_2) \\
          \multicolumn{2}{@{}r}{}
        \end{array}
        &
        \begin{array}{@{}r@{\,:=\,}l@{}}
          \tro(\int{\opint{\vec{X}}{\vec{x}}}{Y{=}y})                                & \int{\opint{\vec{X}}{\vec{x}}}{Y{=}y} \\
          \tro(\int{\opint{\vec{X}}{\vec{x}}}{\lnot \varphi})                           & \tro(\lnot \int{\opint{\vec{X}}{\vec{x}}}{\varphi}) \\
          \tro(\int{\opint{\vec{X}}{\vec{x}}}{(\varphi_1 \land \varphi_2)})                & \tro(\int{\opint{\vec{X}}{\vec{x}}}{\varphi_1} \land \int{\opint{\vec{X}}{\vec{x}}}{\varphi_2}) \\
          \tro(\int{\opint{\vec{X}}{\vec{x}}}{\int{\opint{\vec{Y}}{\vec{y}}}{\varphi}}) & \tro(\int{\opint{\vec{X'}}{\vec{x'}}, \opint{\vec{Y}}{\vec{y}}}{\varphi}) \\
        \end{array}
      \end{array}
    \]
  \endgroup
  where, in the bottom clause on the right, $\opint{\vec{X'}}{\vec{x'}}$ is the subassignment of $\opint{\vec{X}}{\vec{x}}$ for $\vec{X'} := \vec{X} \setminus \vec{Y}$.\footnote{Note how the translation can be `shortened' by directly defining $\tro(\int{\opint{\vec{X}}{\vec{x}}}{\lnot \varphi}) := \lnot \tro(\int{\opint{\vec{X}}{\vec{x}}}{\varphi})$ and $\tro(\int{\opint{\vec{X}}{\vec{x}}}{(\varphi_1 \land \varphi_2)}) := \tro(\int{\opint{\vec{X}}{\vec{x}}}{\varphi_1}) \land \tro(\int{\opint{\vec{X}}{\vec{x}}}{\varphi_2})$. Still, the provided clauses have the precise shape that is needed to prove $\tro$'s crucial properties.}
\end{definition}

Intuitively, $\tro$ performs two tasks: it turns \LAc-atoms into \LAcres-atoms, and it guarantees that only \LAc-atoms occur inside the scope of intervention operators. The first task is taken care of by case $\tro(Y{=}y)$, which adds an intervention with the empty assignment to every \LAc-atom. The second task is taken care of by the cases in the rightmost column: cases $\tro(\int{\opint{\vec{X}}{\vec{x}}}{\lnot \varphi})$ and $\tro(\int{\opint{\vec{X}}{\vec{x}}}{(\varphi_1 \land \varphi_2)}) $ push the intervention operator $\int{\opint{\vec{X}}{\vec{x}}}$ deeper into the formula on its right by commuting over $\lnot$ and distributing over $\land$. A repetitive application of those cases leads to formulas in which the expression directly in front of $\int{\opint{\vec{X}}{\vec{x}}}$ is either $Y{=}y$ or another intervention $\int{\opint{\vec{Y}}{\vec{y}}}{\varphi}$. In the former, the formula is in the desired shape, as the case $\tro(\int{\opint{\vec{X}}{\vec{x}}}{Y{=}y})$ acknowledges. In the latter, case $\tro(\int{\opint{\vec{X}}{\vec{x}}}{\int{\opint{\vec{Y}}{\vec{y}}}{\varphi}})$ turns the two sequential interventions into a single one so the process can continue, now pushing the new intervention deeper into $\varphi$.

{\smallskip}

The following proposition contains the crucial properties of $\tro$.

\begin{proposition}\label{pro:LAtoLAres}
  For every $\varphi \in \LAc$,
  \begin{multicols}{3}
    \begin{enumerate}
      \item\label{pro:LAtoLAres:trans} $\tro(\varphi) \in \LAcres$,
      \item\label{pro:LAtoLAres:syn} $\vdashloc \varphi \leftrightarrow \tro(\varphi)$,
      \item\label{pro:LAtoLAres:sem} $\models \varphi \leftrightarrow \tro(\varphi)$.
    \end{enumerate}
  \end{multicols}
  \begin{proof}
    A standard strategy for proving that all formulas in a given language satisfy certain property is to work by induction on the formulas' structure, thus relying on an inductive hypothesis (IH) stating that every \emph{subformula} of the given $\varphi$ has the property. However, this form of induction is of not use here. For example, when arguing that $\int{\opint{\vec{X}}{\vec{x}}}{\lnot \varphi}$ has certain property, one would like to rely on properties of $\lnot \int{\opint{\vec{X}}{\vec{x}}}{\varphi}$, but the later is not a subformula of the former.

    {\smallskip}
    The proofs of this proposition use induction, but the induction will be not on the structure of the formulas, but rather on their \emph{complexity} $\com:\LAc \to \Nat\setminus\set{0}$, defined as
    \begin{ctabular}{c@{\qquad}c}
      \begin{tabular}{r@{\;:=\;}l}
        $\com(Y{=}y)$      & $1$ \\
        $\com(\lnot \varphi)$ & $1 + \com(\varphi)$
      \end{tabular}
      &
      \begin{tabular}{r@{\;:=\;}l}
        $\com(\varphi_1 \land \varphi_2)$                  & $1 + \max\set{\com(\varphi_1), \com(\varphi_2)}$ \\
        $\com(\int{\opint{\vec{X}}{\vec{x}}}{\varphi})$ & $2\com(\varphi)$
      \end{tabular}
    \end{ctabular}
    The important feature of $\com$ is the following: for all assignments $\opint{\vec{X}}{\vec{x}}$ and $\opint{\vec{Y}}{\vec{y}}$, and all formulas $\varphi, \varphi_1, \varphi_2 \in \LAc$,
    \begin{compactitemize}
      \item $\com(\varphi) \geqslant \com(\psi)$ for every $\psi \in \sub(\varphi)$, with $\sub$ the standard subformula function. This is shown by structural induction.%
\footnote{\textbf{Base case ($\bs{Y{=}y}$)}. Note that $\com(Y{=}y)=1$. Now take any $\psi$ in $\sub(Y{=}y) = \set{Y{=}y}$; clearly, $\com(Y{=}y) = \com(\psi)$. \textbf{Inductive case $\bs{\lnot \varphi}$}, with the IH being $\com(\varphi) \geqslant \com(\psi)$ for every $\psi \in \sub(\varphi)$. Note that $\com(\lnot \varphi)=1+\com(\varphi)$ so $\com(\lnot \varphi) > \com(\varphi)$. Now take any $\psi$ in $\sub(\lnot \varphi) = \set{\lnot \varphi} \cup \sub(\varphi)$: if $\psi = \lnot \varphi$ then $\com(\lnot \varphi) = \com(\psi)$, and if $\psi \in \sub(\varphi)$ then $\com(\varphi) \geqslant \com(\psi)$ (by IH) and hence $\com(\lnot \varphi) > \com(\psi)$. \textbf{Inductive case $\bs{\varphi_1 \land \varphi_2}$}, with the IH being $\com(\varphi_i) \geqslant \com(\psi)$ for every $\psi \in \sub(\varphi_i)$ and $i \in \set{1,2}$. Note that $\com(\varphi_1 \land \varphi_2) = 1 + \max\set{\com(\varphi_1), \com(\varphi_2)}$ so $\com(\varphi_1 \land \varphi_2) > \com(\varphi_i)$ for $i \in \set{1,2}$. Now take any $\psi$ in $\sub(\varphi_1 \land \varphi_2) = \set{\varphi_1 \land \varphi_2} \cup \sub(\varphi_1) \cup \sub(\varphi_2)$: if $\psi = \varphi_1 \land \varphi_2$ then $\com(\varphi_1 \land \varphi_2) = \com(\psi)$, and if $\psi \in \sub(\varphi_i)$ then $\com(\varphi_i) \geqslant \com(\psi)$ (by IH) and hence $\com(\varphi_1 \land \varphi_2) > \com(\psi)$. \textbf{Inductive case $\bs{\int{\opint{\vec{X}}{\vec{x}}}{\varphi}}$}, with the IH being $\com(\varphi) \geqslant \com(\psi)$ for every $\psi \in \sub(\varphi)$. Note that $\com(\int{\opint{\vec{X}}{\vec{x}}}{\varphi})=2\com(\varphi)$ so $\com(\int{\opint{\vec{X}}{\vec{x}}}{\varphi}) > \com(\varphi)$. Now take any $\psi$ in $\sub(\int{\opint{\vec{X}}{\vec{x}}}{\varphi}) = \set{\int{\opint{\vec{X}}{\vec{x}}}{\varphi}} \cup \sub(\varphi)$: if $\psi = \int{\opint{\vec{X}}{\vec{x}}}{\varphi}$ then $\com(\int{\opint{\vec{X}}{\vec{x}}}{\varphi}) = \com(\psi)$, and if $\psi \in \sub(\varphi)$ then $\com(\varphi) \geqslant \com(\psi)$ (by IH) and hence $\com(\int{\opint{\vec{X}}{\vec{x}}}{\varphi}) > \com(\psi)$.}

      \item $\com(\int{\opint{\vec{X}}{\vec{x}}}{\lnot \varphi}) > \com(\lnot \int{\opint{\vec{X}}{\vec{x}}}{\varphi})$.
      Indeed,
      \begin{smallltabular}{@{--\;\;}l}
        $\com(\int{\opint{\vec{X}}{\vec{x}}}{\lnot \varphi}) = 2\com(\lnot \varphi) = 2(1+\com(\varphi)) = 2+2\com(\varphi)$; \\
        $\com(\lnot \int{\opint{\vec{X}}{\vec{x}}}{\varphi}) = 1+\com(\int{\opint{\vec{X}}{\vec{x}}}{\varphi}) = 1+2\com(\varphi)$.
      \end{smallltabular}

      \item $\com(\int{\opint{\vec{X}}{\vec{x}}}{(\varphi_1 \land \varphi_2)}) > \com(\int{\opint{\vec{X}}{\vec{x}}}{\varphi_1} \land \int{\opint{\vec{X}}{\vec{x}}}{\varphi_2})$.
      Indeed,
      \begin{smallltabular}{@{--\;\;}l@{\;=\;}l}
        $\com(\int{\opint{\vec{X}}{\vec{x}}}{(\varphi_1 \land \varphi_2)}) = 2\com(\varphi_1 \land \varphi_2)$ & $2(1 + \max\set{\com(\varphi_1), \com(\varphi_2)})$ \\
        \multicolumn{1}{l@{\;=\;}}{}                                                                             & $2 + 2\max\set{\com(\varphi_1), \com(\varphi_2)}$; \\
        $\com(\int{\opint{\vec{X}}{\vec{x}}}{\varphi_1} \land \int{\opint{\vec{X}}{\vec{x}}}{\varphi_2})$ & $1 + \max\set{\com(\int{\opint{\vec{X}}{\vec{x}}}{\varphi_1}), \com(\int{\opint{\vec{X}}{\vec{x}}}{\varphi_2})}$ \\
        \multicolumn{1}{l@{\;=\;}}{}                                                                          & $1 + \max\set{2\com(\varphi_1), 2\com(\varphi_2)}$ \\
        \multicolumn{1}{l@{\;=\;}}{}                                                                          & $1 + 2\max\set{\com(\varphi_1), \com(\varphi_2)}$.
      \end{smallltabular}

      \item $\com(\int{\opint{\vec{X}}{\vec{x}}}{\int{\opint{\vec{Y}}{\vec{y}}}{\varphi}}) > \com(\int{\opint{\vec{X'}}{\vec{x'}}, \opint{\vec{Y}}{\vec{y}}}{\varphi})$.
      Indeed,
      \begin{smallltabular}{@{--\;\;}l}
        $\com(\int{\opint{\vec{X}}{\vec{x}}}{\int{\opint{\vec{Y}}{\vec{y}}}{\varphi}}) = 2\com(\int{\opint{\vec{Y}}{\vec{y}}}{\varphi}) = 2(2)\com(\varphi) = 4\com(\varphi)$; \\
        $\com(\int{\opint{\vec{X'}}{\vec{x'}}, \opint{\vec{Y}}{\vec{y}}}{\varphi}) = 2\com(\varphi)$.
      \end{smallltabular}
    \end{compactitemize}
    Thus, in every case of the definition of the translation, the complexity of the formulas under $\tro$ on the right-hand side is strictly smaller than the complexity of the formula under $\tro$ on the left-hand side. Because of this, the application of $\tro$ to any formula $\varphi \in \LAc$ will eventually end.

    {\medskip}

    With this tool, here are the proofs.
    \begin{compactenumerate}
      \item The proof is by induction on $\com(\varphi)$. The base case is for formulas $\varphi$ with $\com(\varphi) = 1$; the inductive case is for formulas $\varphi$ with $\com(\varphi) > 1$, with the IH stating that $\tro(\psi) \in \LAcres$ for all formulas $\psi \in \LAc$ with $\com(\psi) < \com(\varphi)$.
      \begin{compactitemize}
        \item \textbf{Case $\bs{\com(\varphi)=1}$.} The only formulas in \LAc with $\com(\varphi)=1$ are \LAc-atoms, so it should be proved that $\tro(Y{=}y) \in \LAcres$. This is straightforward, as $\tro(Y{=}y) = \int{\;}{Y{=}y}$ is an \LAcres-atom.

        \item \textbf{Case $\bs{\com(\varphi)>1}$.} Here are the cases.
        \begin{compactitemize}
          \item \textbf{Case $\bs{\lnot \varphi}$.} Since $\com(\lnot \varphi) > \com(\varphi)$, from IH it follows that $\tro(\varphi)$ is in \LAcres, and hence so is $\lnot \tro(\varphi) = \tro(\lnot \varphi)$.

          \item \textbf{Case $\bs{\varphi_1 \land \varphi_2}$.} Since $\com(\varphi_1 \land \varphi_2) > \com(\varphi_i)$ for $i \in \set{1,2}$, from IH it follows that both $\tro(\varphi_1)$ and $\tro(\varphi_2)$ are in \LAcres, and thus so is $\tro(\varphi_1) \land \tro(\varphi_2) = \tro(\varphi_1 \land \varphi_2)$.

          \item \textbf{Case $\bs{\int{\opint{\vec{X}}{\vec{x}}}{Y{=}y}}$.} From the definition of $\tro(\int{\opint{\vec{X}}{\vec{x}}}{Y{=}y})$, the goal $\tro(\int{\opint{\vec{X}}{\vec{x}}}{Y{=}y}) \in \LAcres$ is straightforward.

          \item \textbf{Case $\bs{\int{\opint{\vec{X}}{\vec{x}}}{\lnot \varphi}}$.} Since $\com(\int{\opint{\vec{X}}{\vec{x}}}{\lnot \varphi}) > \com(\lnot \int{\opint{\vec{X}}{\vec{x}}}{\varphi})$, from IH it follows that $\tro(\lnot \int{\opint{\vec{X}}{\vec{x}}}{\varphi}) \in \LAcres$. By definition, the latter formula is the same as $\tro(\int{\opint{\vec{X}}{\vec{x}}}{\lnot \varphi})$, which completes the case.

          \item \textbf{Case $\bs{\int{\opint{\vec{X}}{\vec{x}}}{(\varphi_1 \land \varphi_2)}}$.} Since $\com(\int{\opint{\vec{X}}{\vec{x}}}{(\varphi_1 \land \varphi_2)}) > \com(\int{\opint{\vec{X}}{\vec{x}}}{\varphi_1} \land \int{\opint{\vec{X}}{\vec{x}}}{\varphi_2})$, from IH it follows that $\tro(\int{\opint{\vec{X}}{\vec{x}}}{\varphi_1} \land \int{\opint{\vec{X}}{\vec{x}}}{\varphi_2}) \in \LAcres$. By definition, the latter formula is the same as $\tro(\int{\opint{\vec{X}}{\vec{x}}}{(\varphi_1 \land \varphi_2)})$, which completes the case.

          \item \textbf{Case $\bs{\int{\opint{\vec{X}}{\vec{x}}}{\int{\opint{\vec{Y}}{\vec{y}}}{\varphi}}}$.} Since $\com(\int{\opint{\vec{X}}{\vec{x}}}{\int{\opint{\vec{Y}}{\vec{y}}}{\varphi}}) > \com(\int{\opint{\vec{X'}}{\vec{x'}}, \opint{\vec{Y}}{\vec{y}}}{\varphi})$, from IH it follows that $\tro(\int{\opint{\vec{X'}}{\vec{x'}}, \opint{\vec{Y}}{\vec{y}}}{\varphi}) \in \LAcres$. By definition, the latter formula is $\tro(\int{\opint{\vec{X}}{\vec{x}}}{\int{\opint{\vec{Y}}{\vec{y}}}{\varphi}})$, which completes the case.
        \end{compactitemize}
      \end{compactitemize}

      \item  The proof is by induction on $\com(\varphi)$. The base case is for formulas $\varphi$ with $\com(\varphi) = 1$; the inductive case is for formulas $\varphi$ with $\com(\varphi) > 1$, with the IH stating that $\vdashloc \psi \leftrightarrow \tro(\psi)$ holds for all formulas $\psi \in \LAc$ with $\com(\psi) < \com(\varphi)$.
      \begin{compactitemize}
        \item \textbf{Case $\bs{\com(\varphi)=1}$.} The only formulas in \LAc with $\com(\varphi)=1$ are \LAc-atoms, so it should be proved that $\vdashloc Y{=}y \leftrightarrow \tro(Y{=}y)$, i.e., that $\vdashloc Y{=}y \leftrightarrow \int{\;}{Y{=}y}$. This follows immediately from axiom \ax{A_{[]}}.

        \item \textbf{Case $\bs{\com(\varphi)>1}$.} Here are the cases.
        \begin{compactitemize}
          \item \textbf{Case $\bs{\lnot \varphi}$.} Since $\com(\lnot \varphi) > \com(\varphi)$, from IH it follows that $\vdashloc \varphi \leftrightarrow \tro(\varphi)$. Then $\vdashloc \lnot \varphi \leftrightarrow \lnot \tro(\varphi)$ (propositional reasoning), and thus the definition of $\tro$ yields the required $\vdashloc \lnot \varphi \leftrightarrow \tro(\lnot \varphi)$.

          \item \textbf{Case $\bs{\varphi_1 \land \varphi_2}$.} Since $\com(\varphi_1 \land \varphi_2) > \com(\varphi_i)$ for $i \in \set{1,2}$, from IH it follows that $\vdashloc \varphi_i \leftrightarrow \tro(\varphi_i)$. Then $\vdashloc (\varphi_1 \land \varphi_2) \leftrightarrow (\tro(\varphi_1) \land \tro(\varphi_2))$ (propositional reasoning), and thus the definition of $\tro$ yields the required $\vdashloc (\varphi_1 \land \varphi_2) \leftrightarrow \tro(\varphi_1 \land \varphi_2)$.

          \item \textbf{Case $\bs{\int{\opint{\vec{X}}{\vec{x}}}{Y{=}y}}$.} From the definition of $\tro(\int{\opint{\vec{X}}{\vec{x}}}{Y{=}y})$, the goal $\vdashloc \int{\opint{\vec{X}}{\vec{x}}}{Y{=}y} \leftrightarrow \tro(\int{\opint{\vec{X}}{\vec{x}}}{Y{=}y})$ is straightforward.

          \item \textbf{Case $\bs{\int{\opint{\vec{X}}{\vec{x}}}{\lnot \varphi}}$.} Since $\com(\int{\opint{\vec{X}}{\vec{x}}}{\lnot \varphi}) > \com(\lnot \int{\opint{\vec{X}}{\vec{x}}}{\varphi})$, from IH it follows that $\vdashloc \lnot \int{\opint{\vec{X}}{\vec{x}}}{\varphi} \leftrightarrow \tro(\lnot \int{\opint{\vec{X}}{\vec{x}}}{\varphi})$. But axiom \ax{A_\lnot} gives us $\vdashloc \int{\opint{\vec{X}}{\vec{x}}}{\lnot \varphi} \leftrightarrow \lnot \int{\opint{\vec{X}}{\vec{x}}}{\varphi}$, so $\vdashloc \int{\opint{\vec{X}}{\vec{x}}}{\lnot \varphi} \leftrightarrow \tro(\lnot \int{\opint{\vec{X}}{\vec{x}}}{\varphi})$ and hence, from $\tro$'s definition, $\vdashloc \int{\opint{\vec{X}}{\vec{x}}}{\lnot \varphi} \leftrightarrow \tro(\int{\opint{\vec{X}}{\vec{x}}}{\lnot \varphi})$.

          \item \textbf{Case $\bs{\int{\opint{\vec{X}}{\vec{x}}}{(\varphi_1 \land \varphi_2)}}$.} Since $\com(\int{\opint{\vec{X}}{\vec{x}}}{(\varphi_1 \land \varphi_2)}) > \com(\int{\opint{\vec{X}}{\vec{x}}}{\varphi_1} \land \int{\opint{\vec{X}}{\vec{x}}}{\varphi_2})$, IH gives us $\vdashloc (\int{\opint{\vec{X}}{\vec{x}}}{\varphi_1} \land \int{\opint{\vec{X}}{\vec{x}}}{\varphi_2}) \leftrightarrow \tro(\int{\opint{\vec{X}}{\vec{x}}}{\varphi_1} \land \int{\opint{\vec{X}}{\vec{x}}}{\varphi_2})$. But, by axiom \ax{A_\land}, $\vdashloc \int{\opint{\vec{X}}{\vec{x}}}{(\varphi_1 \land \varphi_2)} \leftrightarrow (\int{\opint{\vec{X}}{\vec{x}}}{\varphi_1} \land \int{\opint{\vec{X}}{\vec{x}}}{\varphi_2})$; then, it follows that $\vdashloc \int{\opint{\vec{X}}{\vec{x}}}{(\varphi_1 \land \varphi_2)} \leftrightarrow \tro(\int{\opint{\vec{X}}{\vec{x}}}{\varphi_1} \land \int{\opint{\vec{X}}{\vec{x}}}{\varphi_2})$ and hence, from $\tro$'s definition we obtain the required $\vdashloc \int{\opint{\vec{X}}{\vec{x}}}{(\varphi_1 \land \varphi_2)} \leftrightarrow \tro(\int{\opint{\vec{X}}{\vec{x}}}{(\varphi_1 \land \varphi_2)})$.

          \item \textbf{Case $\bs{\int{\opint{\vec{X}}{\vec{x}}}{\int{\opint{\vec{Y}}{\vec{y}}}{\varphi}}}$.} Since $\com(\int{\opint{\vec{X}}{\vec{x}}}{\int{\opint{\vec{Y}}{\vec{y}}}{\varphi}}) > \com(\int{\opint{\vec{X'}}{\vec{x'}}, \opint{\vec{Y}}{\vec{y}}}{\varphi})$, from IH it follows that $\vdashloc \int{\opint{\vec{X'}}{\vec{x'}}, \opint{\vec{Y}}{\vec{y}}}{\varphi} \leftrightarrow \tro(\int{\opint{\vec{X'}}{\vec{x'}}, \opint{\vec{Y}}{\vec{y}}}{\varphi}))$. But $\vdashloc \int{\opint{\vec{X}}{\vec{x}}}{\int{\opint{\vec{Y}}{\vec{y}}}{\varphi}} \leftrightarrow \int{\opint{\vec{X'}}{\vec{x'}}, \opint{\vec{Y}}{\vec{y}}}{\varphi}$ (from axiom \ax{A_{[][]}}), so $\vdashloc \int{\opint{\vec{X}}{\vec{x}}}{\int{\opint{\vec{Y}}{\vec{y}}}{\varphi}} \leftrightarrow \tro(\int{\opint{\vec{X'}}{\vec{x'}}, \opint{\vec{Y}}{\vec{y}}}{\varphi})$; hence, the definition of $\tro$ yields the required $\vdashloc \int{\opint{\vec{X}}{\vec{x}}}{\int{\opint{\vec{Y}}{\vec{y}}}{\varphi}} \leftrightarrow \tro(\int{\opint{\vec{X}}{\vec{x}}}{\int{\opint{\vec{Y}}{\vec{y}}}{\varphi}})$.
        \end{compactitemize}
      \end{compactitemize}

      \item By the previous item, $\vdashloc \varphi \leftrightarrow \tro(\varphi)$. But \LOc is sound within recursive causal models; therefore, $\models \varphi \leftrightarrow \tro(\varphi)$.
    \end{compactenumerate}
  \end{proof}
\end{proposition}

\myparagraphit{Completeness, step 2: a canonical model for \LAcres.} Having shown that every formula in \LAc is both semantically and \LOc-provably equivalent to a formula in \LAcres, now it will be shown that the axiom system \LOcres, the fragment of \LOc without axioms \ax{A_{[]}}, \ax{A_\lnot}, \ax{A_\land} and \ax{A_{[][]}}, is complete for \LAcres over recursive causal models.  This will be done by showing, via the construction of a canonical model, that any \LOcres-consistent set of \LAcres-formulas is satisfiable in a recursive causal model. The construction is almost exactly as that in \citet{halpern2000axiomatizing}, with just minor adjustments.

{\smallskip}

Fix the signature $\Sig = \tuple{\XV, \NV, \Ran}$ (Definition \ref{def:signature}). Let $\mcss$ be the set of all maximally \LOcres-consistent sets of \LAcres-formulas. The process starts by showing how each $\Gamma \in \mcss$ gives raise to a recursive causal model.

\begin{definition}[Building a causal model]\label{def:cm}
  Let $\Gamma \in \mcss$ be a maximally \LOcres-con\-sis\-tent set of \LAcres-formulas.
  \begin{compactitemize}
    \item Let $\vec{U}$ be a tuple with all exogenous variables in $\XV$. For each endogenous variable $V \in \NV$, let $\vec{Y}$ be the tuple of all endogenous variables in $\NV \setminus \set{V}$. The structural function $\fun^{\Gamma}_V$ is defined, for each $\vec{u} \in \Ran(\vec{U})$ and $\vec{y} \in \Ran(\vec{Y})$, as
    \[
      \fun^{\Gamma}_V(\vec{u}, \vec{y}) = v
      \quad\text{if and only if}\quad
      \int{\opint{\vec{U}}{\vec{u}},\opint{\vec{Y}}{\vec{y}}}{V{=}v} \in \Gamma
    \]
    Axioms \ax{A_1} and \ax{A_2} ensure that $\fun^{\Gamma}_V$ is well-defined, as they guarantee $\Gamma$ has one and only one formula of the form $\int{\opint{\vec{U}}{\vec{u}},\opint{\vec{Y}}{\vec{y}}}{V{=}v}$ when $\vec{u}$, $\vec{y}$ and $V$ are fixed. The set of structural functions for $\NV$ in $\Gamma$ is defined as $\F^{\Gamma} := \set{\fun^\Gamma_V \mid V \in \NV}$.

    \item The valuation $\A^{\Gamma}$ is defined, for every $Z \in \AV$, as
    \[
      \A^\Gamma(Z) = z
      \quad\text{if and only if}\quad
      \int{\;}{Z{=}z} \in \Gamma
    \]
    Axioms \ax{A_1} and \ax{A_2} ensure that $\A^{\Gamma}$ is well-defined: they guarantee $\Gamma$ has one and only one formula of the form $\int{\;}{Z{=}z}$ for a fixed $Z$.
  \end{compactitemize}
\end{definition}

The next proposition shows that the just defined structure is indeed a recursive causal model.

\begin{proposition}\label{pro:rcm}
  Take $\Gamma \in \mcss$. The tuple $\tuple{\Sig, \F^{\Gamma}, \A^{\Gamma}}$ is a proper recursive causal model, that is, \begin{inlineenum} \item $\A^\Gamma$ complies with $\F^\Gamma$, and \item $\F^{\Gamma}$ is recursive\end{inlineenum}.
  \begin{proof}
    ~
    \begin{compactenumerate}
      \item Towards a contradiction, suppose $\A^\Gamma$ does not comply with $\F^\Gamma$. Then, there is $V \in \NV$ such that $\A^\Gamma(V) = v$ but $\fun^\Gamma_V(\A^\Gamma(\vec{U}), \A^\Gamma(\vec{Y})) \neq v$, with $\vec{U}$ a tuple with all exogenous variables in $\XV$ and $\vec{Y}$ a tuple with all endogenous variables in $\NV \setminus \set{V}$. Take $\A^\Gamma(\vec{U}) = \vec{u}$ and $\A^\Gamma(\vec{Y}) = \vec{y}$.

      From $\A^\Gamma$'s definition, $\A^\Gamma(\vec{U}) = \vec{u}$, $\A^\Gamma(\vec{Y}) = \vec{y}$ and $\A^\Gamma(V) = v$ imply that the formulas in $\set{\int{\;}{V{=}v}} \cup \set{\int{\;}{U_i{=}u_i} \mid U_i \in \vec{U}} \cup \set{\int{\;}{Y_i{=}y_i} \mid Y_i \in \vec{Y}}$ are all in $\Gamma$. This and axiom \ax{A_3} imply that $\int{\opint{\vec{U}}{\vec{u}}, \opint{\vec{Y}}{\vec{y}}}{V{=}v} \in \Gamma$.\footnote{Indeed, suppose $\set{\int{\;}{X_1{=}x_1}, \ldots, \int{\;}{X_\ell{=}x_\ell}} \in \Gamma$; take any formula $\int{\;}{X_k{=}x_k}$ with $k \in \intint{2}{\ell}$. Since $\Gamma$ is maximally consistent, $\int{\;}{X_1{=}x_1} \land \int{\;}{X_k{=}x_k}$ is in $\Gamma$, and so is the instance of axiom \ax{A_3} given by $(\int{\;}{X_1{=}x_1} \land \int{\;}{X_k{=}x_k}) \rightarrow \int{\opint{X_1}{x_1}}{X_k{=}x_k}$; hence, by modus ponens, $\int{\opint{X_1}{x_1}}{X_k{=}x_k} \in \Gamma$.  But this holds for any $k \in \intint{2}{\ell}$, so in fact $\set{\int{\opint{X_1}{x_1}}{X_2{=}x_2}, \ldots, \int{\opint{X_1}{x_1}}{X_\ell{=}x_\ell}} \in \Gamma$. A repetition of this argument, starting now from $\int{\opint{X_1}{x_1}}{X_2{=}x_2}$ and any formula $\int{\opint{X_1}{x_1}}{X_k{=}x_k}$ with $k \in \intint{3}{\ell}$, will produce $\set{\int{\opint{X_1}{x_1},\opint{X_2}{x_2}}{X_3{=}x_3}, \ldots, \int{\opint{X_1}{x_1},\opint{X_2}{x_2}}{X_\ell{=}x_\ell}} \in \Gamma$. Further repetitions will lead to the required $\int{\opint{X_1}{x_1}, \ldots, \opint{X_{\ell-1}}{x_{\ell-1}}}{X_\ell{=}x_\ell} \in \Gamma$.} But, from $\fun^\Gamma_V$'s definition, $\fun^\Gamma_V(\A^\Gamma(\vec{U}), \A^\Gamma(\vec{Y})) \neq v$ implies $\int{\opint{\vec{U}}{\vec{u}}, \opint{\vec{Y}}{\vec{y}}}{V{=}v} \notin \Gamma$, a contradiction.

      \item Towards a contradiction, suppose $\F^{\Gamma}$ is not recursive, i.e., suppose $\parof{\F^{\Gamma}}$ is not asymmetric. Then, there are $X_1 , X_2 \in \AV$ with $X_1 \parof{\F^{\Gamma}} X_2$ and $X_2 \parof{\F^{\Gamma}} X_1$, that is, there are $Y_1, \ldots, Y_p, W_1, \ldots, W_q$ in $\AV$ such that
      \begingroup
        \small
          \[
            X_1 \dirparof{\F^{\Gamma}} Y_1 \dirparof{\F^{\Gamma}} \cdots \dirparof{\F^{\Gamma}} Y_p \dirparof{\F^{\Gamma}} X_2 \dirparof{\F^{\Gamma}} W_1 \dirparof{\F^{\Gamma}} \cdots \dirparof{\F^{\Gamma}} W_q \dirparof{\F^{\Gamma}} X_1.
          \]
      \endgroup
      Now, note how, for any two variables $Z_1, Z_2 \in \AV$, if $Z_1 \dirparof{\F^{\Gamma}} Z_2$ then $Z_1 \syndirparof Z_2 \in \Gamma$.\footnote{Indeed, let $\vec{Z^-}$ be a vector containing all variables in $\AV \setminus \set{Z_1, Z_2}$, and suppose $Z_1 \dirparof{\F^{\Gamma}} Z_2$. By definition of $\dirparof{\F^{\Gamma}}$, there is a vector $\vec{z^-} \in \Ran(\vec{Z^-})$ and there are $z_1, z'_1 \in \Ran(Z_1)$ with $z_1 \neq z'_1$ such that, if $\fun^{\Gamma}_{Z_2}(\vec{z^-}, z_1) = z_2$ and $\fun^{\Gamma}_{Z_2}(\vec{z^-}, z'_1) = z'_2$ (with $\fun^{\Gamma}_{Z_2}$ the structural function for $X_2$ in $\F^\Gamma$), then $z_2 \neq z'_2$. Thus, from the definition of the structural functions in $\F^{\Gamma}$, it follows that $\int{\opint{\vec{Z^-}}{\vec{z^-}}, \opint{Z_1}{z_1}}{Z_2{=}z_2} \in \Gamma$ and $\int{\opint{\vec{Z^-}}{\vec{z^-}}, \opint{Z_1}{z'_1}}{Z_2{=}z'_2} \in \Gamma$ for $\vec{z^-} \in \Ran(\vec{Z^-})$, $z_1 \neq z'_1$ and $z_2 \neq z'_2$. Since $\Gamma$ is maximally consistent, the conjunction of both formulas is also in $\Gamma$, and hence so is $Z_1 \syndirparof Z_2$.} Thus, all formulas in
      \begingroup
        \small
          \[
            \left\{
              \begin{array}{l}
                X_1 \syndirparof Y_1,\, Y_1 \syndirparof Y_2,\, \ldots, Y_{p-1} \syndirparof Y_p,\, Y_p \syndirparof X_2, \\
                X_2 \syndirparof W_1,\, W_1 \syndirparof W_2,\, \ldots, W_{q-1} \syndirparof W_q,\, W_q \syndirparof X_1
              \end{array}
            \right\}
          \]
      \endgroup
      are in $\Gamma$, and so is their conjunction. But, by axiom \ax{A_6}, $(X_1 \syndirparof Y_1 \;\land\; \cdots \;\land\; W_{q-1} \syndirparof W_{q}) \,\rightarrow\, \lnot (W_{q} \syndirparof X_1) \in \Gamma$. This makes $\Gamma$ inconsistent; a contradiction.
    \end{compactenumerate}
  \end{proof}
\end{proposition}

In the following useful lemma, an expression of the form $\int{\opint{\vec{X}}{\vec{x}}}{\vec{Y}{=}\vec{y}}$ for $\vec{Y} = (Y_1, \ldots, Y_k)$ abbreviates the \LAcres-formula $\int{\opint{\vec{X}}{\vec{x}}}{Y_1{=}y_1} \land \cdots \land \int{\opint{\vec{X}}{\vec{x}}}{Y_k{=}y_k}$.

\begin{lemma}\label{lem:cut}
  $\vdashlocres
    \left( \int{\opint{\vec{X}}{\vec{x}}}{\vec{Y}{=}\vec{y}}
      \land
      \int{\opint{\vec{X}}{\vec{x}}, \opint{\vec{Y}}{\vec{y}}}{\vec{Z}{=}\vec{z}}
    \right)
    \rightarrow
    \int{\opint{\vec{X}}{\vec{x}}}{\vec{Z}{=}\vec{z}}$.
  \begin{proof}
    With the aim of reaching a contradiction, suppose that the set of formulas $\set{ \int{\opint{\vec{X}}{\vec{x}}}{\vec{Y}{=}\vec{y}}, \int{\opint{\vec{X}}{\vec{x}}, \opint{\vec{Y}}{\vec{y}}}{\vec{Z}{=}\vec{z}}, \lnot \int{\opint{\vec{X}}{\vec{x}}}{\vec{Z}{=}\vec{z}} }$ is \LOcres-consistent. Thus, by a standard Lindenbaum's construction (see, e.g., \citealp[Chapter 4]{BlackburnRijkeVenema2001}), there is a maximally \LOcres-consistent set of \LAcres-formulas $\Gamma' \in \mcss$ containing it: $\set{ \int{\opint{\vec{X}}{\vec{x}}}{\vec{Y}{=}\vec{y}}, \int{\opint{\vec{X}}{\vec{x}}, \opint{\vec{Y}}{\vec{y}}}{\vec{Z}{=}\vec{z}}, \lnot \int{\opint{\vec{X}}{\vec{x}}}{\vec{Z}{=}\vec{z}} } \subseteq \Gamma'$. Note how this forces the vector $\vec{Z}$ to have possible values other than $\vec{z}$ (i.e., $\card{\Ran(\vec{Z})} > 1$).\footnote{Otherwise, $\Ran(\vec{Z}) = \set{\vec{z}}$, and thus $\vdashlocres \int{\opint{\vec{X}}{\vec{x}}}{\vec{Z}{=}\vec{z}}$ (an instance of axiom \ax{A_2}), which implies $\int{\opint{\vec{X}}{\vec{x}}}{\vec{Z}{=}\vec{z}} \in \Gamma'$ and thus makes $\Gamma'$ \LOcres-inconsistent.}

    From axiom \ax{A_2}, the just obtained $\card{\Ran(\vec{Z})} > 1$ and propositional reasoning,
    \[
      \vdashlocres
        \lnot \int{\opint{\vec{X}}{\vec{x}}}{\vec{Z}{=}\vec{z}}
        \rightarrow
        \bigvee_{\vec{z'} \in \Ran(\vec{Z})\setminus\set{\vec{z}}} \int{\opint{\vec{X}}{\vec{x}}}{\vec{Z}{=}\vec{z'}}.
    \]
    Then, by modus ponens,
    \[ \bigvee_{\vec{z'} \in \Ran(\vec{Z})\setminus\set{\vec{z}}} \int{\opint{\vec{X}}{\vec{x}}}{\vec{Z}{=}\vec{z'}} \in \Gamma', \]
    which implies $\int{\opint{\vec{X}}{\vec{x}}}{\vec{Z}{=}\vec{z'}} \in \Gamma'$ for some $\vec{z'} \neq \vec{z}$. But consider the following instance of axiom \ax{A_3}:
    \[
      \vdashlocres
        \left(
          \int{\opint{\vec{X}}{\vec{x}}}{(\vec{Y}{=}\vec{y})}
          \land
          \int{\opint{\vec{X}}{\vec{x}}}{(\vec{Z}{=}\vec{z'})}
        \right)
        \rightarrow
        \int{\opint{\vec{X}}{\vec{x}}, \opint{\vec{Y}}{\vec{y}}}{(\vec{Z}{=}\vec{z'})}.
    \]
    As $\int{\opint{\vec{X}}{\vec{x}}}{(\vec{Y}{=}\vec{y})}$ and $\int{\opint{\vec{X}}{\vec{x}}}{(\vec{Z}{=}\vec{z'})}$ are both in $\Gamma'$, so is their conjunction, and therefore so is $\int{\opint{\vec{X}}{\vec{x}}, \opint{\vec{Y}}{\vec{y}}}{(\vec{Z}{=}\vec{z'})}$ (by modus ponens). Moreover: from $\vdashlocres \int{\opint{\vec{X}}{\vec{x}}, \opint{\vec{Y}}{\vec{y}}}{\vec{Z}{=}\vec{z'}} \rightarrow \lnot \int{\opint{\vec{X}}{\vec{x}}, \opint{\vec{Y}}{\vec{y}}}{\vec{Z}{=}\vec{z}}$ (an instance of axiom \ax{A_1}, using $\vec{z'} \neq \vec{z}$), it follows that $\lnot \int{\opint{\vec{X}}{\vec{x}}, \opint{\vec{Y}}{\vec{y}}}{\vec{Z}{=}\vec{z}} \in \Gamma'$. This makes $\Gamma'$ \LOcres-inconsistent: a contradiction.
  \end{proof}
\end{lemma}

The next step is to prove a truth lemma: every $\gamma \in \LAcres$ is such that $\tuple{\Sig, \F^{\Gamma}, \A^{\Gamma}} \models \gamma$ if and only if $\gamma \in \Gamma$. The following proposition takes care of the crucial base case.

\begin{proposition}\label{pro:LAres:truth-lemma:atoms}
  Let $\Gamma \in \mcss$ be a maximally \LOcres-con\-sis\-tent set of formulas in \LAcres. Let $\opint{\vec{X}}{\vec{x}}$ be an assignment, with $\vec{X}$ a tuple of variables in $\AV$; take $Y \in \AV$ and $y \in \Ran(Y)$. Then,
  \begin{ctabular}{c}
    $\int{\opint{\vec{X}}{\vec{x}}}{Y{=}y} \in \Gamma
    \qquad\text{if and only if}\qquad
    \tuple{\Sig, \F^\Gamma, \A^\Gamma} \models \int{\opint{\vec{X}}{\vec{x}}}{Y{=}y}$.
  \end{ctabular}
  \begin{proof}
    By sem. interpretation, the right-hand side $\tuple{\Sig, \F^\Gamma, \A^\Gamma} \models \int{\opint{\vec{X}}{\vec{x}}}{Y{=}y}$ is equivalent to $\newval{\A}{\F}{\opint{\vec{X}}{\vec{x}}}(Y) = y$. Then, the proof will show that, for any assignment $\opint{\vec{X}}{\vec{x}}$ on $\AV$, any $Y \in \AV$ and any $y \in \Ran(Y)$,
    \[
      \int{\opint{\vec{X}}{\vec{x}}}{Y{=}y} \in \Gamma
      \qquad\text{if and only if}\qquad
      \newval{\A^\Gamma}{\F}{\opint{\vec{X}}{\vec{x}}}(Y) = y
    \]
    Thus, take any $\opint{\vec{X}}{\vec{x}}$, any $Y \in \AV$ and any $y \in \Ran(Y)$.

    {\bigskip\medskip}

    There are two main cases. First, suppose $Y \in \XV$.
    \begin{compactitemize}
      \item Suppose further that $Y$ occurs in $\vec{X}$, so $Y=X_i$ for some $i \in \intint{1}{\card{\vec{X}}}$. \prooflr Suppose $\int{\opint{\vec{X}}{\vec{x}}}{X_i{=}y} \in \Gamma$. By axiom \ax{A_4}, $\int{\opint{\vec{X}}{\vec{x}}}{X_i{=}x_i} \in \Gamma$; thus, axiom \ax{A_1} and the consistency of $\Gamma$ imply $y=x_i$. Now, from the definition of the value of intervened variables after an intervention (Definition \ref{def:int}), it follows that $\newval{\A^\Gamma}{\F}{\opint{\vec{X}}{\vec{x}}}(X_i) = x_i$; this, together with $y=x_i$, produces the required $\newval{\A^\Gamma}{\F}{\opint{\vec{X}}{\vec{x}}}(X_i) = y$. \proofrl Suppose $\newval{\A^\Gamma}{\F}{\opint{\vec{X}}{\vec{x}}}(X_i) = y$. From Definition \ref{def:int} again it follows that $\newval{\A^\Gamma}{\F}{\opint{\vec{X}}{\vec{x}}}(X_i) = x_i$, so $y=x_i$. Now, by axiom \ax{A_4} again, $\int{\opint{\vec{X}}{\vec{x}}}{X_i{=}x_i} \in \Gamma$ so, since $y = x_i$, it follows that $\int{\opint{\vec{X}}{\vec{x}}}{X_i{=}y} \in \Gamma$.

      \item Suppose $Y$ does not occur in $\vec{X}$. By axiom \ax{A_7}, $\int{\opint{\vec{X}}{\vec{x}}}{Y{=}y} \in \Gamma$ if and only if $\int{\;}{Y{=}y} \in \Gamma$; by the definition of $\A^\Gamma$ (Definition \ref{def:cm}), $\int{\;}{Y{=}y} \in \Gamma$ if and only if $\A^\Gamma(Y) = y$; by the definition of the value an intervened valuation assigns to a non-intervened exogenous variable (Definition \ref{def:int}), $\A^\Gamma(Y) = y$ if and only if $\newval{\A^\Gamma}{\F}{\opint{\vec{X}}{\vec{x}}}(Y) = y$.
    \end{compactitemize}

    {\bigskip}

    Suppose now $Y \in \NV$. The proof proceeds by induction on the number of \emph{non-intervened endogenous} variables, i.e., by induction on the size of $\NV \setminus \vec{X}$.
    \begin{compactitemize}
      \item \textbf{Case $\bs{\card{\NV \setminus \vec{X}}=0}$.} This is the case when every endogenous variable is being intervened; in particular, $Y$ is, i.e., $Y$ occurs in $\vec{X}$. Then, the argument for the case $Y \in \XV$ with $Y$ occurring in $\vec{X}$ shows that the equivalence holds.

      \item \textbf{Case $\bs{\card{\NV \setminus \vec{X}}=1}$.} If $Y$ is being intervened (i.e., $Y$ occurs in $\vec{X}$), then the argument for the case $\card{\NV \setminus \vec{X}}=0$ is enough.

      If $Y$ is the lone non-intervened endogenous variable, then $\vec{X}$ contains all variables in $\NV \setminus \set{Y}$. Define $\opint{\vec{U'}}{\vec{u'}}$ as the assignment over the exogenous variables not in $\vec{X}$ (i.e., $U' \in \vec{U'}$ if and only if both $U' \in \XV$ and $U' \notin \vec{X}$) by taking $u'_i := \A^\Gamma(U'_i)$. From the definition of $\A^\Gamma$, it is clear that $\int{\;}{U'_i{=}u'_i} \in \Gamma$ for all $U'_i \in \vec{U'}$, that is, $\int{\;}{\vec{U'}{=}\vec{u'}} \in \Gamma$. Note how the disjoint vectors $\vec{X}$ and $\vec{U'}$ contain, together, exactly all the variables in $\AV \setminus \set{Y}$. Note also how, from the definition of intervention (Definition \ref{def:int}), it follows that
      \[
        \newval{\A^\Gamma}{\F}{\opint{\vec{X}}{\vec{x}}}(Y)
        =
        \newval{\A^\Gamma}{\F}{\opint{\vec{X}\vec{U'}}{\vec{x}\vec{u'} }}(Y)
        =
        \fun^{\Gamma}_Y(\vec{x},\vec{u'}).
      \]
      Now, for the equivalence, take $\newval{\A^\Gamma}{\F}{\opint{\vec{X}}{\vec{x}}}(Y) = y$. By the just stated equality, this holds if and only if $\fun^{\Gamma}_Y(\vec{x},\vec{u'}) = y$. But then, by the construction of $\fun^{\Gamma}_Y$ (Definition \ref{def:cm}), $\fun^{\Gamma}_Y(\vec{x},\vec{u'}) = y$ holds if and only if $\int{\opint{\vec{X}}{\vec{x}}, \opint{\vec{U'}}{\vec{u'}}}{Y{=}y} \in \Gamma$. Finally, in the presence of $\int{\opint{\vec{X}}{\vec{x}}}{\vec{U'}{=}\vec{u'}} \in \Gamma$ (a consequence of the previous $\int{\;}{\vec{U'}{=}\vec{u'}} \in \Gamma$ and axiom \ax{A_7}), $\int{\opint{\vec{X}}{\vec{x}}, \opint{\vec{U'}}{\vec{u'}}}{Y{=}y} \in \Gamma$ holds if and only if the required $\int{\opint{\vec{X}}{\vec{x}}}{Y{=}y} \in \Gamma$ holds (by Lemma \ref{lem:cut} in one direction, and by axiom \ax{A_3} in the other).

      \item \textbf{Case $\bs{\card{\NV \setminus \vec{X}} = k > 1}$.} If $Y$ is being intervened, equivalence follows as shown in the case $\card{\NV \setminus \vec{X}}=0$.

      Suppose $Y$ is not being intervened. Define $\opint{\vec{U'}}{\vec{u'}}$ as in the previous case.

      \prooflr Suppose $\int{\opint{\vec{X}}{\vec{x}}}{Y{=}y} \in \Gamma$. The strategy is to build a complete valuation $\A^*$, and then show that $\A^*$ \begin{inlineenum} \item agrees with $\A^\Gamma$ on the values of all exogenous variables not in $\vec{X}$, \item follows $\opint{\vec{X}}{\vec{x}}$ for the values of exogenous variables in $\vec{X}$, and \item complies with all structural functions in $\newf{\F^\Gamma}{\opint{\vec{X}}{\vec{x}}}$\end{inlineenum}. There is a unique valuation satisfying these three requirements (by Proposition \ref{pro:rcm}, $\F^\Gamma$ is recursive), so it will follow that $\A^* = \newval{\A^\Gamma}{\F}{\opint{\vec{X}}{\vec{x}}}$. As it will be shown, the assumption implies $\A^*(Y) = y$ and thus also the required $\newval{\A^\Gamma}{\F}{\opint{\vec{X}}{\vec{x}}}(Y) = y$.

      Recall that $\vec{U'}$ contains exactly all exogenous variables not in $\vec{X}$; let $\vec{V'}$ be the vector containing exactly all \emph{endogenous} variables not in $\vec{X}$. Then, define
      \begin{compactitemize}
        \item $\A^*(X_i) := x_i$ for $X_i \in \vec{X}$;
        \item $\A^*(U'_i) := u'_i$ for $U'_i \in \vec{U'}$;
        \item $\A^*(V'_i) := v'_i$ if and only if $\int{\opint{\vec{X}}{\vec{x}}}{V'_i{=}v'_i} \in \Gamma$, for $V'_i \in \vec{V'}$.\footnote{By axioms \ax{A_1} and \ax{A_2}, this uniquely determines the value of each variable in $\vec{V'}$.}
      \end{compactitemize}

      Now, for the three points mentioned above.
      \begin{compactenumerate}
        \item $\A^*$ agrees with $\A^\Gamma$ on the values of all exogenous variables not in $\vec{X}$ (i.e., variables in $\vec{U'}$) because $\vec{u'}$ is directly taken from $\A^\Gamma$.

        \item $\A^*$ follows $\opint{\vec{X}}{\vec{x}}$ for the values of all (in particular, the exogenous) variables in $\vec{X}$.

        \item It is only left to show that $\A^*$ complies with $\newf{\F^\Gamma}{\vec{X}=\vec{x}}$. For notation, use $z^*$ to denote the value a variable $Z$ receives according to $\A^*$. Note how, since $\card{\NV \setminus \vec{X}} > 1$, there are at least two endogenous variables that are not being intervened (i.e., there are at least two variables in $\vec{V'}$); denote them by $W_1$ and $W_2$. From the definition of the values in $\vec{v'}$, it follows that $\int{\opint{\vec{X}}{\vec{x}}}{W_1{=}w^*_1} \in \Gamma$ and $\int{\opint{\vec{X}}{\vec{x}}}{W_2{=}w^*_2} \in \Gamma$.

        For the proof, it should be shown that, for every \emph{endogenous} variable $Z$, the value $z^*$ complies with the structural function for $Z$ in $\newf{\F^\Gamma}{\vec{X}=\vec{x}}$. Take any endogenous variable $Z$ different from $W_1$. If $Z$ is in $\vec{X}$, from axiom \ax{A_4} it follows that $\int{\opint{\vec{X}}{\vec{x}}, \opint{W_1}{w^*_1}}{Z{=}z^*} \in \Gamma$. Otherwise, $Z$ is not in $\vec{X}$, so $Z$ is in $\vec{V'}$ and therefore $\int{\opint{\vec{X}}{\vec{x}}}{Z{=}z^*} \in \Gamma$. But $\int{\opint{\vec{X}}{\vec{x}}}{W_1{=}w^*_1} \in \Gamma$ so, by axiom \ax{A_3}, $\int{\opint{\vec{X}}{\vec{x}}, \opint{W_1}{w^*_1}}{Z{=}z^*} \in \Gamma$. Thus, $\int{\opint{\vec{X}}{\vec{x}}, \opint{W_1}{w^*_1}}{Z{=}z^*} \in \Gamma$ holds for every $Z \in \NV$ different from $W_1$. Now, since $\card{\NV \setminus (\vec{X} \cup \set{W_1})} = k-1$, from inductive hypothesis it follows that $\newval{\A^\Gamma}{\F}{\opint{\vec{X}}{\vec{x}},\opint{W_1}{w_1^*}}(Z) = z^*$. Moreover: $\A^*$ agrees with $\A^\Gamma$ in all variables not in $\set{\vec{X}, W_1}$, so $\A^*$ complies with the structural function for $Z$ from $\newf{\F^\Gamma}{\opint{\vec{X}}{\vec{x}}, \opint{W_1}{w_1^*}}$. But $Z$ is different from $W_1$, and thus its structural equation in $\newf{\F^\Gamma}{\opint{\vec{X}}{\vec{x}}, \opint{W_1}{w_1^*}}$ is exactly the same as that in $\newf{\F^\Gamma}{\opint{\vec{X}}{\vec{x}}}$. Hence, $\A^*$ complies with the structural function for $Z$ from $\newf{\F^\Gamma}{\opint{\vec{X}}{\vec{x}}}$.

        Thus, for any $Z$ different from $W_1$, the valuation $\A^*$ complies with the structural function for $Z$ at $\newf{\F^\Gamma}{\opint{\vec{X}}{\vec{x}}}$. An analogous reasoning shows that, for any $Z$ different from $W_2$, the valuation $\A^*$ complies with the structural function for $Z$ at $\newf{\F^\Gamma}{\opint{\vec{X}}{\vec{x}}}$. Thus, for every endogenous variable $Z$, the valuation $\A^*$ complies with the structural function for $Z$ at $\newf{\F^\Gamma}{\opint{\vec{X}}{\vec{x}}}$.
      \end{compactenumerate}
      Hence, $\A^* = \newval{\A^\Gamma}{\F}{\opint{\vec{X}}{\vec{x}}}$. For the final point, $Y$ is in $\vec{V'}$; thus, from the assumption $\int{\opint{\vec{X}}{\vec{x}}}{Y{=}y} \in \Gamma$ it follows that $\A^*(Y) = y$, that is, $\newval{\A^\Gamma}{\F}{\opint{\vec{X}}{\vec{x}}}(Y) = y$, as required.

      {\medskip}

      \proofrl Suppose $\newval{\A^\Gamma}{\F}{\opint{\vec{X}}{\vec{x}}}(Y) = y$. Since $\card{\NV \setminus \vec{X}} = k > 1$, there are at least two endogenous variables not in $\vec{X}$. One of them is $Y$; let $W$ be one of the others, and let $w \in \Ran(W)$ be the value satisfying $\newval{\A^\Gamma}{\F}{\opint{\vec{X}}{\vec{x}}}(W) = w$.
      \begin{compactitemize}
        \item Consider the valuation $\newval{\A^\Gamma}{\F}{\opint{\vec{X}}{\vec{x}},\opint{W}{w}}$. It intervenes on one variable more than $\newval{\A^\Gamma}{\F}{\opint{\vec{X}}{\vec{x}}}$ (namely, $W$), but it assigns to it the same value (namely, $w$). Thus, both valuations are identical, and hence $\newval{\A^\Gamma}{\F}{\opint{\vec{X}}{\vec{x}},\opint{W}{w}}(Y) = y$. As $\card{\NV \setminus (\vec{X} \cup \set{W})} = k-1$, from the inductive hypothesis it follows that $\int{\opint{\vec{X}}{\vec{x}}, \opint{W}{w}}{Y{=}y} \in \Gamma$.
        \item Consider the valuation $\newval{\A^\Gamma}{\F}{\opint{\vec{X}}{\vec{x}},\opint{Y}{y}}$. It intervenes on one variable more than $\newval{\A^\Gamma}{\F}{\opint{\vec{X}}{\vec{x}}}$ (namely, $Y$), but it assigns to it the same value (namely, $y$). Thus, both valuations are identical, and hence $\newval{\A^\Gamma}{\F}{\opint{\vec{X}}{\vec{x}},\opint{Y}{y}}(W) = w$. As $\card{\NV \setminus (\vec{X} \cup \set{Y})} = k-1$, from the inductive hypothesis it follows that $\int{\opint{\vec{X}}{\vec{x}}, \opint{Y}{y}}{W{=}w} \in \Gamma$.
      \end{compactitemize}
      Thus, $\int{\opint{\vec{X}}{\vec{x}}, \opint{W}{w}}{Y{=}y} \in \Gamma$ and $\int{\opint{\vec{X}}{\vec{x}}, \opint{Y}{y}}{W{=}w} \in \Gamma$. Then, since $Y \neq W$, the required $\int{\opint{\vec{X}}{\vec{x}}}{Y{=}y} \in \Gamma$ follows from axiom \ax{A_5}.
    \end{compactitemize}
  \end{proof}
\end{proposition}

Here is the Truth Lemma at its fullest.

\begin{lemma}[Truth Lemma, \LAcres]\label{lem:LAres:truth-lemma}
  Take $\Gamma \in \mcss$. Then, for all $\gamma \in \LAcres$,
  \begin{ctabular}{c}
    $\tuple{\Sig, \F^\Gamma, \A^\Gamma} \models \gamma
    \qquad\text{if and only if}\qquad
    \gamma \in \Gamma.$
  \end{ctabular}
  \begin{proof}
    The proof is by induction on $\gamma \in \Gamma$.
    \begin{compactitemize}
      \item \textbf{Case $\bs{\int{\opint{\vec{X}}{\vec{x}}}{Y{=}y}}$.} Proposition \ref{pro:LAres:truth-lemma:atoms}.

      \item \textbf{Case $\bs{\lnot \gamma}$} and \textbf{Case $\bs{\gamma_1 \land \gamma_2}$.} Immediate from IHs and the properties of a maximally consistent set.
    \end{compactitemize}
  \end{proof}
\end{lemma}

Then, the full argument for completeness of \LOcres for formulas in \LAcres.

\begin{theorem}[Completeness of \LOcres for \LAcres]\label{teo:completeness-lores}
  The system \LOcres is strongly complete for the language \LAcres based on $\Sig$ with respect to recursive causal models for $\Sig$. In other words, for every set of \LAcres-formulas $\Gamma^- \cup \set{\gamma}$, if $\Gamma^- \models \gamma$ then $\Gamma^- \vdashlocres \gamma$.
  \begin{proof}
    The argument works by contraposition, so take any \LOcres-consistent set of formulas $\Gamma^-$. Since \LAcres is enumerable, $\Gamma^-$ can be extended into a maximally $\LOcres$-consistent set $\Gamma$ via a standard Lindenbaum's construction (see, e.g., \citealp[Chapter 4]{BlackburnRijkeVenema2001}). By Lemma \ref{lem:LAres:truth-lemma}, all formulas in $\Gamma$ are satisfiable in $\tuple{\Sig, \F^\Gamma, \A^\Gamma}$, which by Proposition \ref{pro:rcm} is a recursive causal model. Thus, $\Gamma^-$ is satisfiable in a recursive causal model.
  \end{proof}
\end{theorem}

{\bigskip}

Finally, the argument for strong completeness of \LOc for formulas in \LAc\label{proof:thm:rcm-LA} proceeds in three steps.
\begin{enumerate}
  \item Take $\Psi \cup \set{\varphi} \subseteq \LAc$; suppose $\Psi \models \varphi$, i.e., suppose that, for every causal model $\tuple{\Sig, \F, \A}$, if $\tuple{\Sig, \F, \A} \models \Psi$ then $\tuple{\Sig, \F, \A} \models \varphi$ or, in other words, for every $\tuple{\Sig, \F, \A}$,
  \[
    \tuple{\Sig, \F, \A} \models \psi \;\text{for all}\; \psi \in \Psi
    \qquad\text{implies}\qquad
    \tuple{\Sig, \F, \A} \models \varphi.
  \]
  Since $\models \varphi' \leftrightarrow \tro(\varphi')$ for every $\varphi' \in \LAc$ (Proposition \ref{pro:LAtoLAres}.\ref{pro:LAtoLAres:sem}), it follows that, for every $\tuple{\Sig, \F, \A}$,
  \begin{compactitemize}
    \item $\tuple{\Sig, \F, \A} \models \tro(\psi)$ for all $\psi \in \Psi$ if and only if $\tuple{\Sig, \F, \A} \models \psi$ for all $\psi \in \Psi$, and
    \item $\tuple{\Sig, \F, \A} \models \varphi$ if and only if $\tuple{\Sig, \F, \A} \models \tro(\varphi)$.
  \end{compactitemize}
  Thus, for every $\tuple{\Sig, \F, \A}$,
  \[
    \tuple{\Sig, \F, \A} \models \tro(\psi) \;\text{for all}\; \psi \in \Psi
    \qquad\text{implies}\qquad
    \tuple{\Sig, \F, \A} \models \tro(\varphi).
  \]
  By defining $\tro(\Psi) := \set{\tro(\psi) \mid \psi \in \Psi}$, the previous actually states that $\tuple{\Sig, \F, \A} \models \tro(\Psi)$ implies $\tuple{\Sig, \F, \A} \models \tro(\varphi)$ for every $\tuple{\Sig, \F, \A}$; in other words, $\tro(\Psi) \models \tro(\varphi)$.

  \item Since $\tro(\varphi') \in \LAcres$ for every $\varphi' \in \LAc$ (Proposition \ref{pro:LAtoLAres}.\ref{pro:LAtoLAres:trans}), it follows that $\tro(\Psi) \cup \set{\tro(\varphi)} \subseteq \LAcres$; therefore, the just obtained $\tro(\Psi) \models \tro(\varphi)$ and Theorem \ref{teo:completeness-lores} imply $\tro(\Psi) \vdashlocres \tro(\varphi)$. Since \LOcres is a subsystem of \LOc, it follows that $\tro(\Psi) \vdashloc \tro(\varphi)$.

  \item Since $\tro(\Psi) \vdashloc \tro(\varphi)$, there are $\psi'_1, \ldots, \psi'_n \in \tro(\Psi)$ such that
  \[ \vdashloc \left(\psi'_1 \land \cdots \land \psi'_n\right) \rightarrow \tro(\varphi). \]
  Then, from the definition of $\tro(\Psi)$, there are $\psi_1, \ldots, \psi_n \in \Psi$ such that
  \[ \vdashloc \left(\tro(\psi_1) \land \cdots \land \tro(\psi_n)\right) \rightarrow \tro(\varphi). \]
  But $\vdashloc \varphi' \leftrightarrow \tro(\varphi')$ for every $\varphi' \in \LAc$ (Proposition \ref{pro:LAtoLAres}.\ref{pro:LAtoLAres:syn}); hence (by propositional reasoning on the first),
  \[
    \vdashloc \left(\psi_1 \land \cdots \land \psi_n\right) \rightarrow \left(\tro(\psi_1) \land \cdots \land \tro(\psi_n)\right)
    \quad\text{and}\quad
    \vdashloc \tro(\varphi) \rightarrow \varphi.
  \]
  and therefore
  \[ \vdashloc \left(\psi_1 \land \cdots \land \psi_n\right) \rightarrow \varphi. \]
  Hence, $\Psi \vdashloc \varphi$, as required.
\end{enumerate}

\end{appendix:full}

\begin{appendix:short}

\myparagraph{Soundness.} For \ax{P} and \ax{MP} this is straightforward; for \ax{A_1-A_5} and \ax{A_\land} this has been proved in \citet{halpern2000axiomatizing}. Axioms \ax{A_6} and \ax{A_\lnot} are sound because we work on \emph{recursive} causal models. For the first, a recursive $\F$ produces a relation $\dirparof{\F}$ without cycles (Footnote~\ref{ftn:recursive}), syntactically characterised by $\syndirparof$. For the second, in a recursive causal model, the value of each variable is uniquely determined.

For \ax{A_7} note how, for any assignment $\vec{X}{=}\vec{x}$, the valuations $\A$ and $\newval{\A}{\F}{\opint{\vec{X}}{\vec{x}}}$ coincide in the value of \emph{exogenous} variables not occurring in $\vec{X}$ (Definition \ref{def:int}). For \ax{A_{[]}}, an intervention with the empty assignment does not affect the given causal model. Finally, \ax{A_{[][]}} states that, when two interventions are performed in a row, the second overrides the first in the variables they both act upon.

\myparagraph{Completeness.} For completeness, it will be shown that axioms \ax{A_{[]}}, \ax{A_\lnot}, \ax{A_\land} and \ax{A_{[][]}} define a translation from \LAc to a language \LAcres for which axioms \ax{A_1}-\ax{A_7}, \ax{P} and rule \ax{MP} are complete. Here are the steps.

\myparagraphit{Completeness, step 1: from \LAc to \LAcres.} Formulas $\gamma$ of the language \LAcres \citep{halpern2000axiomatizing} consists of Boolean combinations of expressions of the form $\int{\opint{\vec{X}}{\vec{x}}}{Y{=}y}$. They are semantically interpreted over causal models as before.

\begin{definition}[Translation $\tro$]\label{def:tro}
  Define $\tro:\LAc \to \LAcres$ as
  \begingroup
    \small
    \[
      \renewcommand{\arraystretch}{1.6}
      \begin{array}{@{}c@{\quad}c@{}}
        \begin{array}{@{}r@{\,:=\,}l@{}}
          \tro(Y{=}y)               & \int{\;}{Y{=}y} \\
          \tro(\lnot\varphi)           & \lnot \tro(\varphi) \\
          \tro(\varphi_1 \land \varphi_2) & \tro(\varphi_1) \land \tro(\varphi_2) \\
          \multicolumn{2}{@{}r}{}
        \end{array}
        &
        \begin{array}{@{}r@{\,:=\,}l@{}}
          \tro(\int{\opint{\vec{X}}{\vec{x}}}{Y{=}y})                                & \int{\opint{\vec{X}}{\vec{x}}}{Y{=}y} \\
          \tro(\int{\opint{\vec{X}}{\vec{x}}}{\lnot \varphi})                           & \tro(\lnot \int{\opint{\vec{X}}{\vec{x}}}{\varphi}) \\
          \tro(\int{\opint{\vec{X}}{\vec{x}}}{(\varphi_1 \land \varphi_2)})                & \tro(\int{\opint{\vec{X}}{\vec{x}}}{\varphi_1} \land \int{\opint{\vec{X}}{\vec{x}}}{\varphi_2}) \\
          \tro(\int{\opint{\vec{X}}{\vec{x}}}{\int{\opint{\vec{Y}}{\vec{y}}}{\varphi}}) & \tro(\int{\opint{\vec{X'}}{\vec{x'}}, \opint{\vec{Y}}{\vec{y}}}{\varphi}) \\
        \end{array}
      \end{array}
    \]
  \endgroup
  with $\opint{\vec{X'}}{\vec{x'}}$ the subassignment of $\opint{\vec{X}}{\vec{x}}$ for $\vec{X'} := \vec{X} \setminus \vec{Y}$.
\end{definition}

\noindent The following proposition indicates the crucial properties of $\tro$.

\begin{proposition}\label{pro:LAtoLAres}
  For every $\varphi \in \LAc$,
  \begin{multicols}{3}
    \begin{enumerate}
      \item\label{pro:LAtoLAres:trans} $\tro(\varphi) \in \LAcres$,
      \item\label{pro:LAtoLAres:syn} $\vdashloc \varphi \leftrightarrow \tro(\varphi)$,
      \item\label{pro:LAtoLAres:sem} $\models \varphi \leftrightarrow \tro(\varphi)$.
    \end{enumerate}
  \end{multicols}
  \begin{proof}[Sketch]
    The proofs of this proposition use induction on a notion of \emph{complexity} $\com:\LAc \to \Nat\setminus\set{0}$, defined as
    \begin{smallctabular}{c@{\qquad}c}
      \begin{tabular}{r@{\;:=\;}l}
        $\com(Y{=}y)$         & $1$ \\
        $\com(\lnot \varphi)$ & $1 + \com(\varphi)$
      \end{tabular}
      &
      \begin{tabular}{r@{\;:=\;}l}
        $\com(\varphi_1 \land \varphi_2)$               & $1 + \max\set{\com(\varphi_1), \com(\varphi_2)}$ \\
        $\com(\int{\opint{\vec{X}}{\vec{x}}}{\varphi})$ & $2\com(\varphi)$
      \end{tabular}
    \end{smallctabular}
    Note: for all assignments $\opint{\vec{X}}{\vec{x}}$ and $\opint{\vec{Y}}{\vec{y}}$, and all formulas $\varphi, \varphi_1, \varphi_2 \in \LAc$,
    \begingroup
      \small
      \begin{center}
        $\begin{array}{r@{\;}c@{\;}l}
          \toprule
          \com(\varphi) & \geqslant & \com(\psi) \;\text{for every}\; \psi \in \sub(\varphi) \\
          \midrule
          \com(\int{\opint{\vec{X}}{\vec{x}}}{\lnot \varphi}) & > & \com(\lnot \int{\opint{\vec{X}}{\vec{x}}}{\varphi}) \\
          \com(\int{\opint{\vec{X}}{\vec{x}}}{(\varphi_1 \land \varphi_2)}) & > & \com(\int{\opint{\vec{X}}{\vec{x}}}{\varphi_1} \land \int{\opint{\vec{X}}{\vec{x}}}{\varphi_2}) \\
          \com(\int{\opint{\vec{X}}{\vec{x}}}{\int{\opint{\vec{Y}}{\vec{y}}}{\varphi}}) & > & \com(\int{\opint{\vec{X'}}{\vec{x'}}, \opint{\vec{Y}}{\vec{y}}}{\varphi}) \\
          \bottomrule
        \end{array}$
      \end{center}
    \endgroup
    \noindent Thus, in every case in Definition~\ref{pro:LAtoLAres}, the complexity of the formulas under $\tro$ on the right-hand side is strictly smaller than the complexity of the formula under $\tro$ on the left-hand side. Hence, $\tro$ is a proper recursive translation. With this, the first and the second item can be proved by induction on $\com(\varphi)$ (the first using $\tro$'s definition and the inductive hypothesis (IH); the second using, additionally, propositional reasoning and axioms \ax{A_{[]}}, \ax{A_\lnot}, \ax{A_\land} and \ax{A_{[][]}}). The third item follows from the second and the system's soundness.
  \end{proof}
\end{proposition}

\myparagraphit{Completeness, step 2: a canonical model for \LAcres.} The next step is to show that \LOcres, the fragment of \LOc without axioms \ax{A_{[]}}, \ax{A_\lnot}, \ax{A_\land} and \ax{A_{[][]}}, is complete for \LAcres over recursive causal models. This is done by showing, via the construction of a canonical model, that any \LOcres-consistent set of \LAcres-formulas is satisfiable in a recursive causal model. The construction is almost exactly as that in \citet{halpern2000axiomatizing}, so here we will only describe the main tasks.

{\smallskip}

Fix the signature $\Sig = \tuple{\XV, \NV, \Ran}$. Let $\mcss$ be the set of all maximally \LOcres-consistent sets of \LAcres-formulas. The first task is to show that each $\Gamma \in \mcss$ gives raise to a recursive causal model. A valuation $\A^\Gamma$ is defined by picking as $\A(Z)$ (for $Z \in \AV$) the unique $z$ such that $\int{\;}{Z{=}z} \in \Gamma$. Similarly, the structural function for an endogenous variable $V$ is read off from the formulas of the form $\int{\opint{\vec{U}}{\vec{u}},\opint{\vec{Y}}{\vec{y}}}{V{=}v}$ (with $\vec{U} = \XV$ and $\vec{Y} = \NV \setminus \set{V}$) that occur in $\Gamma$.  It can be shown that the resulting structure, $\tuple{\Sig, \F^{\Gamma}, \A^{\Gamma}}$ is a proper recursive causal model (i.e., $\F^{\Gamma}$ is recursive and $\A^\Gamma$ complies with $\F^\Gamma$).

The second task is to prove a truth lemma: every $\gamma \in \LAcres$ is such that $\tuple{\Sig, \F^{\Gamma}, \A^{\Gamma}} \models \gamma$ if and only if $\gamma \in \Gamma$. The crucial base case is as in \citet{halpern2000axiomatizing}; the Boolean cases rely on the IH. With it, one can prove the following.

\begin{theorem}[Completeness of \LOcres for \LAcres]\label{teo:completeness-lores}
  The system \LOcres is strongly complete for \LAcres based on $\Sig$ w.r.t. recursive causal models for $\Sig$. In other words, for every set of \LAcres-formulas $\Gamma^- \cup \set{\gamma}$, if $\Gamma^- \models \gamma$ then $\Gamma^- \vdashlocres \gamma$.
\end{theorem}

\noindent Finally, the argument for strong completeness of \LOc for formulas in \LAc\label{proof:thm:rcm-LA} is as follows. First, take $\Psi \cup \set{\varphi} \subseteq \LAc$ and suppose $\Psi \models \varphi$. Since $\models \varphi' \leftrightarrow \tro(\varphi')$ for every $\varphi' \in \LAc$ (Proposition \ref{pro:LAtoLAres}.\ref{pro:LAtoLAres:sem}), by defining $\tro(\Psi) := \set{\tro(\psi) \mid \psi \in \Psi}$ it follows that $\tro(\Psi) \models \tro(\varphi)$. Now, $\tro(\varphi') \in \LAcres$ for every $\varphi' \in \LAc$ (Proposition \ref{pro:LAtoLAres}.\ref{pro:LAtoLAres:trans}); hence, $\tro(\Psi) \cup \set{\tro(\varphi)} \subseteq \LAcres$. Therefore, the just obtained $\tro(\Psi) \models \tro(\varphi)$ and Theorem \ref{teo:completeness-lores} imply $\tro(\Psi) \vdashlocres \tro(\varphi)$. Since \LOcres is a subsystem of \LOc, it follows that $\tro(\Psi) \vdashloc \tro(\varphi)$. But $\vdashloc \varphi' \leftrightarrow \tro(\varphi')$ for every $\varphi' \in \LAc$ (Proposition \ref{pro:LAtoLAres}.\ref{pro:LAtoLAres:syn}); hence, $\Psi \vdashloc \varphi$, as required.
\end{appendix:short}

\subsection{Proof of Theorem \ref{thm:rcm-LAk}}\label{app:rcm-LAk}

\begin{appendix:full}
\myparagraph{Soundness.} For the soundness of axioms and rules in Table \ref{tbl:rcm-LA}, note that they do not make use of the $K$ operator; hence, their truth-value in a given pointed epistemic causal model $(\tuple{\Sig, \F, \T}, \A)$ only depends on the causal model $\tuple{\Sig, \F, \A}$, for which these axioms and rules are sound (Theorem \ref{thm:rcm-LA}).

For axioms and rules on Table \ref{tbl:rcm-LAk}, those in the \emph{epistemic} part are known to be sound on relational structures with a single equivalence relation \citep{RAK1995,BlackburnRijkeVenema2001}, which is equivalent to having a simple set of epistemic alternatives, as epistemic causal models have. For the remaining axioms, take any $(\tuple{\Sig, \F, \T}, \A)$. For \ax{CM}, note how \begin{inlineenum} \item $K\int{\opint{\vec{X}}{\vec{x}}}\xi$ holds at $(\tuple{\Sig, \F, \T}, \A)$ iff $\xi$ holds at $(\tuple{\Sig, \F_{\opint{\vec{X}}{\vec{x}}}, \newval{\T}{\F}{\opint{\vec{X}}{\vec{x}}}}, \newval{\A'}{\F}{\opint{\vec{X}}{\vec{x}}})$ for every $\A' \in \T$, and \item $ \int{\opint{\vec{X}}{\vec{x}}}K\xi$ holds at $(\tuple{\Sig, \F, \T}, \A)$ iff $\xi$ holds at $(\tuple{\Sig, \F_{\opint{\vec{X}}{\vec{x}}}, \newval{\T}{\F}{\opint{\vec{X}}{\vec{x}}}}, (\newval{\A}{\F}{\opint{\vec{X}}{\vec{x}}})')$ for every $(\newval{\A}{\F}{\opint{\vec{X}}{\vec{x}}})' \in \newval{\T}{\F}{\opint{\vec{X}}{\vec{x}}}$\end{inlineenum}. Then, observe that, by Definition \ref{def:int:epis}, the set of relevant valuations for the second, $\newval{\T}{\F}{\opint{\vec{X}}{\vec{x}}}$, is exactly the set of relevant valuations for the first, $\set{\newval{\A'}{\F}{\opint{\vec{X}}{\vec{x}}} \mid \A' \in \T}$. For \ax{KL}, it is enough to recall that all valuations in $\T$ comply with the same structural functions.

\myparagraph{Completeness.} The argument uses again a translation. The following rule will be useful.

\begin{lemma}\label{lem:rek}
  Let $\xi_1, \xi_2$ be \LAkc formulas.
  \begin{ctabular}{ll}
    \ax{RE_K}: & if $\vdashlokc \xi_1 \leftrightarrow \xi_2$ then $\vdashlokc K\xi_1 \leftrightarrow K\xi_2$.
  \end{ctabular}
  \begin{proof}
    Suppose $\vdashlokc \xi_1 \leftrightarrow \xi_2$. By \ax{P} and \ax{MP} we obtain $\vdashlokc \xi_1 \rightarrow \xi_2$ and $\vdashlokc \xi_2 \rightarrow \xi_1$. By rule \ax{N_K} we get $\vdashlokc K(\xi_1 \rightarrow \xi_2)$ and $\vdashlokc K(\xi_2 \rightarrow \xi_1)$. By axiom \ax{K} we obtain $\vdashlokc K\xi_1 \rightarrow K\xi_2$ and $\vdashlokc K\xi_2 \rightarrow K\xi_1$. By the tautology $(A\rightarrow B \land B\rightarrow A)\rightarrow (A \leftrightarrow B)$ and \ax{MP} we obtain $\vdashlokc K\xi_1 \leftrightarrow K\xi_2$.
  \end{proof}
\end{lemma}

\myparagraphit{Completeness, step 1: from \LAkc to \LAkcres.} Here is the translation's target language, formed by free application of Boolean and $K$ operators to atoms of the form $\int{\opint{\vec{X}}{\vec{x}}}{Y{=}y}$.

\begin{definition}[Language \LAkcres]\label{def:LAkcres}
  Formulas $\delta$ of the language \LAkcres based on the signature $\Sig$ are given by
  \[ \delta ::= \int{\opint{\vec{X}}{\vec{x}}}{Y{=}y} \mid \lnot \delta \mid \delta \land \delta \mid K\delta \]
  for $Y \in \AV$, $y \in \Ran(Y)$ and $\opint{\vec{X}}{\vec{x}}$ an assignment on $\Sig$. Given an epistemic causal model $\ec = \tuple{\Sig, \F, \T}$ and a $\A \in \T$, the semantic interpretation of the languages works as before.
\end{definition}

{\smallskip}

Here is the translation.

\begin{definition}[Translation $\trt$]\label{def:trt}
  Define $\trt:\LAkc \to \LAkcres$ as
  \begingroup
    \small
    \[
      \renewcommand{\arraystretch}{1.6}
      \begin{array}{c@{\;\quad\;}c}
        \begin{array}{@{}r@{\,:=\,}l@{}}
          \trt(Y{=}y)             & \int{\;}{Y{=}y} \\
          \trt(\lnot\xi)          & \lnot \trt(\xi) \\
          \trt(\xi_1 \land \xi_2) & \trt(\xi_1) \land \trt(\xi_2) \\
          \trt(K\xi)              & K\trt(\xi) \\
          \multicolumn{2}{@{}r}{}
        \end{array}
        &
        \begin{array}{@{}r@{\,:=\,}l@{}}
          \trt(\int{\opint{\vec{X}}{\vec{x}}}{Y{=}y})                               & \int{\opint{\vec{X}}{\vec{x}}}{Y{=}y} \\
          \trt(\int{\opint{\vec{X}}{\vec{x}}}{\lnot \xi})                           & \trt(\lnot \int{\opint{\vec{X}}{\vec{x}}}{\xi}) \\
          \trt(\int{\opint{\vec{X}}{\vec{x}}}{(\xi_1 \land \xi_2)})                 & \trt(\int{\opint{\vec{X}}{\vec{x}}}{\xi_1} \land \int{\opint{\vec{X}}{\vec{x}}}{\xi_2}) \\
          \trt(\int{\opint{\vec{X}}{\vec{x}}}{K\xi})                                & \trt(K \int{\opint{\vec{X}}{\vec{x}}}{\xi}) \\
          \trt(\int{\opint{\vec{X}}{\vec{x}}}{\int{\opint{\vec{Y}}{\vec{y}}}{\xi}}) & \trt(\int{\opint{\vec{X'}}{\vec{x'}}, \opint{\vec{Y}}{\vec{y}}}{\xi}) \\
        \end{array}
      \end{array}
    \]
  \endgroup
  where, in the bottom clause on the right, $\opint{\vec{X'}}{\vec{x'}}$ is the subassignment of $\opint{\vec{X}}{\vec{x}}$ for $\vec{X'} := \vec{X} \setminus \vec{Y}$.
\end{definition}

The translation $\trt$ works as $\tro$, taking additional care of `pushing' intervention operators inside the scope of knowledge operators (i.e., $K$ and $\int{\opint{\vec{X}}{\vec{x}}}$ commute). The following proposition contains $\trt$'s crucial properties.

\begin{proposition}\label{pro:LAktoLAkres}
  For every $\xi \in \LAkc$,
  \begin{multicols}{3}
    \begin{enumerate}
      \item\label{pro:LAktoLAkres:trans} $\trt(\xi) \in \LAkcres$,
      \item\label{pro:LAktoLAkres:syn} $\vdashlokc \xi \leftrightarrow \trt(\xi)$,
      \item\label{pro:LAktoLAkres:sem} $\models \xi \leftrightarrow \trt(\xi)$.
    \end{enumerate}
  \end{multicols}
  \begin{proof}
    The proofs of the items in this proposition rely again on an induction on the formulas' complexity $\com:\LAkc \to \Nat\setminus\set{0}$, which is this time defined as
    \begin{ctabular}{c@{\qquad}c}
      \begin{tabular}[t]{r@{\;:=\;}l}
        $\com(Y{=}y)$             & $1$ \\
        $\com(\lnot \xi)$         & $1 + \com(\xi)$ \\
        $\com(\xi_1 \land \xi_2)$ & $1 + \max\set{\com(\xi_1), \com(\xi_2)}$ \\
      \end{tabular}
      &
      \begin{tabular}[t]{r@{\;:=\;}l}
        $\com(K\xi)$                                & $1 + \com(\xi)$ \\
        $\com(\int{\opint{\vec{X}}{\vec{x}}}{\xi})$ & $2\com(\xi)$
      \end{tabular}
    \end{ctabular}
    Here is the important feature of $\com$: for every assignments $\opint{\vec{X}}{\vec{x}}$ and $\opint{\vec{Y}}{\vec{y}}$, and every formulas $\xi, \xi_1, \xi_2 \in \LAkc$,
    \begin{compactitemize}
      \item $\com(\xi) \geqslant \com(\psi)$ for every $\psi \in \sub(\xi)$. This is shown by structural induction.%
      \footnote{\textbf{Base case ($\bs{Y{=}y}$)}. Note that $\com(Y{=}y)=1$. Now take any $\psi$ in $\sub(Y{=}y) = \set{Y{=}y}$; clearly, $\com(Y{=}y) = \com(\psi)$. \textbf{Inductive case $\bs{\lnot \xi}$}, with the IH being $\com(\xi) \geqslant \com(\psi)$ for every $\psi \in \sub(\xi)$. Note that $\com(\lnot \xi)=1+\com(\xi)$ so $\com(\lnot \xi) > \com(\xi)$. Now take any $\psi$ in $\sub(\lnot \xi) = \set{\lnot \xi} \cup \sub(\xi)$: if $\psi = \lnot \xi$ then $\com(\lnot \xi) = \com(\psi)$, and if $\psi \in \sub(\xi)$ then $\com(\xi) \geqslant \com(\psi)$ (by IH) and hence $\com(\lnot \xi) > \com(\psi)$. \textbf{Inductive case $\bs{\xi_1 \land \xi_2}$}, with the IH being $\com(\xi_i) \geqslant \com(\psi)$ for every $\psi \in \sub(\xi_i)$ and $i \in \set{1,2}$. Note that $\com(\xi_1 \land \xi_2) = 1 + \max\set{\com(\xi_1), \com(\xi_2)}$ so $\com(\xi_1 \land \xi_2) > \com(\xi_i)$ for $i \in \set{1,2}$. Now take any $\psi$ in $\sub(\xi_1 \land \xi_2) = \set{\xi_1 \land \xi_2} \cup \sub(\xi_1) \cup \sub(\xi_2)$: if $\psi = \xi_1 \land \xi_2$ then $\com(\xi_1 \land \xi_2) = \com(\psi)$, and if $\psi \in \sub(\xi_i)$ then $\com(\xi_i) \geqslant \com(\psi)$ (by IH) and hence $\com(\xi_1 \land \xi_2) > \com(\psi)$. \textbf{Inductive case $\bs{K\xi}$}, with the IH being $\com(\xi) \geqslant \com(\psi)$ for every $\psi \in \sub(\xi)$. Note that $\com(K\xi)=1+\com(\xi)$ so $\com(K\xi) > \com(\xi)$. Now take any $\psi$ in $\sub(K\xi) = \set{K\xi} \cup \sub(\xi)$: if $\psi = K\xi$ then $\com(K\xi) = \com(\psi)$, and if $\psi \in \sub(\xi)$ then $\com(\xi) \geqslant \com(\psi)$ (by IH) and hence $\com(K\xi) > \com(\psi)$. \textbf{Inductive case $\bs{\int{\opint{\vec{X}}{\vec{x}}}{\xi}}$}, with the IH being $\com(\xi) \geqslant \com(\psi)$ for every $\psi \in \sub(\xi)$. Note that $\com(\int{\opint{\vec{X}}{\vec{x}}}{\xi})=2\com(\xi)$ so $\com(\int{\opint{\vec{X}}{\vec{x}}}{\xi}) > \com(\xi)$. Now take any $\psi$ in $\sub(\int{\opint{\vec{X}}{\vec{x}}}{\xi}) = \set{\int{\opint{\vec{X}}{\vec{x}}}{\xi}} \cup \sub(\xi)$: if $\psi = \int{\opint{\vec{X}}{\vec{x}}}{\xi}$ then $\com(\int{\opint{\vec{X}}{\vec{x}}}{\xi}) = \com(\psi)$, and if $\psi \in \sub(\xi)$ then $\com(\xi) \geqslant \com(\psi)$ (by IH) and hence $\com(\int{\opint{\vec{X}}{\vec{x}}}{\xi}) > \com(\psi)$.}

      \item $\com(\int{\opint{\vec{X}}{\vec{x}}}{\lnot \xi}) > \com(\lnot \int{\opint{\vec{X}}{\vec{x}}}{\xi})$.
      Indeed,
      \begin{smallltabular}{@{--\;\;}l}
        $\com(\int{\opint{\vec{X}}{\vec{x}}}{\lnot \xi}) = 2\com(\lnot \xi) = 2(1+\com(\xi)) = 2+2\com(\xi)$; \\
        $\com(\lnot \int{\opint{\vec{X}}{\vec{x}}}{\xi}) = 1+\com(\int{\opint{\vec{X}}{\vec{x}}}{\xi}) = 1+2\com(\xi)$.
      \end{smallltabular}

      \item $\com(\int{\opint{\vec{X}}{\vec{x}}}{(\xi_1 \land \xi_2)}) > \com(\int{\opint{\vec{X}}{\vec{x}}}{\xi_1} \land \int{\opint{\vec{X}}{\vec{x}}}{\xi_2})$.
      Indeed,
      \begin{smallltabular}{@{--\;\;}l@{\;=\;}l}
        $\com(\int{\opint{\vec{X}}{\vec{x}}}{(\xi_1 \land \xi_2)}) = 2\com(\xi_1 \land \xi_2)$ & $2(1 + \max\set{\com(\xi_1), \com(\xi_2)})$ \\
        \multicolumn{1}{l@{\;=\;}}{}                                                                             & $2 + 2\max\set{\com(\xi_1), \com(\xi_2)}$; \\
        $\com(\int{\opint{\vec{X}}{\vec{x}}}{\xi_1} \land \int{\opint{\vec{X}}{\vec{x}}}{\xi_2})$ & $1 + \max\set{\com(\int{\opint{\vec{X}}{\vec{x}}}{\xi_1}), \com(\int{\opint{\vec{X}}{\vec{x}}}{\xi_2})}$ \\
        \multicolumn{1}{l@{\;=\;}}{}                                                                          & $1 + \max\set{2\com(\xi_1), 2\com(\xi_2)}$ \\
        \multicolumn{1}{l@{\;=\;}}{}                                                                          & $1 + 2\max\set{\com(\xi_1), \com(\xi_2)}$.
      \end{smallltabular}

      \item $\com(\int{\opint{\vec{X}}{\vec{x}}}{K \xi}) > \com(K \int{\opint{\vec{X}}{\vec{x}}}{\xi})$.
      Indeed,
      \begin{smallltabular}{@{--\;\;}l}
        $\com(\int{\opint{\vec{X}}{\vec{x}}}{K \xi}) = 2\com(K \xi) = 2(1+\com(\xi)) = 2+2\com(\xi)$; \\
        $\com(K \int{\opint{\vec{X}}{\vec{x}}}{\xi}) = 1+\com(\int{\opint{\vec{X}}{\vec{x}}}{\xi}) = 1+2\com(\xi)$.
      \end{smallltabular}

      \item $\com(\int{\opint{\vec{X}}{\vec{x}}}{\int{\opint{\vec{Y}}{\vec{y}}}{\xi}}) > \com(\int{\opint{\vec{X'}}{\vec{x'}}, \opint{\vec{Y}}{\vec{y}}}{\xi})$.
      Indeed,
      \begin{smallltabular}{@{--\;\;}l}
        $\com(\int{\opint{\vec{X}}{\vec{x}}}{\int{\opint{\vec{Y}}{\vec{y}}}{\xi}}) = 2\com(\int{\opint{\vec{Y}}{\vec{y}}}{\xi}) = 2(2)\com(\xi) = 4\com(\xi)$; \\
        $\com(\int{\opint{\vec{X'}}{\vec{x'}}, \opint{\vec{Y}}{\vec{y}}}{\xi}) = 2\com(\xi)$.
      \end{smallltabular}
    \end{compactitemize}
    Thus, in every case of the definition of the translation, the complexity of the formulas under $\trt$ on the right-hand side is strictly smaller than the complexity of the formula under $\trt$ on the left-hand side. Thus, the application of $\trt$ to any formula $\xi \in \LAkc$ will eventually end.

    {\medskip}

    Then, the properties.
    \begin{compactenumerate}
      \item The proof is by induction on $\com(\xi)$.
      The base case is for formulas $\xi$ with $\com(\xi) = 1$; the inductive case is for formulas $\xi$ with $\com(\xi) > 1$, with the IH stating that $\trt(\psi) \in \LAkcres$ for all formulas $\psi \in \LAkc$ with $\com(\psi) < \com(\xi)$.
      \begin{compactitemize}
        \item \textbf{Case $\bs{\com(\xi)=1}$.} The only formulas in \LAkc with $\com(\xi)=1$ are \LAkc-atoms, so it should be proved that $\trt(Y{=}y) \in \LAkcres$. This is straightforward, as $\trt(Y{=}y) = \int{\;}{Y{=}y}$ is an \LAkcres-atom.

        \item \textbf{Case $\bs{\com(\xi)>1}$.} Here are the cases.
        \begin{compactitemize}
          \item \textbf{Case $\bs{\lnot \xi}$.} Since $\com(\lnot \xi) > \com(\xi)$, from IH it follows that $\trt(\xi)$ is in \LAkcres, and hence so is $\lnot \trt(\xi) = \trt(\lnot \xi)$.

          \item \textbf{Case $\bs{\xi_1 \land \xi_2}$.} Since $\com(\xi_1 \land \xi_2) > \com(\xi_i)$ for $i \in \set{1,2}$, from IH it follows that both $\trt(\xi_1)$ and $\trt(\xi_2)$ are in \LAkcres, and thus so is $\trt(\xi_1) \land \trt(\xi_2) = \trt(\xi_1 \land \xi_2)$.

          \item \textbf{Case $\bs{K \xi}$.} Since $\com(K \xi) > \com(\xi)$, from IH it follows that $\trt(\xi)$ is in \LAkcres, and hence so is $K \trt(\xi) = \trt(K \xi)$.

          \item \textbf{Case $\bs{\int{\opint{\vec{X}}{\vec{x}}}{Y{=}y}}$.} From the definition of $\trt(\int{\opint{\vec{X}}{\vec{x}}}{Y{=}y})$, the goal $\trt(\int{\opint{\vec{X}}{\vec{x}}}{Y{=}y}) \in \LAkcres$ is straightforward.

          \item \textbf{Case $\bs{\int{\opint{\vec{X}}{\vec{x}}}{\lnot \xi}}$.} Since $\com(\int{\opint{\vec{X}}{\vec{x}}}{\lnot \xi}) > \com(\lnot \int{\opint{\vec{X}}{\vec{x}}}{\xi})$, from IH it follows that $\trt(\lnot \int{\opint{\vec{X}}{\vec{x}}}{\xi}) \in \LAkcres$. By definition, the latter formula is the same as $\trt(\int{\opint{\vec{X}}{\vec{x}}}{\lnot \xi})$, which completes the case.

          \item \textbf{Case $\bs{\int{\opint{\vec{X}}{\vec{x}}}{(\xi_1 \land \xi_2)}}$.} Recall that $\com(\int{\opint{\vec{X}}{\vec{x}}}{(\xi_1 \land \xi_2)}) > \com(\int{\opint{\vec{X}}{\vec{x}}}{\xi_1} \land \int{\opint{\vec{X}}{\vec{x}}}{\xi_2})$, from IH it follows that $\trt(\int{\opint{\vec{X}}{\vec{x}}}{\xi_1} \land \int{\opint{\vec{X}}{\vec{x}}}{\xi_2}) \in \LAkcres$. By definition, the latter formula is the same as $\trt(\int{\opint{\vec{X}}{\vec{x}}}{(\xi_1 \land \xi_2)})$, which completes the case.

          \item \textbf{Case $\bs{\int{\opint{\vec{X}}{\vec{x}}}{K \xi}}$.} Since $\com(\int{\opint{\vec{X}}{\vec{x}}}{K \xi}) > \com(K \int{\opint{\vec{X}}{\vec{x}}}{\xi})$, from IH it follows that $\trt(K \int{\opint{\vec{X}}{\vec{x}}}{\xi}) \in \LAkcres$. By definition, the latter formula is the same as $\trt(\int{\opint{\vec{X}}{\vec{x}}}{K \xi})$, which completes the case.

          \item \textbf{Case $\bs{\int{\opint{\vec{X}}{\vec{x}}}{\int{\opint{\vec{Y}}{\vec{y}}}{\xi}}}$.} As $\com(\int{\opint{\vec{X}}{\vec{x}}}{\int{\opint{\vec{Y}}{\vec{y}}}{\xi}}) > \com(\int{\opint{\vec{X'}}{\vec{x'}}, \opint{\vec{Y}}{\vec{y}}}{\xi})$, from IH it follows that $\trt(\int{\opint{\vec{X'}}{\vec{x'}}, \opint{\vec{Y}}{\vec{y}}}{\xi}) \in \LAkcres$. By definition, the latter formula is $\trt(\int{\opint{\vec{X}}{\vec{x}}}{\int{\opint{\vec{Y}}{\vec{y}}}{\xi}})$, which completes the case.
        \end{compactitemize}
      \end{compactitemize}

      \item  The proof is by induction on $\com(\xi)$.
      The base case is for formulas $\xi$ with $\com(\xi) = 1$; the inductive case is for formulas $\xi$ with $\com(\xi) > 1$, with the IH stating that $\vdashlokc \psi \leftrightarrow \trt(\psi)$ holds for all formulas $\psi \in \LAkc$ with $\com(\psi) < \com(\xi)$.
      \begin{compactitemize}
        \item \textbf{Case $\bs{\com(\xi)=1}$.} The only formulas in \LAkc with $\com(\xi)=1$ are \LAkc-atoms, so it should be proved that $\vdashlokc Y{=}y \leftrightarrow \trt(Y{=}y)$, i.e., that $\vdashlokc Y{=}y \leftrightarrow \int{\;}{Y{=}y}$. This follows immediately from the axiom \ax{A_{[]}}.

        \item \textbf{Case $\bs{\com(\xi)>1}$.} Here is where all non-\LAc-atomic formulas fall. Here are the cases.
        \begin{compactitemize}
          \item \textbf{Case $\bs{\lnot \xi}$.} Since $\com(\lnot \xi) > \com(\xi)$, from IH it follows that $\vdashlokc \xi \leftrightarrow \trt(\xi)$. Then $\vdashlokc \lnot \xi \leftrightarrow \lnot \trt(\xi)$ (propositional reasoning), and thus the definition of $\trt$ yields the required $\vdashlokc \lnot \xi \leftrightarrow \trt(\lnot \xi)$.

          \item \textbf{Case $\bs{\xi_1 \land \xi_2}$.} Since $\com(\xi_1 \land \xi_2) > \com(\xi_i)$ for $i \in \set{1,2}$, from IH it follows that $\vdashlokc \xi_i \leftrightarrow \trt(\xi_i)$. Then $\vdashlokc (\xi_1 \land \xi_2) \leftrightarrow (\trt(\xi_1) \land \trt(\xi_2))$ (propositional reasoning), and thus the definition of $\trt$ yields the required $\vdashlokc (\xi_1 \land \xi_2) \leftrightarrow \trt(\xi_1 \land \xi_2)$.

          \item \textbf{Case $\bs{K \xi}$.} Since $\com(K \xi) > \com(\xi)$, from IH it follows that $\vdashlokc \xi \leftrightarrow \trt(\xi)$. Then $\vdashlokc K \xi \leftrightarrow K \trt(\xi)$ (by \ax{RE_K} on Lemma \ref{lem:rek}), and thus the definition of $\trt$ yields the required $\vdashlokc K \xi \leftrightarrow \trt(K \xi)$.

          \item \textbf{Case $\bs{\int{\opint{\vec{X}}{\vec{x}}}{Y{=}y}}$.} From the definition of $\trt(\int{\opint{\vec{X}}{\vec{x}}}{Y{=}y})$, the goal $\vdashlokc \int{\opint{\vec{X}}{\vec{x}}}{Y{=}y} \leftrightarrow \trt(\int{\opint{\vec{X}}{\vec{x}}}{Y{=}y})$ is straightforward.

          \item \textbf{Case $\bs{\int{\opint{\vec{X}}{\vec{x}}}{\lnot \xi}}$.} Since $\com(\int{\opint{\vec{X}}{\vec{x}}}{\lnot \xi}) > \com(\lnot \int{\opint{\vec{X}}{\vec{x}}}{\xi})$, from IH it follows that $\vdashlokc \lnot \int{\opint{\vec{X}}{\vec{x}}}{\xi} \leftrightarrow \trt(\lnot \int{\opint{\vec{X}}{\vec{x}}}{\xi})$. But axiom \ax{A_\lnot} gives us $\vdashlokc \int{\opint{\vec{X}}{\vec{x}}}{\lnot \xi} \leftrightarrow \lnot \int{\opint{\vec{X}}{\vec{x}}}{\xi}$, so $\vdashlokc \int{\opint{\vec{X}}{\vec{x}}}{\lnot \xi} \leftrightarrow \trt(\lnot \int{\opint{\vec{X}}{\vec{x}}}{\xi})$ and hence, from $\trt$'s definition, $\vdashlokc \int{\opint{\vec{X}}{\vec{x}}}{\lnot \xi} \leftrightarrow \trt(\int{\opint{\vec{X}}{\vec{x}}}{\lnot \xi})$.

          \item \textbf{Case $\bs{\int{\opint{\vec{X}}{\vec{x}}}{(\xi_1 \land \xi_2)}}$.} Recall that $\com(\int{\opint{\vec{X}}{\vec{x}}}{(\xi_1 \land \xi_2)}) > \com(\int{\opint{\vec{X}}{\vec{x}}}{\xi_1} \land \int{\opint{\vec{X}}{\vec{x}}}{\xi_2})$, IH gives us $\vdashlokc (\int{\opint{\vec{X}}{\vec{x}}}{\xi_1} \land \int{\opint{\vec{X}}{\vec{x}}}{\xi_2}) \leftrightarrow \trt(\int{\opint{\vec{X}}{\vec{x}}}{\xi_1} \land \int{\opint{\vec{X}}{\vec{x}}}{\xi_2})$. But, by axiom \ax{A_\land}, $\vdashlokc \int{\opint{\vec{X}}{\vec{x}}}{(\xi_1 \land \xi_2)} \leftrightarrow (\int{\opint{\vec{X}}{\vec{x}}}{\xi_1} \land \int{\opint{\vec{X}}{\vec{x}}}{\xi_2})$; then, it follows that $\vdashlokc \int{\opint{\vec{X}}{\vec{x}}}{(\xi_1 \land \xi_2)} \leftrightarrow \trt(\int{\opint{\vec{X}}{\vec{x}}}{\xi_1} \land \int{\opint{\vec{X}}{\vec{x}}}{\xi_2})$, and hence the definition of $\trt$ yields the required $\vdashlokc \int{\opint{\vec{X}}{\vec{x}}}{(\xi_1 \land \xi_2)} \leftrightarrow \trt(\int{\opint{\vec{X}}{\vec{x}}}{(\xi_1 \land \xi_2)})$.

          \item \textbf{Case $\bs{\int{\opint{\vec{X}}{\vec{x}}}{K \xi}}$.} Since $\com(\int{\opint{\vec{X}}{\vec{x}}}{K \xi}) > \com(K \int{\opint{\vec{X}}{\vec{x}}}{\xi})$, from IH it follows that $\vdashlokc K \int{\opint{\vec{X}}{\vec{x}}}{\xi} \leftrightarrow \trt(K \int{\opint{\vec{X}}{\vec{x}}}{\xi})$. But axiom \ax{CM} gives us $\vdashlokc \int{\opint{\vec{X}}{\vec{x}}}{K \xi} \leftrightarrow K \int{\opint{\vec{X}}{\vec{x}}}{\xi}$, so $\vdashlokc \int{\opint{\vec{X}}{\vec{x}}}{K \xi} \leftrightarrow \trt(K \int{\opint{\vec{X}}{\vec{x}}}{\xi})$ and hence, from $\trt$'s definition, $\vdashlokc \int{\opint{\vec{X}}{\vec{x}}}{K \xi} \leftrightarrow \trt(\int{\opint{\vec{X}}{\vec{x}}}{K \xi})$.

          \item \textbf{Case $\bs{\int{\opint{\vec{X}}{\vec{x}}}{\int{\opint{\vec{Y}}{\vec{y}}}{\xi}}}$.} As $\com(\int{\opint{\vec{X}}{\vec{x}}}{\int{\opint{\vec{Y}}{\vec{y}}}{\xi}}) > \com(\int{\opint{\vec{X'}}{\vec{x'}}, \opint{\vec{Y}}{\vec{y}}}{\xi})$, from IH it follows that $\vdashlokc \int{\opint{\vec{X'}}{\vec{x'}}, \opint{\vec{Y}}{\vec{y}}}{\xi} \leftrightarrow \trt(\int{\opint{\vec{X'}}{\vec{x'}}, \opint{\vec{Y}}{\vec{y}}}{\xi}))$. But $\vdashlokc \int{\opint{\vec{X}}{\vec{x}}}{\int{\opint{\vec{Y}}{\vec{y}}}{\xi}} \leftrightarrow \int{\opint{\vec{X'}}{\vec{x'}}, \opint{\vec{Y}}{\vec{y}}}{\xi}$ (from axiom \ax{A_{[][]}}), so $\vdashlokc \int{\opint{\vec{X}}{\vec{x}}}{\int{\opint{\vec{Y}}{\vec{y}}}{\xi}} \leftrightarrow \trt(\int{\opint{\vec{X'}}{\vec{x'}}, \opint{\vec{Y}}{\vec{y}}}{\xi})$; hence, the definition of $\trt$ yields the required $\vdashlokc \int{\opint{\vec{X}}{\vec{x}}}{\int{\opint{\vec{Y}}{\vec{y}}}{\xi}} \leftrightarrow \trt(\int{\opint{\vec{X}}{\vec{x}}}{\int{\opint{\vec{Y}}{\vec{y}}}{\xi}})$.
        \end{compactitemize}
      \end{compactitemize}

      \item By the previous item, $\vdashlokc \xi \leftrightarrow \trt(\xi)$. But \LOkc is sound within recursive causal models; therefore, $\models \xi \leftrightarrow \trt(\xi)$.
    \end{compactenumerate}
  \end{proof}
\end{proposition}

\myparagraphit{Completeness, step 2: a canonical model for \LAkcres.} Having shown that every formula in \LAkc is both semantically and \LOkc-provably equivalent to a formula in \LAkcres, now it will be shown that the axiom system \LOkcres, the fragment of \LOkc without the axioms in the third part of Table \ref{tbl:rcm-LA} (axioms \ax{A_{[]}}, \ax{A_\lnot}, \ax{A_\land}, \ax{A_{[][]}}) and without axiom \ax{CM}, is complete for \LAkcres over epistemic (recursive) causal models. This will be done by showing, via the construction of a canonical model, that any \LOkcres-consistent set of \LAkcres-formulas is satisfiable in an epistemic (recursive) causal model. The construction uses the construction of a recursive causal model of Definition \ref{pro:rcm} to build an \emph{epistemic} recursive causal model (following the ideas in \citealp{RAK1995}), taking additional care of guaranteeing that all valuations comply with the same set of structural functions.

{\smallskip}

Let $\mcssk$ be the set of all maximally \LOkcres-consistent sets of \LAkcres-formulas; take $\Delta \in \mcssk$. Note that, while the language \LAkcres extends the language \LAcres with a free use of the epistemic modality $K$, the system \LOkcres extends \LOcres with rule \ax{N_K} and axioms \ax{K}, \ax{T}, \ax{4}, \ax{5} and \ax{KL}. Hence, $\Delta\vert_{\LAcres}$ (the restriction of $\Delta$ to formulas in \LAcres) is a \LOcres-consistent set of \LAcres-formulas and thus an element of $\mcss$. Then, each such $\Delta$ defines a set of structural functions for $\NV$, namely $\F^{\Delta\vert_{\LAcres}}$ (Definition \ref{def:cm}).

\begin{definition}[Building an epistemic causal model]\label{def:ecm}
  Take $\Delta \in \mcssk$.
  \begin{itemize}
    \item Define the set $\mcssk^\Delta \subseteq \mcssk$ as $\mcssk^\Delta := \set{ \Psi \in \mcssk \mid \F^{\Delta\vert_{\LAcres}} = \F^{\Psi\vert_{\LAcres}}}$, so it contains the maximally consistent sets in $\mcssk$ whose structural functions coincide with those of $\Delta$. Obviously, $\Delta \in \mcssk^\Delta$.

    \item Define $R^\Delta \subseteq \mcssk^\Delta \times \mcssk^\Delta$ as $(\Psi_1, \Psi_2) \in R^\Delta$ if and only if $K\delta \in \Psi_1$ implies $\delta \in \Psi_2$ for every $\delta \in \LAkcres$. This is the standard definition of the relation in modal canonical models (see, e.g., \citealp{RAK1995,BlackburnRijkeVenema2001}). The elements of $\mcss^\Delta$ are maximally \LOkcres-consistent sets, and \LOkcres includes axioms \ax{T}, \ax{4} and \ax{5}; thus, it follows from standard modal results (see, e.g., the just mentioned references) that $R^\Delta$ is an equivalence relation. In particular, axiom \ax{T} implies $(\Delta, \Delta) \in R^\Delta$.

    \item Define $\T^{\Delta} := \set{ \A^{\Psi\vert_{\LAcres}} \mid (\Delta, \Psi) \in R^\Delta}$, so the set contains the valuation function (see Definition \ref{def:cm}) of each maximally consistent set $\Psi \in \mcss^\Delta$ that is $R^\Delta$-reachable from $\Delta$. In particular, $(\Delta, \Delta) \in R^\Delta$ implies $\A^{\Delta\vert_{\LAcres}} \in \T^\Delta$.
  \end{itemize}
  The structure $\ec^\Delta$ is given by $\tuple{\Sig, \F^{\Delta\vert_{\LAcres}}, \T^\Delta}$.
\end{definition}

Now it will be shown that $\ec^\Delta$ satisfies an appropriate truth lemma.

\begin{lemma}[Truth Lemma, \LAcres]\label{lem:LAkres:truth-lemma}
  Take $\Delta \in \mcssk$; recall that $\A^{\Delta\vert_{\LAcres}} \in \T^\Delta$. Then, for every $\delta \in \LAkcres$,
  \begin{ctabular}{c}
    $(\ec^\Delta, \A^{\Delta\vert_{\LAcres}}) \models \delta
    \quad\text{if and only if}\quad
    \delta \in \Delta$
  \end{ctabular}
  \begin{proof}
    The proof is by structural induction on $\delta$.
    \begin{inductionproof}
      \item [Case $\bs{\int{\opint{\vec{X}}{\vec{x}}}Z{=}z}$.] The formula's truth-value at $(\tuple{\Sig, \F^{\Delta\vert_{\LAcres}}, \T^\Delta}, \A^{\Delta\vert_{\LAcres}})$ is independent from the set of valuations $\T^\Delta$. Thus,
      \begingroup
        \small
          \[
            (\tuple{\Sig, \F^{\Delta\vert_{\LAcres}}, \T^\Delta}, \A^{\Delta\vert_{\LAcres}}) \models \int{\opint{\vec{X}}{\vec{x}}}Z{=}z
            \quad\text{iff}\quad
            \tuple{\Sig, \F^{\Delta\vert_{\LAcres}}, \A^{\Delta\vert_{\LAcres}}} \models \int{\opint{\vec{X}}{\vec{x}}}Z{=}z
          \]
      \endgroup
      By Proposition \ref{pro:LAres:truth-lemma:atoms}, the right-hand side is equivalent to $\int{\opint{\vec{X}}{\vec{x}}}Z{=}z \in \Delta\vert_{\LAcres}$, that is, to $\int{\opint{\vec{X}}{\vec{x}}}Z{=}z \in \Delta$.

      \item [Cases $\bs{\lnot}$ and $\bs{\land}$.] Immediate from the inductive hypotheses and the properties of a maximally consistent set.

      \item[Case $\bs{K}$] {$\bs{(\Rightarrow)}$} Suppose $(\ec^\Delta, \A^{\Delta\vert_{\LAcres}}) \models K\delta$. Define the set
      \[ \Psi^- := \set{\psi \in \LAkcres \mid K\psi \in \Delta} \cup \set{\lnot \delta} \]
      and, for the sake of a contradiction, suppose it is \LOkcres-consistent. Then, $\Psi^-$ could be extended into a maximally \LOkcres-consistent set $\Psi \in \mcssk$. Note how, by axiom \ax{KL} and modus ponens, each one of the formulas in $\Delta$ that define $\Delta$'s set of structural functions (those of the form $\int{\opint{\vec{X}}{\vec{x}}}Y{=}y$ for $Y\in \NV$ and $\vec{X} = \AV \setminus \set{Y}$) is in $\Psi$;\footnote{More precisely, $\int{\opint{\vec{X}}{\vec{x}}}Y{=}y \in \Delta$ for $Y\in \NV$ and $\vec{X} = \AV \setminus \set{Y}$ implies (axiom \ax{KL} and modus ponens) $K\int{\opint{\vec{X}}{\vec{x}}}Y{=}y \in \Delta$, so $\int{\opint{\vec{X}}{\vec{x}}}Y{=}y$ is in $\Psi^-$ and thus in $\Psi$.} thus, $\F^{\Psi\vert_{\LAcres}} = \F^{\Delta\vert_{\LAcres}}$, so $\Psi \in \mcssk^\Delta$. From this and the construction of $\Psi^-$, it follows that $R^\Delta \Delta \Psi$, so $\A^{\Psi\vert_{\LAcres}} \in \T^\Delta$. Finally, $\lnot \delta \in \Psi$ so $\delta \not\in \Psi$ and hence, by IH, $(\ec^\Delta, \A^{\Psi\vert_{\LAcres}}) \not\models \delta$, that is, $(\ec^\Delta, \A^{\Psi\vert_{\LAcres}}) \models \lnot \delta$. But then $\A^{\Psi\vert_{\LAcres}}$ is such that $\A^{\Psi\vert_{\LAcres}} \in \T^\Delta$ and yet $(\ec^\Delta, \A^{\Psi\vert_{\LAcres}}) \models \lnot \delta$, contradicting the initial assumption $(\ec^\Delta, \A^{\Delta\vert_{\LAcres}}) \models K\delta$.

      Thus, $\Psi^-$ is \LOkcres-inconsistent, so there are $\set{\psi_1, \ldots, \psi_m} \subseteq \Psi^-$ such that
      \[
        \vdashlokcres (\psi_1 \land \cdots \land \psi_m \land \lnot \delta) \rightarrow \bot,
        \qquad\text{that is,}\qquad
        \vdashlokcres (\psi_1 \land \cdots \land \psi_m) \rightarrow \delta.
      \]
      Now, by \ax{N_K},
      \[ \vdashlokcres K\big((\psi_1 \land \cdots \land \psi_m) \rightarrow \delta\big) \]
      and, by \ax{K},
      \[
        \vdashlokcres
          K\big((\psi_1 \land \cdots \land \psi_m) \rightarrow \delta\big)
          \rightarrow
          \big( K(\psi_1 \land \cdots \land \psi_m) \rightarrow K\delta \big).
      \]
      Hence, by modus ponens,
      \[ \vdashlokcres K(\psi_1 \land \cdots \land \psi_m) \rightarrow K\delta, \]
      that is (using that $\vdashlokcres K(\delta_1 \land \delta_2) \leftrightarrow (K\delta_1 \land K\delta_2)$),
      \[ \vdashlokcres (K\psi_1 \land \cdots \land K\psi_m) \rightarrow K\delta. \]
      Finally, recall that $\psi_i \in \set{\psi \in \LAkcres \mid K\psi \in \Delta}$. Hence,
      \[
       \set{ K\psi_1, \ldots, K\psi_m} \subseteq \Delta,
       \qquad\text{that is,}\qquad
       K\psi_1 \land \cdots \land K\psi_m \in \Delta
      \]
      and therefore we obtain the required $K\delta \in \Delta$.

      {\smallskip}

      {$\bs{(\Leftarrow)}$} Suppose $K\delta \in \Delta$, and take any $\A^{\Psi\vert_{\LAcres}} \in \T^\Delta$, so $R^\Delta \Delta \Psi$. From the latter and $K\delta \in \Delta$ it follows that $\delta \in \Psi$. By induction hypothesis on the latter, $(\ec^\Delta, \A^{\Psi\vert_{\LAcres}}) \models \delta$. Hence, $(\ec^\Delta, \A^{\Delta\vert_{\LAcres}}) \models K\delta$.
    \end{inductionproof}
  \end{proof}
\end{lemma}

It is only left to check that $\ec^\Delta$ is indeed an epistemic causal model.

\begin{proposition}\label{pro:ercm}
  Take $\Delta \in \mcss$. The structure $\ec^\Delta = \tuple{\Sig, \F^{\Delta\vert_{\LAcres}}, \T^\Delta}$ is such that every valuation in $\T^\Delta$ complies with $\F^{\Delta\vert_{\LAcres}}$, which is a recursive set of structural equations.
  \begin{proof}
    Each $\A^{\Psi\vert_{\LAcres}} \in \T^\Gamma$ complies with \emph{its} set of structural functions $\F^{\Psi\vert_{\LAcres}}$ (second item in Proposition \ref{pro:rcm}). But $\A^{\Psi\vert_{\LAcres}} \in \T^\Gamma$, so $\Psi \in \mcss^\Delta$, which implies $\F^{\Psi\vert_{\LAcres}} = \F^{\Delta\vert_{\LAcres}}$. Thus, $\A^{\Psi\vert_{\LAcres}}$ complies with $\F^{\Delta\vert_{\LAcres}}$, which it has been already proved to be recursive (second item in Proposition \ref{pro:rcm}.
  \end{proof}
\end{proposition}

Then, the full argument for completeness of \LOkcres for formulas in \LAkcres.

\begin{theorem}[Completeness of \LOkcres for \LAkcres]\label{teo:completeness-lokres}
  The axiom system \LOkcres is strongly complete for the language \LAkcres based on $\Sig$ with respect to epistemic (recursive) causal models for $\Sig$. In other words, for every set of \LAkcres-formulas $\Delta^- \cup \set{\delta}$, if $\Delta^- \models \delta$ then $\Delta^- \vdashlokcres \delta$.
  \begin{proof}
    By contraposition; take any \LOkcres-consistent set of formulas $\Delta^-$. Since \LAcres is enumerable, $\Delta^-$ can be extended into a maximally $\LOkcres$-consistent set $\Delta$ via a standard Lindenbaum's construction (see, e.g., \citealp[Chapter 4]{BlackburnRijkeVenema2001}). By Lemma \ref{lem:LAkres:truth-lemma}, all formulas in $\Delta$ are satisfiable in $(\ec^\Delta, \A^{\Delta\vert_{\LAcres}})$, which by Proposition \ref{pro:ercm} is an epistemic (recursive) causal model. Thus, $\Delta^-$ is satisfiable in an epistemic (recursive) causal model.
  \end{proof}
\end{theorem}

{\medskip}

Finally, the argument for strong completeness of \LOkc for formulas in \LAkc\label{proof:thm:rcm-LAk} is analogous to that in the proof of Theorem \ref{thm:rcm-LA} (Page \pageref{proof:thm:rcm-LA}). Take $\Psi \cup \set{\xi} \subseteq \LAkc$ and suppose $\Psi \models \xi$. By Proposition \ref{pro:LAktoLAkres}.\ref{pro:LAktoLAkres:sem}, it follows that $\trt(\Psi) \models \trt(\xi)$. But, by Proposition \ref{pro:LAktoLAkres}.\ref{pro:LAktoLAkres:trans}, $\trt(\Psi) \cup \set{\trt(\xi)} \subseteq \LAkcres$. Hence, Theorem \ref{teo:completeness-lokres} implies $\trt(\Psi) \vdashlokcres \trt(\varphi)$ and thus $\trt(\Psi) \vdashlokc \trt(\xi)$ (as \LOkcres is a subsystem of \LOkc). Then, by Proposition \ref{pro:LAktoLAkres}.\ref{pro:LAktoLAkres:syn}, the required $\Psi \vdashlokc \xi$ follows.

\end{appendix:full}

\begin{appendix:short}
\myparagraph{Soundness.} Axioms and rules in Table \ref{tbl:rcm-LA} do not involve the $K$ operator, so their truth-value in a pointed epistemic causal model $(\tuple{\Sig, \F, \T}, \A)$ only depends on the causal model $\tuple{\Sig, \F, \A}$, for which they are sound (Theorem \ref{thm:rcm-LA}).

For Table \ref{tbl:rcm-LAk}, axioms and rules in the \emph{epistemic} part are sound on relational structures with a single equivalence relation \citep{RAK1995,BlackburnRijkeVenema2001}, which is equivalent to having a simple set of epistemic alternatives, as epistemic causal models have. For \ax{CM} recall: \begin{inlineenum} \item $K\int{\opint{\vec{X}}{\vec{x}}}\xi$ holds at $(\tuple{\Sig, \F, \T}, \A)$ iff $\xi$ holds at $(\tuple{\Sig, \F_{\opint{\vec{X}}{\vec{x}}}, \newval{\T}{\F}{\opint{\vec{X}}{\vec{x}}}}, \newval{\A'}{\F}{\opint{\vec{X}}{\vec{x}}})$ for every $\A' \in \T$, and \item $ \int{\opint{\vec{X}}{\vec{x}}}K\xi$ holds at $(\tuple{\Sig, \F, \T}, \A)$ iff $\xi$ holds at $(\tuple{\Sig, \F_{\opint{\vec{X}}{\vec{x}}}, \newval{\T}{\F}{\opint{\vec{X}}{\vec{x}}}}, (\newval{\A}{\F}{\opint{\vec{X}}{\vec{x}}})')$ for every $(\newval{\A}{\F}{\opint{\vec{X}}{\vec{x}}})' \in \newval{\T}{\F}{\opint{\vec{X}}{\vec{x}}}$\end{inlineenum}. Then note how, by Definition \ref{def:int:epis}, the set of relevant valuations for the first, $\set{\newval{\A'}{\F}{\opint{\vec{X}}{\vec{x}}} \mid \A' \in \T}$, is exactly as that for the second, $\newval{\T}{\F}{\opint{\vec{X}}{\vec{x}}}$. For \ax{KL} recall: all valuations in $\T$ comply with the same structural functions.

\myparagraph{Completeness.} Again, by translation, now also using the following rule.

\begin{lemma}\label{lem:rek}
  Let $\xi_1, \xi_2$ be \LAkc formulas.
  \begin{ctabular}{ll}
    \ax{RE_K}: & if $\vdashlokc \xi_1 \leftrightarrow \xi_2$ then $\vdashlokc K\xi_1 \leftrightarrow K\xi_2$.
  \end{ctabular}
\end{lemma}

\myparagraphit{Completeness, step 1: from \LAkc to \LAkcres.} Formulas $\delta$ in the language $\LAkcres$ are built by using Boolean and $K$ operators on `atoms' of the form $\int{\opint{\vec{X}}{\vec{x}}}{Y{=}y}$. They are semantically interpreted just as formulas in \LAkc.

{\smallskip}

Here is the translation.

\begin{definition}[Translation $\trt$]\label{def:trt}
  Define $\trt:\LAkc \to \LAkcres$ as
  \begingroup
    \small
    \[
      \renewcommand{\arraystretch}{1.6}
      \begin{array}{c@{\;\quad\;}c}
        \begin{array}{@{}r@{\,:=\,}l@{}}
          \trt(Y{=}y)             & \int{\;}{Y{=}y} \\
          \trt(\lnot\xi)          & \lnot \trt(\xi) \\
          \trt(\xi_1 \land \xi_2) & \trt(\xi_1) \land \trt(\xi_2) \\
          \trt(K\xi)              & K\trt(\xi) \\
          \multicolumn{2}{@{}r}{}
        \end{array}
        &
        \begin{array}{@{}r@{\,:=\,}l@{}}
          \trt(\int{\opint{\vec{X}}{\vec{x}}}{Y{=}y})                               & \int{\opint{\vec{X}}{\vec{x}}}{Y{=}y} \\
          \trt(\int{\opint{\vec{X}}{\vec{x}}}{\lnot \xi})                           & \trt(\lnot \int{\opint{\vec{X}}{\vec{x}}}{\xi}) \\
          \trt(\int{\opint{\vec{X}}{\vec{x}}}{(\xi_1 \land \xi_2)})                 & \trt(\int{\opint{\vec{X}}{\vec{x}}}{\xi_1} \land \int{\opint{\vec{X}}{\vec{x}}}{\xi_2}) \\
          \trt(\int{\opint{\vec{X}}{\vec{x}}}{K\xi})                                & \trt(K \int{\opint{\vec{X}}{\vec{x}}}{\xi}) \\
          \trt(\int{\opint{\vec{X}}{\vec{x}}}{\int{\opint{\vec{Y}}{\vec{y}}}{\xi}}) & \trt(\int{\opint{\vec{X'}}{\vec{x'}}, \opint{\vec{Y}}{\vec{y}}}{\xi}) \\
        \end{array}
      \end{array}
    \]
  \endgroup
  with $\opint{\vec{X'}}{\vec{x'}}$ the subassignment of $\opint{\vec{X}}{\vec{x}}$ for $\vec{X'} := \vec{X} \setminus \vec{Y}$.
\end{definition}

\noindent The following proposition contains $\trt$'s crucial properties.

\begin{proposition}\label{pro:LAktoLAkres}
  For every $\xi \in \LAkc$,
  \begin{multicols}{3}
    \begin{enumerate}
      \item\label{pro:LAktoLAkres:trans} $\trt(\xi) \in \LAkcres$,
      \item\label{pro:LAktoLAkres:syn} $\vdashlokc \xi \leftrightarrow \trt(\xi)$,
      \item\label{pro:LAktoLAkres:sem} $\models \xi \leftrightarrow \trt(\xi)$.
    \end{enumerate}
  \end{multicols}
  \begin{proof}
    The proofs rely again on the formulas' complexity $\com:\LAkc \to \Nat\setminus\set{0}$, this time defined as
    \begin{smallctabular}{c@{\qquad}c}
      \begin{tabular}[t]{r@{\;:=\;}l}
        $\com(Y{=}y)$             & $1$ \\
        $\com(\lnot \xi)$         & $1 + \com(\xi)$ \\
        $\com(\xi_1 \land \xi_2)$ & $1 + \max\set{\com(\xi_1), \com(\xi_2)}$ \\
      \end{tabular}
      &
      \begin{tabular}[t]{r@{\;:=\;}l}
        $\com(K\xi)$                                & $1 + \com(\xi)$ \\
        $\com(\int{\opint{\vec{X}}{\vec{x}}}{\xi})$ & $2\com(\xi)$
      \end{tabular}
    \end{smallctabular}
    Note: for every assignments $\opint{\vec{X}}{\vec{x}}$ and $\opint{\vec{Y}}{\vec{y}}$, and every $\xi, \xi_1, \xi_2 \in \LAkc$,
    \begingroup
      \small
      \begin{center}
        $\begin{array}{r@{\;}c@{\;}l}
          \toprule
          \com(\xi) & \geqslant & \com(\psi) \;\text{for every}\; \psi \in \sub(\xi) \\
          \midrule
          \com(\int{\opint{\vec{X}}{\vec{x}}}{\lnot \xi}) & > & \com(\lnot \int{\opint{\vec{X}}{\vec{x}}}{\xi}) \\
          \com(\int{\opint{\vec{X}}{\vec{x}}}{(\xi_1 \land \xi_2)}) & > & \com(\int{\opint{\vec{X}}{\vec{x}}}{\xi_1} \land \int{\opint{\vec{X}}{\vec{x}}}{\xi_2}) \\
          \com(\int{\opint{\vec{X}}{\vec{x}}}{K \xi}) & > & \com(K \int{\opint{\vec{X}}{\vec{x}}}{\xi}) \\
          \com(\int{\opint{\vec{X}}{\vec{x}}}{\int{\opint{\vec{Y}}{\vec{y}}}{\xi}}) & > & \com(\int{\opint{\vec{X'}}{\vec{x'}}, \opint{\vec{Y}}{\vec{y}}}{\xi}) \\
          \bottomrule
        \end{array}$
      \end{center}
    \endgroup
    \noindent Thus, $\trt$ is a proper recursive translation. The proof of first two properties use induction on $\com(\xi)$. The base case is for formulas $\xi$ with $\com(\xi) = 1$ ($Y{=}y$); the inductive case is for formulas $\xi$ with $\com(\xi) > 1$ ($\lnot \xi, \xi_1 \land \xi_2, K\xi$, $\int{\opint{\vec{X}}{\vec{x}}}{Y{=}y}, \int{\opint{\vec{X}}{\vec{x}}}{\lnot \xi}, \int{\opint{\vec{X}}{\vec{x}}}{(\xi_1 \land \xi_2)}$, $\int{\opint{\vec{X}}{\vec{x}}}{K \xi}$ and $\int{\opint{\vec{X}}{\vec{x}}}{\int{\opint{\vec{Y}}{\vec{y}}}{\xi}}$).
    \begin{compactenumerate}
      \item The IH states that $\trt(\psi) \in \LAkcres$ for all formulas $\psi \in \LAkc$ with $\com(\psi) < \com(\xi)$. All cases are straightforward from $\trt$'s definition and the respective IH.

      \item The IH states that $\vdashlokc \psi \leftrightarrow \trt(\psi)$ holds for all formulas $\psi \in \LAkc$ with $\com(\psi) < \com(\xi)$. All cases follow from $\trt$'s definition, the IH, propositional reasoning and the axiom system (\ax{A_{[]}} for formulas of the form $Y{=}y$, \ax{RE_K} on Lemma \ref{lem:rek} for formulas of the form $K\xi$, \ax{A_\lnot} for formulas of the form $\int{\opint{\vec{X}}{\vec{x}}}{\lnot \xi}$, \ax{A_\land} for formulas of the form $\int{\opint{\vec{X}}{\vec{x}}}{(\xi_1 \land \xi_2)}$, \ax{CM} for formulas of the form $\int{\opint{\vec{X}}{\vec{x}}}{K \xi}$, \ax{A_{[][]}} for formulas of the form $\int{\opint{\vec{X}}{\vec{x}}}{\int{\opint{\vec{Y}}{\vec{y}}}{\xi}}$).

      \item By the previous item, $\vdashlokc \xi \leftrightarrow \trt(\xi)$. But \LOkc is sound within recursive causal models; therefore, $\models \xi \leftrightarrow \trt(\xi)$.
    \end{compactenumerate}
  \end{proof}
\end{proposition}

\myparagraphit{Completeness, step 2: a canonical model for \LAkcres.} Next it is shown that \LOkcres, the fragment of \LOkc without axioms \ax{A_{[]}}, \ax{A_\lnot}, \ax{A_\land}, \ax{A_{[][]}} and \ax{CM}, is complete for \LAkcres over epistemic (recursive) causal models. This is done by showing, via the construction of a canonical model, that any \LOkcres-consistent set of \LAkcres-formulas is satisfiable in an epistemic (recursive) causal model. The construction uses the ideas of \citet{halpern2000axiomatizing} for building a recursive causal model, and then ideas from \citet{RAK1995} to extend it to a recursive epistemic causal model, taking additional care of guaranteeing that all valuations comply with the same set of structural functions.

{\smallskip}

Let $\mcssk$ be the set of all maximally \LOkcres-consistent sets of \LAkcres-formulas; take $\Delta \in \mcssk$. Since \LAkcres extends \LAcres and \LOkcres extends \LOcres, the set $\Delta\vert_{\LAcres}$ (the restriction of $\Delta$ to formulas in \LAcres) is a maximally \LOcres-consistent set of \LAcres-formulas. Then (completeness proof of Theorem~\ref{thm:rcm-LA}), each such $\Delta$ defines a set of structural functions for $\NV$, namely $\F^{\Delta\vert_{\LAcres}}$. This is the basis for defining an epistemic causal model $\ec^\Delta = \tuple{\Sig, \F^{\Delta\vert_{\LAcres}}, \T^\Delta}$, where the valuations in $\T^\Delta$ are built as follows. First, define the set $\mcssk^\Delta \subseteq \mcssk$ containing the elements of $\mcssk$ whose structural functions coincide with those of $\Delta$ (obviously, $\Delta \in \mcssk^\Delta$). Then define $\T^{\Delta}$ as those sets in $\mcssk^\Delta$ that fit the epistemic information in $\Delta$.\footnote{This is done in two stages. First, define an epistemic indistinguishability relation $R^\Delta \subseteq \mcssk^\Delta \times \mcssk^\Delta$ as typically done in modal canonical models (see, e.g., \citealp{RAK1995,BlackburnRijkeVenema2001}). Due to axioms \ax{T}, \ax{4} and \ax{5}, the relation $R^\Delta$ is an equivalence relation (in particular, $(\Delta, \Delta) \in R^\Delta$). Then, $\T^{\Delta}$ contains the valuation function induced by each $\Psi \in \mcss^\Delta$ satisfying $R^\Delta\Delta\Psi$ (since $(\Delta, \Delta) \in R^\Delta$, we have $\A^{\Delta\vert_{\LAcres}} \in \T^\Delta$).}

\smallskip

It can be shown, by structural induction on $\delta \in \LAkcres$, that $\ec^\Delta$ satisfies a truth lemma. The base case (formulas of the form $\int{\opint{\vec{X}}{\vec{x}}}Z{=}z$) is the base case of the truth lemma for Theorem~\ref{thm:rcm-LA}; the Boolean cases use their IH. For $K \xi$, the proof goes as in standard epistemic logic, using the fact that, by \ax{KL}, the formulas in $\Delta$ defining $\F^{\Delta\vert_{\LAcres}}$ are also in every other element of $\mcssk^\Delta$. Moreover, $\ec^\Delta$ is an epistemic recursive causal model (every valuation in $\T^\Delta$ complies with $\F^{\Delta\vert_{\LAcres}}$, which is recursive). Thus, we get the following.

\begin{theorem}[Completeness of \LOkcres for \LAkcres]\label{teo:completeness-lokres}
  Given a signature $\Sig$, the system \LOkcres is strong\-ly complete for \LAkcres w.r.t. epistemic recursive causal models.
\end{theorem}

\noindent Finally, the argument for strong completeness of \LOkc for formulas in \LAkc\label{proof:thm:rcm-LAk} is analogous to that in the proof of Theorem \ref{thm:rcm-LA} (Page \pageref{proof:thm:rcm-LA}).
\end{appendix:short}

\subsection{Proof of Theorem \ref{thm:rcm-LAka}}\label{app:rcm-LAka}

\begin{appendix:full}
\myparagraph{Soundness.} The soundness of axioms and rules in Tables \ref{tbl:rcm-LA} and \ref{tbl:rcm-LAk} has been already argued for (see the \hyperref[app:rcm-LAk]{proof} of Theorem \ref{thm:rcm-LAk}). The soundness of \ax{N_=} and \ax{N_!} comes from the fact that both an intervention and an announcement on an epistemic (recursive) causal model returns an epistemic (recursive) causal model (Definitions \ref{def:int:epis} and \ref{def:ann:epis}, respectively). For \ax{_!RE} note that, if $\ec$ is an epistemic causal model and $\models \alpha_1 \leftrightarrow \alpha_2$, then $\ec^{\alpha_1}$ and $\ec^{\alpha_2}$ are identical. Then, while \ax{!_{=}}, \ax{!_\land}, \ax{!_\lnot}, \ax{!_K} and \ax{K_!} are standard axioms for a modality $[\alpha!]$ describing the effects of a model operation defined as a partial and deterministic function (see, e.g., \citealp{wang2013axiomatizations}), \ax{!_!} is well-known from public announcement logic \citep[Section 7.4]{vanDitmarschEtAl2007}.

Finally, for the soundness of \ax{=_!}, take any epistemic causal model $\tuple{\F, \T}$ (we omit the signature here, for simplicity) and any $\A \in \T$. By expanding the left-hand side,
\begin{smallctabular}{@{}l@{\;\;}c@{\;\;}l@{}}
  $(\tuple{\F, \T}, \A) \models \int{\opint{\vec{X}}{\vec{x}}}[\alpha!]\chi$
  & iff
  & $(\tuple{ \F_{\opint{\vec{X}}{\vec{x}}}, \newval{\T}{\F}{\opint{\vec{X}}{\vec{x}}} }, \newval{\A}{\F}{\opint{\vec{X}}{\vec{x}}}) \models [\alpha!]\chi$ \\
  & iff
  & $(\tuple{ \F_{\opint{\vec{X}}{\vec{x}}}, \newval{\T}{\F}{\opint{\vec{X}}{\vec{x}}} }, \newval{\A}{\F}{\opint{\vec{X}}{\vec{x}}}) \models \alpha$ \\
  & & implies $(\tuple{ \F_{\opint{\vec{X}}{\vec{x}}}, (\newval{\T}{\F}{\opint{\vec{X}}{\vec{x}}})^{\alpha} }, \newval{\A}{\F}{\opint{\vec{X}}{\vec{x}}}) \models \chi$
\end{smallctabular}
and, by expanding the right-hand side,
\begin{smallctabular}{@{}l@{\;\;}c@{\;\;}l@{}}
  $(\tuple{\F, \T}, \A) \models [\int{\opint{\vec{X}}{\vec{x}}}\alpha!]\int{\opint{\vec{X}}{\vec{x}}}\chi$
  & iff
  & $(\tuple{\F, \T}, \A) \models \int{\opint{\vec{X}}{\vec{x}}}\alpha$ \\
  & & implies $(\tuple{ \F, \T^{\int{\opint{\vec{X}}{\vec{x}}}\alpha} }, \A) \models \int{\opint{\vec{X}}{\vec{x}}}\chi$ \\
  & iff
  & $(\tuple{ \F_{\opint{\vec{X}}{\vec{x}}}, \newval{\T}{\F}{\opint{\vec{X}}{\vec{x}}} }, \newval{\A}{\F}{\opint{\vec{X}}{\vec{x}}}) \models \alpha$ \\
  & & implies $(\tuple{ \F_{\opint{\vec{X}}{\vec{x}}}, \newval{(\T^{\int{\opint{\vec{X}}{\vec{x}}}\alpha})}{\F}{\opint{\vec{X}}{\vec{x}}} }, \newval{\A}{\F}{\opint{\vec{X}}{\vec{x}}}) \models \chi$ \\
\end{smallctabular}
Note that the resulting statements coincide in their antecedents. Moreover, their consequents differ only on the set of valuation functions where $\chi$ is evaluated. Thus, it is enough to check that these sets, $(\newval{\T}{\F}{\opint{\vec{X}}{\vec{x}}})^{\alpha}$ and $\newval{(\T^{\int{\opint{\vec{X}}{\vec{x}}}\alpha})}{\F}{\opint{\vec{X}}{\vec{x}}}$, are identical. For this note how, on the one hand,
\begin{smallctabular}{@{}lcl@{}}
  $\newval{\A}{\F}{\opint{\vec{X}}{\vec{x}}} \in (\newval{\T}{\F}{\opint{\vec{X}}{\vec{x}}})^{\alpha}$
  & iff
  & $\newval{\A}{\F}{\opint{\vec{X}}{\vec{x}}} \in \newval{\T}{\F}{\opint{\vec{X}}{\vec{x}}}$ and $(\tuple{ \F_{\opint{\vec{X}}{\vec{x}}}, \newval{\T}{\F}{\opint{\vec{X}}{\vec{x}}} }, \newval{\A}{\F}{\opint{\vec{X}}{\vec{x}}}) \models \alpha$ \\
  & iff
  & $\A \in \T$ and $(\tuple{ \F_{\opint{\vec{X}}{\vec{x}}}, \newval{\T}{\F}{\opint{\vec{X}}{\vec{x}}} }, \newval{\A}{\F}{\opint{\vec{X}}{\vec{x}}}) \models \alpha$ \\
  & iff
  & $\A \in \T$ and $(\tuple{ \F, \T }, \A) \models \int{\opint{\vec{X}}{\vec{x}}}\alpha$ \\
\end{smallctabular}
and, on the other hand,
\begin{smallctabular}{@{}lcl@{}}
  $\newval{\A}{\F}{\opint{\vec{X}}{\vec{x}}} \in \newval{(\T^{\int{\opint{\vec{X}}{\vec{x}}}\alpha})}{\F}{\opint{\vec{X}}{\vec{x}}}$
  & iff
  & $\A \in \T^{\int{\opint{\vec{X}}{\vec{x}}}\alpha}$ \\
  & iff
  & $\A \in \T$ and $\tuple{ \F, \T }, \A \models \int{\opint{\vec{X}}{\vec{x}}}\alpha$ \\
\end{smallctabular}
This completes this part of the proof.

\myparagraph{Completeness.} For completeness, it will be shown that the axioms on Table \ref{tbl:rcm-LAka}, together with \ax{A_{[]}}, \ax{A_\lnot}, \ax{A_\land} and \ax{CM}, define a translation from \LAfull to \LAkcres, for which the axioms and rules in Tables \ref{tbl:rcm-LA} and \ref{tbl:rcm-LAk} are complete. The following rules will be useful.

\begin{lemma}\label{lem:reka}
  Let $\opint{\vec{X}}{\vec{x}}$ be an assignment on $\Sig$; let $\chi$, $\chi_1$, $\chi_2$, $\alpha$, $\alpha_1$, $\alpha_2$, $[\alpha!]\chi_1$, $[\alpha!]\chi_2$, $[\alpha_1!]\chi$, $[\alpha_2!]\chi$, $\int{\opint{\vec{X}}{\vec{x}}}\chi_1$, $\int{\opint{\vec{X}}{\vec{x}}}\chi_2$ be \LAfull formulas.
  \begin{enumerate}
    \item \ax{RE_!}: if $\vdashlofull \chi_1 \leftrightarrow \chi_2$ then $\vdashlofull [\alpha!]\chi_1 \leftrightarrow [\alpha!]\chi_2$,

    \item \ax{RE_{=}}: If $\vdashlofull \chi_1 \leftrightarrow \chi_2$ then $\vdashlofull \int{\opint{\vec{X}}{\vec{x}}}\chi_1 \leftrightarrow \int{\opint{\vec{X}}{\vec{x}}}\chi_2$.

    \item \ax{{=}!{=}}: $\vdashlofull \int{\opint{\vec{X}}{\vec{x}}}[\alpha!]\int{\opint{\vec{Y}}{\vec{y}}}\chi \,\leftrightarrow\, [\int{\opint{\vec{X}}{\vec{x}}}\alpha!]\int{\opint{\vec{X'}}{\vec{x'}}, \opint{\vec{Y}}{\vec{y}}}{\chi}$,\; with the expression $\opint{\vec{X'}}{\vec{x'}}$ the subassignment of $\opint{\vec{X}}{\vec{x}}$ for $\vec{X'} := \vec{X} \setminus \vec{Y}$.
  \end{enumerate}
  \begin{proof}
    ~
    \begin{enumerate}
      \item As in the proof of \ax{RE_K} on Lemma \ref{lem:rek}, using rule \ax{N_!} and axiom \ax{K_!} instead of rule \ax{N_K} and axiom \ax{K}.

      \item Observe that axioms \ax{A_\lnot} and \ax{A_\land} (together with \ax{P} and \ax{MP}) prove the formula scheme \ax{K_{=}}: $\vdashlofull \int{\opint{\vec{X}}{\vec{x}}}(\chi_1 \rightarrow \chi_2) \rightarrow (\int{\opint{\vec{X}}{\vec{x}}}\chi_1 \rightarrow \int{\opint{\vec{X}}{\vec{x}}}\chi_2)$. Then we proceed as in the case for \ax{RE_K} (Lemma \ref{lem:rek}), using rule \ax{N_{=}} and the scheme \ax{K_{=}} instead of rule \ax{N_K} and axiom \ax{K}.

            \item On the one hand,
      \[
        \vdashlofull
          \int{\opint{\vec{X}}{\vec{x}}}[\alpha!]\int{\opint{\vec{Y}}{\vec{y}}}\chi
          \,\leftrightarrow\,
          [\int{\opint{\vec{X}}{\vec{x}}}\alpha!]\int{\opint{\vec{X}}{\vec{x}}}\int{\opint{\vec{Y}}{\vec{y}}}\chi
      \]
      is an instance of \ax{=_!}. On the other hand, from \ax{A_{[][]}} we get
      \[
        \vdashlofull
          \int{\opint{\vec{X}}{\vec{x}}}{\int{\opint{\vec{Y}}{\vec{y}}}{\chi}}
          \,\leftrightarrow\,
          \int{\opint{\vec{X'}}{\vec{x'}}, \opint{\vec{Y}}{\vec{y}}}{\chi}
      \]
      with $\opint{\vec{X'}}{\vec{x'}}$ the subassignment of $\opint{\vec{X}}{\vec{x}}$ for $\vec{X'} := \vec{X} \setminus \vec{Y}$; hence, from \ax{RE_!} it follows that
      \[
        \vdashlofull
          [\int{\opint{\vec{X}}{\vec{x}}}\alpha!]\int{\opint{\vec{X}}{\vec{x}}}{\int{\opint{\vec{Y}}{\vec{y}}}{\chi}}
          \,\leftrightarrow\,
          [\int{\opint{\vec{X}}{\vec{x}}}\alpha!]\int{\opint{\vec{X'}}{\vec{x'}}, \opint{\vec{Y}}{\vec{y}}}{\chi}
      \]
      Thus, by propositional reasoning from the two pieces,
      \[
        \vdashlofull
          \int{\opint{\vec{X}}{\vec{x}}}[\alpha!]\int{\opint{\vec{Y}}{\vec{y}}}\chi
          \,\leftrightarrow\,
          [\int{\opint{\vec{X}}{\vec{x}}}\alpha!]\int{\opint{\vec{X'}}{\vec{x'}}, \opint{\vec{Y}}{\vec{y}}}{\chi}.
      \]
    \end{enumerate}
  \end{proof}
\end{lemma}

This time, the translation requires two steps: from \LAfull to an intermediate language \LAfullres, and from \LAfullres to the final target \LAkcres.

\begin{definition}[Language \LAfullres]\label{def:LAfullres}
  Given the signature $\Sig$, formulas $\theta$ of the language \LAfullres are given by
  \[ \theta ::= \int{\opint{\vec{X}}{\vec{x}}}{Y{=}y} \mid \lnot \theta \mid \theta \land \theta \mid K \theta \mid [\theta!]\theta \]
  for $Y \in \AV$, $y \in \Ran(Y)$ and $\opint{\vec{X}}{\vec{x}}$ an assignment on $\Sig$. Given a pair $(\ec, \A)$ with $\ec$ an epistemic causal model $\ec = \tuple{\Sig, \F, \T}$ and $\A$ a valuation in $\T$, formulas in \LAfullres are semantically interpreted in the natural way.
\end{definition}

Thus, formulas in $\LAfullres$ (a fragment of \LAfull) are built by the free use of Boolean operators, $K$ and $[\alpha!]$ over `atoms' of the form $\int{\opint{\vec{X}}{\vec{x}}}Y{=}y$.

\myparagraphit{Completeness, step 1: from \LAfull to \LAfullres.} The translation from \LAfull to \LAfullres involves the use of axioms \ax{A_{[]}}, \ax{A_\lnot}, \ax{A_\land} and \ax{CM} for pushing intervention operators through $\lnot$, $\land$ and $K$, and the use of \ax{!_{=}}, \ax{!_\lnot}, \ax{!_\land}, \ax{!_K}, \ax{!_!} and \ax{{=}!{=}} for eliminating public announcement operators \emph{inside the scope of interventions}. Rules \ax{_!RE}, \ax{RE_!} and \ax{RE_{=}} help with the work.

\begin{definition}[Translation $\trr$]\label{def:trr}
  Define $\trr:\LAfull \to \LAfullres$ as
  \begingroup
    \footnotesize
    \[
      \renewcommand{\arraystretch}{1.4}
      \begin{array}{@{}c@{}c@{}}
        \begin{array}{@{}r@{\,:=\,}l@{}}
          \trr(Y{=}y)                                                   & \int{\;}Y{=}y \\
          \trr(\lnot\chi)                                               & \lnot \trr(\chi) \\
          \trr(\chi_1 \land \chi_2)                                     & \trr(\chi_1) \land \trr(\chi_2) \\
          \trr(K\chi)                                                   & K \trr(\chi) \\
          \trr([\alpha!]\chi)                                           & [\trr(\alpha)!]\trr(\chi) \\
        \end{array}
        &
        \begin{array}{@{}r@{\,:=\,}l@{}}
          \trr(\int{\opint{\vec{X}}{\vec{x}}}Y{=}y)                              & \int{\opint{\vec{X}}{\vec{x}}}Y{=}y \\
          \trr(\int{\opint{\vec{X}}{\vec{x}}}\lnot \chi)                         & \trr(\lnot \int{\opint{\vec{X}}{\vec{x}}}\chi) \\
          \trr(\int{\opint{\vec{X}}{\vec{x}}}(\chi_1 \land \chi_2))              & \trr(\int{\opint{\vec{X}}{\vec{x}}}\chi_1 \land \int{\opint{\vec{X}}{\vec{x}}}\chi_2) \\
          \trr(\int{\opint{\vec{X}}{\vec{x}}}K\chi)                              & \trr(K\int{\opint{\vec{X}}{\vec{x}}}\chi) \\
          \multicolumn{2}{l}{} \\
          \trr(\int{\opint{\vec{X}}{\vec{x}}}[\alpha!]Y{=}y)                     & \trr(\int{\opint{\vec{X}}{\vec{x}}}(\alpha \rightarrow Y{=}y)) \\
          \trr(\int{\opint{\vec{X}}{\vec{x}}}[\alpha!]\lnot \chi)                & \trr(\int{\opint{\vec{X}}{\vec{x}}}(\alpha \rightarrow \lnot [\alpha!]\chi)) \\
          \trr(\int{\opint{\vec{X}}{\vec{x}}}[\alpha!](\chi_1 \land \chi_2))     & \trr(\int{\opint{\vec{X}}{\vec{x}}}([\alpha!]\chi_1 \land [\alpha!]\chi_2)) \\
          \trr(\int{\opint{\vec{X}}{\vec{x}}}[\alpha!]K\chi)                     & \trr(\int{\opint{\vec{X}}{\vec{x}}}(\alpha \rightarrow K(\alpha \rightarrow [\alpha!]\chi))) \\
          \trr(\int{\opint{\vec{X}}{\vec{x}}}[\alpha_1!][\alpha_2!]\chi) & \trr(\int{\opint{\vec{X}}{\vec{x}}}[\alpha_1 \land [\alpha_1!]\alpha_2!]\chi) \\
          \trr(\int{\opint{\vec{X}}{\vec{x}}}[\alpha!]\int{\opint{\vec{Y}}{\vec{y}}}\chi) & \trr([\int{\opint{\vec{X}}{\vec{x}}}\alpha!]\int{\opint{\vec{X'}}{\vec{x'}}, \opint{\vec{Y}}{\vec{y}}}{\chi}) \\
          \multicolumn{2}{l}{} \\
          \trr(\int{\opint{\vec{X}}{\vec{x}}}\int{\opint{\vec{Y}}{\vec{y}}}\chi) & \trr(\int{\opint{\vec{X'}}{\vec{x'}}, \opint{\vec{Y}}{\vec{y}}}\chi) \\
        \end{array}
      \end{array}
    \]
  \endgroup
  where, in the two last clauses on the right, $\opint{\vec{X'}}{\vec{x'}}$ is the subassignment of $\opint{\vec{X}}{\vec{x}}$ for $\vec{X'} := \vec{X} \setminus \vec{Y}$.
\end{definition}

\begin{proposition}\label{pro:LAkatoLAkares}
  For every $\chi \in \LAfull$,
  \begin{multicols}{3}
    \begin{enumerate}
      \item\label{pro:LAkatoLAkares:trans} $\trr(\chi) \in \LAfullres$,
      \item\label{pro:LAkatoLAkares:syn} $\vdashlofull \chi \leftrightarrow \trr(\chi)$,
      \item\label{pro:LAkatoLAkares:sem} $\models \chi \leftrightarrow \trr(\chi)$.
    \end{enumerate}
  \end{multicols}
  \begin{proof}
    The proofs of the items in this proposition will be again by induction on the formulas' complexity $\com:\LAfull \to \Nat\setminus\set{0}$, which is this time defined as
    \begin{ctabular}{c@{\qquad}c}
      \begin{tabular}[t]{r@{\;:=\;}l}
        $\com(Y{=}y)$             & $1$ \\
        $\com(\lnot \chi)$         & $1 + \com(\chi)$ \\
        $\com(\chi_1 \land \chi_2)$ & $1 + \max\set{\com(\chi_1), \com(\chi_2)}$ \\
      \end{tabular}
      &
      \begin{tabular}[t]{r@{\;:=\;}l}
        $\com(K\chi)$                                & $1 + \com(\chi)$ \\
        $\com([\alpha!]\chi)$                        & $(7+\com(\alpha))\com(\chi)$ \\
        $\com(\int{\opint{\vec{X}}{\vec{x}}}{\chi})$ & $2\com(\chi)$
      \end{tabular}
    \end{ctabular}
    The case for $[\alpha!]\chi$ seems arbitrary, but it guarantees the following properties:\footnote{\citet[Section 7.4]{vanDitmarschEtAl2007} use, in their Definition 7.21, rather $\com([\alpha!]\chi):=(4+\com(\alpha))\com(\chi)$. Here a modification is needed, as 7 is the least natural number guaranteeing that the inequality $\com(\int{\opint{\vec{X}}{\vec{x}}}[\alpha!]K\chi) > \com(\int{\opint{\vec{X}}{\vec{x}}}(\alpha \rightarrow K(\alpha \rightarrow [\alpha!]\chi)))$ holds.} for all assignments $\opint{\vec{X}}{\vec{x}}$, $\opint{\vec{Y}}{\vec{y}}$ and all formulas $\chi, \chi_1, \chi_2 \in \LAfull$,
    \begingroup
      \small
      \[
        \begin{array}{r@{\;}c@{\;}l}
          \toprule
          \com(\chi) & \geqslant & \com(\psi) \;\text{for every}\; \psi \in \sub(\chi), \\
          \midrule
          \com(\int{\opint{\vec{X}}{\vec{x}}}{\lnot \chi}) & > &\com(\lnot \int{\opint{\vec{X}}{\vec{x}}}{\chi}), \\
          \com(\int{\opint{\vec{X}}{\vec{x}}}{(\chi_1 \land \chi_2)}) & > & \com(\int{\opint{\vec{X}}{\vec{x}}}{\chi_1} \land \int{\opint{\vec{X}}{\vec{x}}}{\chi_2}) \\
          \com(\int{\opint{\vec{X}}{\vec{x}}}{K \chi}) & > & \com(K \int{\opint{\vec{X}}{\vec{x}}}{\chi}) \\
          \midrule
          \com(\int{\opint{\vec{X}}{\vec{x}}}[\alpha!]Y{=}y) & > & \com(\int{\opint{\vec{X}}{\vec{x}}}(\alpha \rightarrow Y{=}y)) \\
          \com(\int{\opint{\vec{X}}{\vec{x}}}[\alpha!]\lnot \chi) & > & \com(\int{\opint{\vec{X}}{\vec{x}}}(\alpha \rightarrow \lnot [\alpha!]\chi)) \\
          \com(\int{\opint{\vec{X}}{\vec{x}}}[\alpha!](\chi_1 \land \chi_2)) & > & \trr(\int{\opint{\vec{X}}{\vec{x}}}([\alpha!]\chi_1 \land [\alpha!]\chi_2)) \\
          \com(\int{\opint{\vec{X}}{\vec{x}}}[\alpha!]K\chi) & > & \com(\int{\opint{\vec{X}}{\vec{x}}}(\alpha \rightarrow K(\alpha \rightarrow [\alpha!]\chi))) \\
          \com(\int{\opint{\vec{X}}{\vec{x}}}[\alpha_1!][\alpha_2!]\chi) & > & \com(\int{\opint{\vec{X}}{\vec{x}}}[\alpha_1 \land [\alpha_1!]\alpha_2!]\chi) \\
          \com(\int{\opint{\vec{X}}{\vec{x}}}[\alpha!]\int{\opint{\vec{Y}}{\vec{y}}}\chi) & > & \com([\int{\opint{\vec{X}}{\vec{x}}}\alpha!]\int{\opint{\vec{X'}}{\vec{x'}}, \opint{\vec{Y}}{\vec{y}}}{\chi}) \; \begin{minipage}[t]{0.3\textwidth}\begin{flushright} with $\opint{\vec{X'}}{\vec{x'}}$ the subassignment of $\opint{\vec{X}}{\vec{x}}$ for $\vec{X'} := \vec{X} \setminus \vec{Y}$ \smallskip \end{flushright}\end{minipage} \\
          \com(\int{\opint{\vec{X}}{\vec{x}}}\int{\opint{\vec{Y}}{\vec{y}}}\chi) & > & \com(\int{\opint{\vec{X'}}{\vec{x'}}, \opint{\vec{Y}}{\vec{y}}}\chi) \\
          \bottomrule
        \end{array}
      \]
    \endgroup
    The first block can be proved by structural induction (with $\sub([\alpha!]\chi) := \set{[\alpha!]\chi} \cup \sub(\alpha) \cup \sub(\chi)$). Items in the second block can be proved as their respective counterparts in Proposition \ref{pro:LAktoLAkres}. For items in the third block, all but the next to last and last are consequences of their `interventionless' counterparts, which are analogous to the cases in public announcement logic \citep[Lemma 7.22]{vanDitmarschEtAl2007}. For the next to last, note that $\com(\int{\opint{\vec{X}}{\vec{x}}}[\alpha!]\int{\opint{\vec{Y}}{\vec{y}}}\chi) = (28 + 4\com(\alpha))\com(\chi)$, yet $\com([\int{\opint{\vec{X}}{\vec{x}}}\alpha!]\int{\opint{\vec{X'}}{\vec{x'}}, \opint{\vec{Y}}{\vec{y}}}{\chi}) = (14+4\com(\alpha)) \com(\chi)$. The last is straightforward.
    {\medskip}

    Then, the properties.
    \begin{compactenumerate}
      \item The proof, by induction on $\com(\chi)$, is analogous to that in Proposition \ref{pro:LAktoLAkres}. The base case (formulas $\chi$ with $\bs{\com(\chi)=1}$, i.e., $Y{=}y$) is straightforward, as $\trr(Y{=}y) = \int{\;}{Y{=}y}$ is an \LAfullres-atom. For the inductive part (formulas $\chi$ with $\bs{\com(\chi) > 1}$, i.e., formulas $\lnot \chi$,\; $\chi_1 \land \chi_2$,\; $K \chi$,\; $[\alpha!]\chi$,\; $\int{\opint{\vec{X}}{\vec{x}}}{Y{=}y}$,\; $\int{\opint{\vec{X}}{\vec{x}}}{\lnot \chi}$,\; $\int{\opint{\vec{X}}{\vec{x}}}{(\chi_1 \land \chi_2)}$,\; $\int{\opint{\vec{X}}{\vec{x}}}{K \chi}$,\; $\int{\opint{\vec{X}}{\vec{x}}}[\alpha!]Y{=}y$,\; $\int{\opint{\vec{X}}{\vec{x}}}[\alpha!]\lnot \chi$,\; $\int{\opint{\vec{X}}{\vec{x}}}[\alpha!](\chi_1 \land \chi_2)$,\; $\int{\opint{\vec{X}}{\vec{x}}}[\alpha!]K\chi$, $\int{\opint{\vec{X}}{\vec{x}}}[\alpha_1!][\alpha_2!]\chi$, $\int{\opint{\vec{X}}{\vec{x}}}[\alpha!]\int{\opint{\vec{Y}}{\vec{y}}}\chi$ and $\int{\opint{\vec{X}}{\vec{x}}}\int{\opint{\vec{Y}}{\vec{y}}}\chi$) use the properties of $\com$ to generate appropriate IHs, and then use the definition of $\trr$ (note: the case for $\int{\opint{\vec{X}}{\vec{x}}}{Y{=}y}$ is straightforward).

      \item The proof, by induction on $\com(\chi)$, is analogous to that in Proposition \ref{pro:LAktoLAkres}. The base case (formulas $\chi$ with $\bs{\com(\chi)=1}$, i.e., $Y{=}y$) follows from axiom \ax{A_{[]}}. The inductive cases rely on the properties of $\com$ to generate appropriate IHs, and then use the definition of $\trr$ with propositional reasoning (cases $\lnot \chi$, $\chi_1 \land \chi_2$ and $\int{\opint{\vec{X}}{\vec{x}}}{Y{=}y}$) plus \ax{RE_K} (case $K \chi$), \ax{_!RE} and \ax{RE_!} from Lemma \ref{lem:reka} (case $[\alpha!]\chi$),\footnote{Indeed, since $\alpha, \chi \in \sub([\alpha!]\chi)$, from IH it follows that $\vdashlofull \alpha \leftrightarrow \trr(\alpha)$ and $\vdashlofull \chi \leftrightarrow \trr(\chi)$. From the first and \ax{_!RE} it follows that $\vdashlofull [\alpha!]\chi \leftrightarrow [\trr(\alpha)!]\chi$; from the second and \ax{RE_!} it follows that $\vdashlofull [\trr(\alpha)!]\chi \leftrightarrow [\trr(\alpha)!]\trr(\chi)$. Then, by propositional reasoning (\ax{P} and \ax{MP}), $\vdashlofull [\alpha!]\chi \leftrightarrow [\trr(\alpha)!]\trr(\chi)$, which by definition of $\trr$ is the required $\vdashlofull [\alpha!]\chi \leftrightarrow \trr([\alpha!]\chi)$.} axiom \ax{A_\lnot} (case $\int{\opint{\vec{X}}{\vec{x}}}{\lnot \chi}$), axiom \ax{A_\land} (case $\int{\opint{\vec{X}}{\vec{x}}}{(\chi_1 \land \chi_2)}$ and axiom \ax{CM} (case $\int{\opint{\vec{X}}{\vec{x}}}{K \chi}$). Cases $\int{\opint{\vec{X}}{\vec{x}}}[\alpha!]Y{=}y$,\; $\int{\opint{\vec{X}}{\vec{x}}}[\alpha!]\lnot \chi$,\; $\int{\opint{\vec{X}}{\vec{x}}}[\alpha!](\chi_1 \land \chi_2)$,\; $\int{\opint{\vec{X}}{\vec{x}}}[\alpha!]K\chi$ \;and\; $\int{\opint{\vec{X}}{\vec{x}}}[\alpha_1!][\alpha_2!]\chi$ follow the same pattern: use the properties of $\com$ to obtain an appropriate IH, then apply \ax{RE_{=}} (Lemma \ref{lem:reka}) over the corresponding axiom, and finally use the definition of $\trr$.\footnote{As an example, consider $\int{\opint{\vec{X}}{\vec{x}}}[\alpha!]Y{=}y$. Since $\com(\int{\opint{\vec{X}}{\vec{x}}}[\alpha!]Y{=}y) > \com(\int{\opint{\vec{X}}{\vec{x}}}(\alpha \rightarrow Y{=}y))$, from IH it follows that $\vdashlofull \int{\opint{\vec{X}}{\vec{x}}}(\alpha \rightarrow Y{=}y) \leftrightarrow \trr(\int{\opint{\vec{X}}{\vec{x}}}(\alpha \rightarrow Y{=}y))$. But applying \ax{RE_{=}} on \ax{!_=} yields $\vdashlofull \int{\opint{\vec{X}}{\vec{x}}}[\alpha!]Y{=}y \leftrightarrow \int{\opint{\vec{X}}{\vec{x}}}(\alpha \rightarrow Y{=}y)$, so $\vdashlofull \int{\opint{\vec{X}}{\vec{x}}}[\alpha!]Y{=}y \leftrightarrow \trr(\int{\opint{\vec{X}}{\vec{x}}}(\alpha \rightarrow Y{=}y))$. Then, from the definition of $\trr$, it follows that $\vdashlofull \int{\opint{\vec{X}}{\vec{x}}}[\alpha!]Y{=}y \leftrightarrow \trr(\int{\opint{\vec{X}}{\vec{x}}}[\alpha!]Y{=}y)$. As another example consider $\int{\opint{\vec{X}}{\vec{x}}}[\alpha_1!][\alpha_2!]\chi$. Since $\com(\int{\opint{\vec{X}}{\vec{x}}}[\alpha_1!][\alpha_2!]\chi) > \com(\int{\opint{\vec{X}}{\vec{x}}}[\alpha_1 \land [\alpha_1!]\alpha_2!]\chi)$, from IH it follows that $\vdashlofull \int{\opint{\vec{X}}{\vec{x}}}[\alpha_1 \land [\alpha_1!]\alpha_2!]\chi \leftrightarrow \trr(\int{\opint{\vec{X}}{\vec{x}}}[\alpha_1 \land [\alpha_1!]\alpha_2!]\chi)$. But applying \ax{RE_{=}} on \ax{!_!} yields $\vdashlofull \int{\opint{\vec{X}}{\vec{x}}}[\alpha_1!][\alpha_2!]\chi \leftrightarrow \int{\opint{\vec{X}}{\vec{x}}}[\alpha_1 \land [\alpha_1!]\alpha_2!]\chi$, so $\vdashlofull \int{\opint{\vec{X}}{\vec{x}}}[\alpha_1!][\alpha_2!]\chi \leftrightarrow \trr(\int{\opint{\vec{X}}{\vec{x}}}[\alpha_1 \land [\alpha_1!]\alpha_2!]\chi)$. Then, from the definition of $\trr$, it follows that $\vdashlofull \int{\opint{\vec{X}}{\vec{x}}}[\alpha_1!][\alpha_2!]\chi \leftrightarrow \trr(\int{\opint{\vec{X}}{\vec{x}}}[\alpha_1 \land [\alpha_1!]\alpha_2!]\chi)$.} The case $\int{\opint{\vec{X}}{\vec{x}}}\int{\opint{\vec{Y}}{\vec{y}}}\chi$ relies on axiom \ax{A_{[][]}}, just as its counterpart in the proof of Proposition \ref{pro:LAtoLAres}. There is just one case left; here are the details.
      \begin{compactitemize}
        \item Case $\int{\opint{\vec{X}}{\vec{x}}}[\alpha!]\int{\opint{\vec{Y}}{\vec{y}}}\chi$. Use again IH, now on $\com(\int{\opint{\vec{X}}{\vec{x}}}[\alpha!]\int{\opint{\vec{Y}}{\vec{y}}}\chi) > \com([\int{\opint{\vec{X}}{\vec{x}}}\alpha!]\int{\opint{\vec{X'}}{\vec{x'}}, \opint{\vec{Y}}{\vec{y}}}{\chi})$ (with $\opint{\vec{X'}}{\vec{x'}}$ as indicated above) to obtain $\vdashlofull [\int{\opint{\vec{X}}{\vec{x}}}\alpha!]\int{\opint{\vec{X'}}{\vec{x'}}, \opint{\vec{Y}}{\vec{y}}}{\chi} \leftrightarrow \trr([\int{\opint{\vec{X}}{\vec{x}}}\alpha!]\int{\opint{\vec{X'}}{\vec{x'}}, \opint{\vec{Y}}{\vec{y}}}{\chi})$. But $\vdashlofull \int{\opint{\vec{X}}{\vec{x}}}[\alpha!]\int{\opint{\vec{Y}}{\vec{y}}}\chi \,\leftrightarrow\, [\int{\opint{\vec{X}}{\vec{x}}}\alpha!]\int{\opint{\vec{X'}}{\vec{x'}}, \opint{\vec{Y}}{\vec{y}}}{\chi}$ (axiom \ax{{=}!{=}}; see Lemma \ref{lem:reka}). Then, by propositional reasoning, it follows that $\vdashlofull \int{\opint{\vec{X}}{\vec{x}}}[\alpha!]\int{\opint{\vec{Y}}{\vec{y}}}\chi \,\leftrightarrow\, \trr([\int{\opint{\vec{X}}{\vec{x}}}\alpha!]\int{\opint{\vec{X'}}{\vec{x'}}, \opint{\vec{Y}}{\vec{y}}}{\chi})$. This is enough, as the right-hand side of the latter is, by definition, $\trr(\int{\opint{\vec{X}}{\vec{x}}}[\alpha!]\int{\opint{\vec{Y}}{\vec{y}}}\chi)$.
      \end{compactitemize}
      \item By the previous item, $\vdashlofull \chi \leftrightarrow \trr(\chi)$. But \LOfull is sound for recursive causal models; therefore, $\models \chi \leftrightarrow \trr(\chi)$.
    \end{compactenumerate}
  \end{proof}
\end{proposition}

\myparagraphit{Completeness, step 2: from \LAfullres to \LAkcres.} The translation from \LAfullres to \LAkcres involves the use of \ax{!_{=}}, \ax{!_\lnot}, \ax{!_\land} and \ax{!_K} for eliminating public announcement operators.

\begin{definition}[Translation $\trf$]\label{def:trf}
  Define $\trf:\LAfullres \to \LAkcres$ as
  \begingroup
    \footnotesize
    \[
      \renewcommand{\arraystretch}{1.4}
      \begin{array}{@{}c@{}c@{}}
        \begin{array}{@{}r@{\,:=\,}l@{}}
          \trf(\int{\opint{\vec{X}}{\vec{x}}}Y{=}y)                     & \int{\opint{\vec{X}}{\vec{x}}}Y{=}y \\
          \trf(\lnot\theta)                                             & \lnot \trf(\theta) \\
          \trf(\theta_1 \land \theta_2)                                 & \trf(\theta_1) \land \trf(\theta_2) \\
          \trf(K\theta)                                                 & K \trf(\theta) \\
        \end{array}
        &
        \begin{array}{@{}r@{\,:=\,}l@{}}
          \trf([\alpha!]\int{\opint{\vec{X}}{\vec{x}}}Y{=}y)                        & \trf(\alpha \rightarrow \int{\opint{\vec{X}}{\vec{x}}}Y{=}y) \\
          \trf([\alpha!]\lnot \theta)                & \trf(\alpha \rightarrow \lnot [\alpha!]\theta) \\
          \trf([\alpha!](\theta_1 \land \theta_2))   & \trf([\alpha!]\theta_1 \land [\alpha!]\theta_2) \\
          \trf([\alpha!]K\theta)                     & \trf(\alpha \rightarrow K(\alpha \rightarrow [\alpha!]\theta)) \\
          \trf([\alpha_1!][\alpha_2!]\theta) & \trf([\alpha_1 \land [\alpha_1!]\alpha_2!]\theta) \\
        \end{array}
      \end{array}
    \]
  \endgroup
\end{definition}

\begin{proposition}\label{pro:LAkarestoLAkres}
  For every $\theta \in \LAfullres$,
  \begin{multicols}{3}
    \begin{enumerate}
      \item\label{pro:LAkarestoLAkres:trans} $\trf(\theta) \in \LAkcres$,
      \item\label{pro:LAkarestoLAkres:syn} $\vdashlofull \theta \leftrightarrow \trf(\theta)$,
      \item\label{pro:LAkarestoLAkres:sem} $\models \theta \leftrightarrow \trf(\theta)$.
    \end{enumerate}
  \end{multicols}
  \begin{proof}
    Again, by induction on the formulas' complexity $\com:\LAfullres \to \Nat\setminus\set{0}$, which is this time defined as
    \begin{ctabular}{c@{\qquad}c}
      \begin{tabular}[t]{r@{\;:=\;}l}
        $\com(\int{\opint{\vec{X}}{\vec{x}}}Y{=}y)$    & $1$ \\
        $\com(\lnot \theta)$                           & $1 + \com(\theta)$ \\
        $\com(\theta_1 \land \theta_2)$                & $1 + \max\set{\com(\theta_1), \com(\theta_2)}$ \\
      \end{tabular}
      &
      \begin{tabular}[t]{r@{\;:=\;}l}
        $\com(K\theta)$                                & $1 + \com(\theta)$ \\
        $\com([\alpha!]\theta)$                        & $(7+\com(\alpha))\com(\theta)$ \\
      \end{tabular}
    \end{ctabular}
    This definition guarantees that, for every assignment $\opint{\vec{X}}{\vec{x}}$ and all formulas $\theta, \theta_1, \theta_2 \in \LAfullres$,
    \begingroup
      \small
      \[
        \begin{array}{r@{\;}c@{\;}l}
          \toprule
          \com(\theta) & \geqslant & \com(\psi) \;\text{for every}\; \psi \in \sub(\theta), \\
          \midrule
          \com([\alpha!]\int{\opint{\vec{X}}{\vec{x}}}Y{=}y) & > & \com(\alpha \rightarrow \int{\opint{\vec{X}}{\vec{x}}}Y{=}y) \\
          \com([\alpha!]\lnot \theta) & > & \com(\alpha \rightarrow \lnot [\alpha!]\theta) \\
          \com([\alpha!](\theta_1 \land \theta_2)) & > & \com([\alpha!]\theta_1 \land [\alpha!]\theta_2) \\
          \com([\alpha!]K\theta) & > & \com(\alpha \rightarrow K(\alpha \rightarrow [\alpha!]\theta)) \\
          \com([\alpha_1!][\alpha_2!]\theta) & > & \com([\alpha_1 \land [\alpha_1!]\alpha_2!]\theta) \\
          \bottomrule
        \end{array}
      \]
    \endgroup
    The first block is proved by structural induction (with $\sub([\alpha!]\theta) := \set{[\alpha!]\theta} \cup \sub(\alpha) \cup \sub(\theta)$). Items in the second are just small variations of the well-known cases from public announcement logic \citep[Section 7.4]{vanDitmarschEtAl2007}.

    Then, the properties.
    \begin{compactenumerate}
      \item The proof, by induction on $\com(\theta)$, is analogous to that in Proposition \ref{pro:LAktoLAkres}. The base case (formulas $\theta$ with $\bs{\com(\theta)=1}$, i.e., $\int{\opint{\vec{X}}{\vec{x}}}Y{=}y$) is straightforward, as $\trf(\int{\opint{\vec{X}}{\vec{x}}}Y{=}y) = \int{\opint{\vec{X}}{\vec{x}}}{Y{=}y}$ is an \LAkcres-atom. For the inductive part (formulas $\theta$ with $\bs{\com(\theta) > 1}$, i.e., formulas $\lnot \theta$,\; $\theta_1 \land \theta_2$,\; $K \theta$,\; $[\alpha!]\int{\opint{\vec{X}}{\vec{x}}}Y{=}y$,\; $[\alpha!]\lnot \theta$,\; $[\alpha!](\theta_1 \land \theta_2)$,\; $[\alpha!]K\theta$ and $[\alpha_1!][\alpha_2!]\theta$) use the properties of $\com$ to generate appropriate IHs, and then use the definition of $\trf$.

      \item The proof, by induction on $\com(\theta)$, is analogous to that in Proposition \ref{pro:LAktoLAkres}. The base case (formulas $\theta$ with $\bs{\com(\theta)=1}$, i.e., $\int{\opint{\vec{X}}{\vec{x}}}Y{=}y$) is by propositional reasoning. The inductive cases rely on the properties of $\com$ to generate appropriate IHs, and then use the definition of $\trf$ with propositional reasoning (cases $\lnot \theta$, $\theta_1 \land \theta_2$) plus \ax{RE_K} (case $K \theta$), axiom \ax{!_{=}} (case $[\alpha!]\int{\opint{\vec{X}}{\vec{x}}}Y{=}y$), axiom \ax{!_\lnot} (case $[\alpha!]\lnot \theta$), axiom \ax{!_\land} (case $[\alpha!](\theta_1 \land \theta_2)$,\; axiom \ax{!_K} (case $[\alpha!]K\theta$) and axiom \ax{!_!} (case $[\alpha_1!][\alpha_2!]\theta$).\footnote{As an example, consider $[\alpha!]\int{\opint{\vec{X}}{\vec{x}}}Y{=}y$. Since $\com([\alpha!]\int{\opint{\vec{X}}{\vec{x}}}Y{=}y) > \com(\alpha \rightarrow \int{\opint{\vec{X}}{\vec{x}}}Y{=}y)$, from IH it follows that $\vdashlofull (\alpha \rightarrow \int{\opint{\vec{X}}{\vec{x}}}Y{=}y) \leftrightarrow \trf(\alpha \rightarrow \int{\opint{\vec{X}}{\vec{x}}}Y{=}y)$. But \ax{!_=} yields $\vdashlofull [\alpha!]\int{\opint{\vec{X}}{\vec{x}}}Y{=}y \leftrightarrow (\alpha \rightarrow \int{\opint{\vec{X}}{\vec{x}}}Y{=}y)$, so $\vdashlofull [\alpha!]\int{\opint{\vec{X}}{\vec{x}}}Y{=}y \leftrightarrow \trf(\alpha \rightarrow \int{\opint{\vec{X}}{\vec{x}}}Y{=}y)$. Then, from the definition of $\trf$, it follows that $\vdashlofull [\alpha!]\int{\opint{\vec{X}}{\vec{x}}}Y{=}y \leftrightarrow \trf([\alpha!]\int{\opint{\vec{X}}{\vec{x}}}Y{=}y)$. As another example, consider $[\alpha_1!][\alpha_2!]\theta$. Since $\com([\alpha_1!][\alpha_2!]\theta) > \com([\alpha_1 \land [\alpha_1!]\alpha_2!]\theta)$, from IH it follows that $\vdashlofull [\alpha_1 \land [\alpha_1!]\alpha_2!]\theta \leftrightarrow \trf([\alpha_1 \land [\alpha_1!]\alpha_2!]\theta)$. But \ax{!_!} yields $\vdashlofull [\alpha_1!][\alpha_2!]\theta \leftrightarrow [\alpha_1 \land [\alpha_1!]\alpha_2!]\theta$, so $\vdashlofull [\alpha_1!][\alpha_2!]\theta \leftrightarrow \trf([\alpha_1 \land [\alpha_1!]\alpha_2!]\theta)$. Then, from the definition of $\trf$, it follows that $\vdashlofull [\alpha_1!][\alpha_2!]\theta \leftrightarrow \trf([\alpha_1!][\alpha_2!]\theta)$.}

      \item By the previous item, $\vdashlofull \theta \leftrightarrow \trf(\theta)$. But \LOfull is sound within recursive causal models; therefore, $\models \theta \leftrightarrow \trf(\theta)$.
    \end{compactenumerate}
  \end{proof}
\end{proposition}

Finally, for strong completeness of \LOfull for formulas in \LAfull, define $\tr:\LAfull \to \LAkcres$ as the composition of $\trr$ and then $\trf$, that is, as $\tr(\chi') := \trf(\trr(\chi'))$ for every $\chi' \in \LAfull$. Note: $\tr$ is properly defined because $\trr(\chi') \in \LAfullres$ (Proposition \ref{pro:LAkatoLAkares}.\ref{pro:LAkatoLAkares:trans}), and thus $\trf$ can be applied to $\trr(\chi')$. Note also how, for every $\chi' \in \LAfull$, the new translation $\tr$ is such that
\begin{itemize}
  \item $\tr(\chi') \in \LAkcres$ (as $\trf(\trr(\chi')) \in \LAkcres$ by Proposition \ref{pro:LAkarestoLAkres}.\ref{pro:LAkarestoLAkres:trans}).

  \item $\vdashlofull \chi' \leftrightarrow \tr(\chi')$. Indeed, take any $\chi' \in \LAfull$. By Proposition \ref{pro:LAkatoLAkares}.\ref{pro:LAkatoLAkares:syn}, $\vdashlofull \chi' \leftrightarrow \trr(\chi')$. But $\trr(\chi') \in \LAfullres$ (Proposition \ref{pro:LAkatoLAkares}.\ref{pro:LAkatoLAkares:trans}) so, by Proposition \ref{pro:LAkarestoLAkres}.\ref{pro:LAkarestoLAkres:syn}, $\vdashlofull \trr(\chi') \leftrightarrow \trf(\trr(\chi'))$. Hence, $\vdashlofull \chi' \leftrightarrow \trf(\trr(\chi'))$, and thus the required $\vdashlofull \chi' \leftrightarrow \tr(\chi')$ follows.

  \item $\models \chi' \leftrightarrow \tr(\chi')$. Indeed, take any $\chi' \in \LAfull$. By Proposition \ref{pro:LAkatoLAkares}.\ref{pro:LAkatoLAkares:sem}, $\models \chi' \leftrightarrow \trr(\chi')$. But $\trr(\chi') \in \LAfullres$ (Proposition \ref{pro:LAkatoLAkares}.\ref{pro:LAkatoLAkares:trans}) so, by Proposition \ref{pro:LAkarestoLAkres}.\ref{pro:LAkarestoLAkres:sem}, $\models \trr(\chi') \leftrightarrow \trf(\trr(\chi'))$. Hence, $\models \chi' \leftrightarrow \trf(\trr(\chi'))$, and thus the required $\models \chi' \leftrightarrow \tr(\chi')$ follows.
\end{itemize}

Then, the argument for strong completeness of \LOfull for formulas in \LAfull is as that for Theorem \ref{thm:rcm-LA} (Page \pageref{proof:thm:rcm-LA}). First, take $\Psi \cup \set{\chi} \subseteq \LAfull$ and suppose $\Psi \models \chi$. Since $\models \chi' \leftrightarrow \tr(\chi')$ for every $\chi' \in \LAfull$ (see above), by defining $\tr(\Psi) := \set{\tr(\psi) \mid \psi \in \Psi}$ it follows that $\tr(\Psi) \models \tr(\chi)$. Now, $\tr(\chi') \in \LAkcres$ for every $\chi' \in \LAfull$ (see above). Hence, $\tr(\Psi) \cup \set{\tr(\chi)} \subseteq \LAkcres$ and therefore, the just obtained $\tr(\Psi) \models \tr(\chi)$ and Theorem \ref{teo:completeness-lokres} imply $\tr(\Psi) \vdashlokcres \tr(\varphi)$. Since \LOkcres is a subsystem of \LOfull, it follows that $\tr(\Psi) \vdashlofull \tr(\chi)$. But $\vdashlofull \chi' \leftrightarrow \tr(\chi')$ for every $\chi' \in \LAfull$ (see above); hence, $\Psi \vdashlofull \chi$, as required.

\end{appendix:full}

\begin{appendix:short}
\myparagraph{Soundness.} For axioms and rules in Tables \ref{tbl:rcm-LA} and \ref{tbl:rcm-LAk} see Theorem \ref{thm:rcm-LAk}. Soundness of \ax{N_=} and \ax{N_!} comes from the fact that both an intervention and an announcement on an epistemic (recursive) causal model returns an epistemic (recursive) causal model (Definitions \ref{def:int:epis} and \ref{def:ann:epis}, respectively). For \ax{_!RE} note that, if $\ec$ is an epistemic causal model and $\models \alpha_1 \leftrightarrow \alpha_2$, then $\ec^{\alpha_1}$ and $\ec^{\alpha_2}$ are identical. Then, while \ax{!_{=}}, \ax{!_\land}, \ax{!_\lnot}, \ax{!_K} and \ax{K_!} are standard axioms for a modality $[\alpha!]$ describing the effects of a model operation defined as a partial and deterministic function (see, e.g., \citealp{wang2013axiomatizations}), \ax{!_!} is well-known from public announcement logic \citep[Section 7.4]{vanDitmarschEtAl2007}.

Finally, for \ax{=_!}, take any epistemic causal model $\tuple{\F, \T}$ (for simplicity, omit the signature) and any $\A \in \T$. Expanding the axiom's left-hand side yields
\begin{smallctabular}{@{}l@{\;\;}c@{\;\;}l@{}}
  $(\tuple{\F, \T}, \A) \models \int{\opint{\vec{X}}{\vec{x}}}[\alpha!]\chi$
  & iff
  & $(\tuple{ \F_{\opint{\vec{X}}{\vec{x}}}, \newval{\T}{\F}{\opint{\vec{X}}{\vec{x}}} }, \newval{\A}{\F}{\opint{\vec{X}}{\vec{x}}}) \models [\alpha!]\chi$ \\
  & iff
  & $(\tuple{ \F_{\opint{\vec{X}}{\vec{x}}}, \newval{\T}{\F}{\opint{\vec{X}}{\vec{x}}} }, \newval{\A}{\F}{\opint{\vec{X}}{\vec{x}}}) \models \alpha$ \\
  & & implies $(\tuple{ \F_{\opint{\vec{X}}{\vec{x}}}, (\newval{\T}{\F}{\opint{\vec{X}}{\vec{x}}})^{\alpha} }, \newval{\A}{\F}{\opint{\vec{X}}{\vec{x}}}) \models \chi$
\end{smallctabular}
and expanding its right-hand yields
\begin{smallctabular}{@{}l@{\;\;}c@{\;\;}l@{}}
  $(\tuple{\F, \T}, \A) \models [\int{\opint{\vec{X}}{\vec{x}}}\alpha!]\int{\opint{\vec{X}}{\vec{x}}}\chi$
  & iff
  & $(\tuple{\F, \T}, \A) \models \int{\opint{\vec{X}}{\vec{x}}}\alpha$ \\
  & & implies $(\tuple{ \F, \T^{\int{\opint{\vec{X}}{\vec{x}}}\alpha} }, \A) \models \int{\opint{\vec{X}}{\vec{x}}}\chi$ \\
  & iff
  & $(\tuple{ \F_{\opint{\vec{X}}{\vec{x}}}, \newval{\T}{\F}{\opint{\vec{X}}{\vec{x}}} }, \newval{\A}{\F}{\opint{\vec{X}}{\vec{x}}}) \models \alpha$ \\
  & & implies $(\tuple{ \F_{\opint{\vec{X}}{\vec{x}}}, \newval{(\T^{\int{\opint{\vec{X}}{\vec{x}}}\alpha})}{\F}{\opint{\vec{X}}{\vec{x}}} }, \newval{\A}{\F}{\opint{\vec{X}}{\vec{x}}}) \models \chi$ \\
\end{smallctabular}
The resulting implications have the same antecedent. The consequents are also the same, since their set of valuation functions coincide:
\begin{smallctabular}{@{}lcl@{}}
  $\newval{\A}{\F}{\opint{\vec{X}}{\vec{x}}} \in (\newval{\T}{\F}{\opint{\vec{X}}{\vec{x}}})^{\alpha}$
  & iff
  & $\newval{\A}{\F}{\opint{\vec{X}}{\vec{x}}} \in \newval{\T}{\F}{\opint{\vec{X}}{\vec{x}}}$ and $(\tuple{ \F_{\opint{\vec{X}}{\vec{x}}}, \newval{\T}{\F}{\opint{\vec{X}}{\vec{x}}} }, \newval{\A}{\F}{\opint{\vec{X}}{\vec{x}}}) \models \alpha$ \\
  & iff
  & $\A \in \T$ and $(\tuple{ \F_{\opint{\vec{X}}{\vec{x}}}, \newval{\T}{\F}{\opint{\vec{X}}{\vec{x}}} }, \newval{\A}{\F}{\opint{\vec{X}}{\vec{x}}}) \models \alpha$ \\
  & iff
  & $\A \in \T$ and $(\tuple{ \F, \T }, \A) \models \int{\opint{\vec{X}}{\vec{x}}}\alpha$ \\
  & iff
  & $\A \in \T^{\int{\opint{\vec{X}}{\vec{x}}}\alpha}$ \\
  & iff
  & $\newval{\A}{\F}{\opint{\vec{X}}{\vec{x}}} \in \newval{(\T^{\int{\opint{\vec{X}}{\vec{x}}}\alpha})}{\F}{\opint{\vec{X}}{\vec{x}}}$.
\end{smallctabular}
This completes this part of the proof.

\myparagraph{Completeness.} Again, by translation, now also using the following rules.

\begin{lemma}\label{lem:reka}
  ~
  \begin{enumerate}
    \item \ax{RE_!}:\; if $\vdashlofull \chi_1 \leftrightarrow \chi_2$ then $\vdashlofull [\alpha!]\chi_1 \leftrightarrow [\alpha!]\chi_2$,

    \item \ax{RE_{=}}:\; If $\vdashlofull \chi_1 \leftrightarrow \chi_2$ then $\vdashlofull \int{\opint{\vec{X}}{\vec{x}}}\chi_1 \leftrightarrow \int{\opint{\vec{X}}{\vec{x}}}\chi_2$.

    \item \ax{{=}!{=}}:\; $\vdashlofull \int{\opint{\vec{X}}{\vec{x}}}[\alpha!]\int{\opint{\vec{Y}}{\vec{y}}}\chi \,\leftrightarrow\, [\int{\opint{\vec{X}}{\vec{x}}}\alpha!]\int{\opint{\vec{X'}}{\vec{x'}}, \opint{\vec{Y}}{\vec{y}}}{\chi}$,\; with the expression $\opint{\vec{X'}}{\vec{x'}}$ the subassignment of $\opint{\vec{X}}{\vec{x}}$ for $\vec{X'} := \vec{X} \setminus \vec{Y}$.
  \end{enumerate}
  \begin{proof}
    Proofs for the first two rules are standard. For the third, while
    \begin{center}
      $\vdashlofull
          \int{\opint{\vec{X}}{\vec{x}}}[\alpha!]\int{\opint{\vec{Y}}{\vec{y}}}\chi
          \,\leftrightarrow\,
          [\int{\opint{\vec{X}}{\vec{x}}}\alpha!]\int{\opint{\vec{X}}{\vec{x}}}\int{\opint{\vec{Y}}{\vec{y}}}\chi$
    \end{center}
    is an instance of \ax{=_!}, from \ax{A_{[][]}} we get
    \begin{center}
      $\vdashlofull
          \int{\opint{\vec{X}}{\vec{x}}}{\int{\opint{\vec{Y}}{\vec{y}}}{\chi}}
          \,\leftrightarrow\,
          \int{\opint{\vec{X'}}{\vec{x'}}, \opint{\vec{Y}}{\vec{y}}}{\chi}$
    \end{center}
    with $\opint{\vec{X'}}{\vec{x'}}$ the subassignment of $\opint{\vec{X}}{\vec{x}}$ for $\vec{X'} := \vec{X} \setminus \vec{Y}$. Hence, using the second equivalence and \ax{RE_!} on the right-hand side of the first,
    \begin{center}
        $\vdashlofull
          [\int{\opint{\vec{X}}{\vec{x}}}\alpha!]\int{\opint{\vec{X}}{\vec{x}}}{\int{\opint{\vec{Y}}{\vec{y}}}{\chi}}
          \,\leftrightarrow\,
          [\int{\opint{\vec{X}}{\vec{x}}}\alpha!]\int{\opint{\vec{X'}}{\vec{x'}}, \opint{\vec{Y}}{\vec{y}}}{\chi}$.
    \end{center}
    Thus, by propositional reasoning from the first and the third,
    \begin{center}
      $\vdashlofull
        \int{\opint{\vec{X}}{\vec{x}}}[\alpha!]\int{\opint{\vec{Y}}{\vec{y}}}\chi
        \,\leftrightarrow\,
        [\int{\opint{\vec{X}}{\vec{x}}}\alpha!]\int{\opint{\vec{X'}}{\vec{x'}}, \opint{\vec{Y}}{\vec{y}}}{\chi}$.
    \end{center}
  \end{proof}
\end{lemma}

\noindent In this case, the translation uses two steps: \begin{inlineenum} \item from \LAfull to an intermediate language \LAfullres whose formulas $\theta$ are built by using of Boolean operators, $K$ and $[\alpha!]$ over `atoms' of the form $\int{\opint{\vec{X}}{\vec{x}}}Y{=}y$ (evaluated over epistemic causal models in the natural way), and \item from \LAfullres to the final target \LAkcres\end{inlineenum}.

\myparagraphit{Completeness, step 1: from \LAfull to \LAfullres.} The translation from \LAfull to \LAfullres uses axioms \ax{A_{[]}}, \ax{A_\lnot}, \ax{A_\land} and \ax{CM} for pushing intervention operators through $\lnot$, $\land$ and $K$, and axioms \ax{!_{=}}, \ax{!_\lnot}, \ax{!_\land}, \ax{!_K}, \ax{!_!} and \ax{{=}!{=}} for eliminating public announcement operators \emph{inside the scope of interventions}.

\begin{definition}[Translation $\trr$]\label{def:trr}
  Define $\trr:\LAfull \to \LAfullres$ as
  \begingroup
    \footnotesize
    \[
      \renewcommand{\arraystretch}{1.4}
      \begin{array}{@{}c@{}}
        \begin{array}{@{}c@{\qquad}c@{}}
          \begin{array}{@{}r@{\,:=\,}l@{}}
            \trr(Y{=}y)                                                   & \int{\;}Y{=}y \\
            \trr(\lnot\chi)                                               & \lnot \trr(\chi) \\
            \trr(\chi_1 \land \chi_2)                                     & \trr(\chi_1) \land \trr(\chi_2) \\
            \trr(K\chi)                                                   & K \trr(\chi) \\
            \trr([\alpha!]\chi)                                           & [\trr(\alpha)!]\trr(\chi) \\
          \end{array}
          &
          \begin{array}{@{}r@{\,:=\,}l@{}}
            \trr(\int{\opint{\vec{X}}{\vec{x}}}Y{=}y)                              & \int{\opint{\vec{X}}{\vec{x}}}Y{=}y \\
            \trr(\int{\opint{\vec{X}}{\vec{x}}}\lnot \chi)                         & \trr(\lnot \int{\opint{\vec{X}}{\vec{x}}}\chi) \\
            \trr(\int{\opint{\vec{X}}{\vec{x}}}(\chi_1 \land \chi_2))              & \trr(\int{\opint{\vec{X}}{\vec{x}}}\chi_1 \land \int{\opint{\vec{X}}{\vec{x}}}\chi_2) \\
            \trr(\int{\opint{\vec{X}}{\vec{x}}}K\chi)                              & \trr(K\int{\opint{\vec{X}}{\vec{x}}}\chi) \\
            \trr(\int{\opint{\vec{X}}{\vec{x}}}\int{\opint{\vec{Y}}{\vec{y}}}\chi) & \trr(\int{\opint{\vec{X'}}{\vec{x'}}, \opint{\vec{Y}}{\vec{y}}}\chi) \\
          \end{array}
        \end{array}
        \\[3.5em]
        \begin{array}{@{}r@{\,:=\,}l@{}}
          \trr(\int{\opint{\vec{X}}{\vec{x}}}[\alpha!]Y{=}y)                     & \trr(\int{\opint{\vec{X}}{\vec{x}}}(\alpha \rightarrow Y{=}y)) \\
          \trr(\int{\opint{\vec{X}}{\vec{x}}}[\alpha!]\lnot \chi)                & \trr(\int{\opint{\vec{X}}{\vec{x}}}(\alpha \rightarrow \lnot [\alpha!]\chi)) \\
          \trr(\int{\opint{\vec{X}}{\vec{x}}}[\alpha!](\chi_1 \land \chi_2))     & \trr(\int{\opint{\vec{X}}{\vec{x}}}([\alpha!]\chi_1 \land [\alpha!]\chi_2)) \\
          \trr(\int{\opint{\vec{X}}{\vec{x}}}[\alpha!]K\chi)                     & \trr(\int{\opint{\vec{X}}{\vec{x}}}(\alpha \rightarrow K(\alpha \rightarrow [\alpha!]\chi))) \\
          \trr(\int{\opint{\vec{X}}{\vec{x}}}[\alpha_1!][\alpha_2!]\chi) & \trr(\int{\opint{\vec{X}}{\vec{x}}}[\alpha_1 \land [\alpha_1!]\alpha_2!]\chi) \\
          \trr(\int{\opint{\vec{X}}{\vec{x}}}[\alpha!]\int{\opint{\vec{Y}}{\vec{y}}}\chi) & \trr([\int{\opint{\vec{X}}{\vec{x}}}\alpha!]\int{\opint{\vec{X'}}{\vec{x'}}, \opint{\vec{Y}}{\vec{y}}}{\chi}) \\
        \end{array}
      \end{array}
    \]
  \endgroup
  with $\opint{\vec{X'}}{\vec{x'}}$ the subassignment of $\opint{\vec{X}}{\vec{x}}$ for $\vec{X'} := \vec{X} \setminus \vec{Y}$.
\end{definition}

\begin{proposition}\label{pro:LAkatoLAkares}
  For every $\chi \in \LAfull$,
  \begin{multicols}{3}
    \begin{enumerate}
      \item\label{pro:LAkatoLAkares:trans} $\trr(\chi) \in \LAfullres$,
      \item\label{pro:LAkatoLAkares:syn} $\vdashlofull \chi \leftrightarrow \trr(\chi)$,
      \item\label{pro:LAkatoLAkares:sem} $\models \chi \leftrightarrow \trr(\chi)$.
    \end{enumerate}
  \end{multicols}
  \begin{proof}
    By the formulas' complexity $\com:\LAfull \to \Nat\setminus\set{0}$, this time defined as
    \begin{smallctabular}{c@{\qquad}c}
      \begin{tabular}[t]{r@{\;:=\;}l}
        $\com(Y{=}y)$             & $1$ \\
        $\com(\lnot \chi)$         & $1 + \com(\chi)$ \\
        $\com(\chi_1 \land \chi_2)$ & $1 + \max\set{\com(\chi_1), \com(\chi_2)}$ \\
      \end{tabular}
      &
      \begin{tabular}[t]{r@{\;:=\;}l}
        $\com(K\chi)$                                & $1 + \com(\chi)$ \\
        $\com([\alpha!]\chi)$                        & $(7+\com(\alpha))\com(\chi)$ \\
        $\com(\int{\opint{\vec{X}}{\vec{x}}}{\chi})$ & $2\com(\chi)$
      \end{tabular}
    \end{smallctabular}
    Thus\footnote{\citet[Definition 7.21]{vanDitmarschEtAl2007} use rather $\com([\alpha!]\chi):=(4+\com(\alpha))\com(\chi)$. Here, 7 is the least natural number guaranteeing $\com(\int{\opint{\vec{X}}{\vec{x}}}[\alpha!]K\chi) > \com(\int{\opint{\vec{X}}{\vec{x}}}(\alpha \rightarrow K(\alpha \rightarrow [\alpha!]\chi)))$.}, for all assignments $\opint{\vec{X}}{\vec{x}}$, $\opint{\vec{Y}}{\vec{y}}$ and all formulas $\chi, \chi_1, \chi_2 \in \LAfull$,
    \begingroup
      \small
      \begin{center}
        $\begin{array}{r@{\;}c@{\;}l}
          \toprule
          \com(\chi) & \geqslant & \com(\psi) \;\text{for every}\; \psi \in \sub(\chi), \\
          \midrule
          \com(\int{\opint{\vec{X}}{\vec{x}}}{\lnot \chi}) & > &\com(\lnot \int{\opint{\vec{X}}{\vec{x}}}{\chi}), \\
          \com(\int{\opint{\vec{X}}{\vec{x}}}{(\chi_1 \land \chi_2)}) & > & \com(\int{\opint{\vec{X}}{\vec{x}}}{\chi_1} \land \int{\opint{\vec{X}}{\vec{x}}}{\chi_2}) \\
          \com(\int{\opint{\vec{X}}{\vec{x}}}{K \chi}) & > & \com(K \int{\opint{\vec{X}}{\vec{x}}}{\chi}) \\
          \midrule
          \com(\int{\opint{\vec{X}}{\vec{x}}}[\alpha!]Y{=}y) & > & \com(\int{\opint{\vec{X}}{\vec{x}}}(\alpha \rightarrow Y{=}y)) \\
          \com(\int{\opint{\vec{X}}{\vec{x}}}[\alpha!]\lnot \chi) & > & \com(\int{\opint{\vec{X}}{\vec{x}}}(\alpha \rightarrow \lnot [\alpha!]\chi)) \\
          \com(\int{\opint{\vec{X}}{\vec{x}}}[\alpha!](\chi_1 \land \chi_2)) & > & \trr(\int{\opint{\vec{X}}{\vec{x}}}([\alpha!]\chi_1 \land [\alpha!]\chi_2)) \\
          \com(\int{\opint{\vec{X}}{\vec{x}}}[\alpha!]K\chi) & > & \com(\int{\opint{\vec{X}}{\vec{x}}}(\alpha \rightarrow K(\alpha \rightarrow [\alpha!]\chi))) \\
          \com(\int{\opint{\vec{X}}{\vec{x}}}[\alpha_1!][\alpha_2!]\chi) & > & \com(\int{\opint{\vec{X}}{\vec{x}}}[\alpha_1 \land [\alpha_1!]\alpha_2!]\chi) \\
          \com(\int{\opint{\vec{X}}{\vec{x}}}[\alpha!]\int{\opint{\vec{Y}}{\vec{y}}}\chi) & > & \com([\int{\opint{\vec{X}}{\vec{x}}}\alpha!]\int{\opint{\vec{X'}}{\vec{x'}}, \opint{\vec{Y}}{\vec{y}}}{\chi}) \hfill \text{ with } \opint{\vec{X'}}{\vec{x'}} \text{ the} \\
           & & \text{subassignment of } \opint{\vec{X}}{\vec{x}} \text{ for } \vec{X'} := \vec{X} \setminus \vec{Y} \\
          \com(\int{\opint{\vec{X}}{\vec{x}}}\int{\opint{\vec{Y}}{\vec{y}}}\chi) & > & \com(\int{\opint{\vec{X'}}{\vec{x'}}, \opint{\vec{Y}}{\vec{y}}}\chi) \\
          \bottomrule
        \end{array}$
      \end{center}
    \endgroup
    \noindent The first is proved by structural induction (with $\sub([\alpha!]\chi) := \set{[\alpha!]\chi} \cup \sub(\alpha) \cup \sub(\chi)$). Items in the second block are proved as their respective counterparts in Proposition \ref{pro:LAktoLAkres}. For items in the third block, all but the last two are consequences of their `interventionless' counterparts, which are analogous to the cases in public announcement logic \citep[Lemma 7.22]{vanDitmarschEtAl2007}. For the next to last, note that $\com(\int{\opint{\vec{X}}{\vec{x}}}[\alpha!]\int{\opint{\vec{Y}}{\vec{y}}}\chi) = (28 + 4\com(\alpha))\com(\chi)$, yet $\com([\int{\opint{\vec{X}}{\vec{x}}}\alpha!]\int{\opint{\vec{X'}}{\vec{x'}}, \opint{\vec{Y}}{\vec{y}}}{\chi}) = (14+4\com(\alpha)) \com(\chi)$. The last is straightforward.

    {\medskip}

    For the properties, the first two are proved by induction on $\com(\chi)$: base cases are for formulas $\chi$ with $\bs{\com(\chi)=1}$ (i.e., $Y{=}y$), and inductive cases are for formulas $\chi$ with $\bs{\com(\chi) > 1}$ (i.e., $\lnot \chi$,\, $\chi_1 \land \chi_2$,\, $K \chi$,\, $[\alpha!]\chi$,\, $\int{\opint{\vec{X}}{\vec{x}}}{Y{=}y}$,\, $\int{\opint{\vec{X}}{\vec{x}}}{\lnot \chi}$,\, $\int{\opint{\vec{X}}{\vec{x}}}{(\chi_1 \land \chi_2)}$,\, $\int{\opint{\vec{X}}{\vec{x}}}{K \chi}$,\, $\int{\opint{\vec{X}}{\vec{x}}}[\alpha!]Y{=}y$,\, $\int{\opint{\vec{X}}{\vec{x}}}[\alpha!]\lnot \chi$,\, $\int{\opint{\vec{X}}{\vec{x}}}[\alpha!](\chi_1 \land \chi_2)$,\, $\int{\opint{\vec{X}}{\vec{x}}}[\alpha!]K\chi$, $\int{\opint{\vec{X}}{\vec{x}}}[\alpha_1!][\alpha_2!]\chi$, $\int{\opint{\vec{X}}{\vec{x}}}[\alpha!]\int{\opint{\vec{Y}}{\vec{y}}}\chi$ and $\int{\opint{\vec{X}}{\vec{x}}}\int{\opint{\vec{Y}}{\vec{y}}}\chi$).

    \begin{compactenumerate}
      \item The proof is analogous to that in Proposition \ref{pro:LAktoLAkres}. The base case is straightforward, and the inductive cases use the above-listed properties of $\com$ to generate appropriate IHs, then using the definition of $\trr$.

      \item The proof is analogous to that in Proposition \ref{pro:LAktoLAkres}. The base case follows from axiom \ax{A_{[]}}. The inductive cases rely on the properties of $\com$ to generate appropriate IHs, and then use the definition of $\trr$ with propositional reasoning (cases $\lnot \chi$, $\chi_1 \land \chi_2$ and $\int{\opint{\vec{X}}{\vec{x}}}{Y{=}y}$) plus \ax{RE_K} (case $K \chi$), \ax{_!RE} and \ax{RE_!} from Lemma \ref{lem:reka} (case $[\alpha!]\chi$),\footnote{Indeed, since $\alpha, \chi \in \sub([\alpha!]\chi)$, from IH it follows that $\vdashlofull \alpha \leftrightarrow \trr(\alpha)$ and $\vdashlofull \chi \leftrightarrow \trr(\chi)$. From the first and \ax{_!RE} it follows that $\vdashlofull [\alpha!]\chi \leftrightarrow [\trr(\alpha)!]\chi$; from the second and \ax{RE_!} it follows that $\vdashlofull [\trr(\alpha)!]\chi \leftrightarrow [\trr(\alpha)!]\trr(\chi)$. Then, by propositional reasoning (\ax{P} and \ax{MP}), $\vdashlofull [\alpha!]\chi \leftrightarrow [\trr(\alpha)!]\trr(\chi)$, which by definition of $\trr$ is the required $\vdashlofull [\alpha!]\chi \leftrightarrow \trr([\alpha!]\chi)$.} axiom \ax{A_\lnot} (case $\int{\opint{\vec{X}}{\vec{x}}}{\lnot \chi}$), axiom \ax{A_\land} (case $\int{\opint{\vec{X}}{\vec{x}}}{(\chi_1 \land \chi_2)}$ and axiom \ax{CM} (case $\int{\opint{\vec{X}}{\vec{x}}}{K \chi}$). Cases $\int{\opint{\vec{X}}{\vec{x}}}[\alpha!]Y{=}y$,\; $\int{\opint{\vec{X}}{\vec{x}}}[\alpha!]\lnot \chi$,\; $\int{\opint{\vec{X}}{\vec{x}}}[\alpha!](\chi_1 \land \chi_2)$,\; $\int{\opint{\vec{X}}{\vec{x}}}[\alpha!]K\chi$ \;and\; $\int{\opint{\vec{X}}{\vec{x}}}[\alpha_1!][\alpha_2!]\chi$ follow the same pattern: use the properties of $\com$ to obtain an appropriate IH, then apply \ax{RE_{=}} (Lemma \ref{lem:reka}) over the corresponding axiom, and finally use the definition of $\trr$.\footnote{As an example, consider $\int{\opint{\vec{X}}{\vec{x}}}[\alpha!]Y{=}y$. Since $\com(\int{\opint{\vec{X}}{\vec{x}}}[\alpha!]Y{=}y) > \com(\int{\opint{\vec{X}}{\vec{x}}}(\alpha \rightarrow Y{=}y))$, from IH it follows that $\vdashlofull \int{\opint{\vec{X}}{\vec{x}}}(\alpha \rightarrow Y{=}y) \leftrightarrow \trr(\int{\opint{\vec{X}}{\vec{x}}}(\alpha \rightarrow Y{=}y))$. But applying \ax{RE_{=}} on \ax{!_=} yields $\vdashlofull \int{\opint{\vec{X}}{\vec{x}}}[\alpha!]Y{=}y \leftrightarrow \int{\opint{\vec{X}}{\vec{x}}}(\alpha \rightarrow Y{=}y)$, so $\vdashlofull \int{\opint{\vec{X}}{\vec{x}}}[\alpha!]Y{=}y \leftrightarrow \trr(\int{\opint{\vec{X}}{\vec{x}}}(\alpha \rightarrow Y{=}y))$. Then, from the definition of $\trr$, it follows that $\vdashlofull \int{\opint{\vec{X}}{\vec{x}}}[\alpha!]Y{=}y \leftrightarrow \trr(\int{\opint{\vec{X}}{\vec{x}}}[\alpha!]Y{=}y)$. As another example consider $\int{\opint{\vec{X}}{\vec{x}}}[\alpha_1!][\alpha_2!]\chi$. Since $\com(\int{\opint{\vec{X}}{\vec{x}}}[\alpha_1!][\alpha_2!]\chi) > \com(\int{\opint{\vec{X}}{\vec{x}}}[\alpha_1 \land [\alpha_1!]\alpha_2!]\chi)$, from IH it follows that $\vdashlofull \int{\opint{\vec{X}}{\vec{x}}}[\alpha_1 \land [\alpha_1!]\alpha_2!]\chi \leftrightarrow \trr(\int{\opint{\vec{X}}{\vec{x}}}[\alpha_1 \land [\alpha_1!]\alpha_2!]\chi)$. But applying \ax{RE_{=}} on \ax{!_!} yields $\vdashlofull \int{\opint{\vec{X}}{\vec{x}}}[\alpha_1!][\alpha_2!]\chi \leftrightarrow \int{\opint{\vec{X}}{\vec{x}}}[\alpha_1 \land [\alpha_1!]\alpha_2!]\chi$, so $\vdashlofull \int{\opint{\vec{X}}{\vec{x}}}[\alpha_1!][\alpha_2!]\chi \leftrightarrow \trr(\int{\opint{\vec{X}}{\vec{x}}}[\alpha_1 \land [\alpha_1!]\alpha_2!]\chi)$. Then, from the definition of $\trr$, it follows that $\vdashlofull \int{\opint{\vec{X}}{\vec{x}}}[\alpha_1!][\alpha_2!]\chi \leftrightarrow \trr(\int{\opint{\vec{X}}{\vec{x}}}[\alpha_1 \land [\alpha_1!]\alpha_2!]\chi)$.} The case $\int{\opint{\vec{X}}{\vec{x}}}\int{\opint{\vec{Y}}{\vec{y}}}\chi$ relies on axiom \ax{A_{[][]}}, just as its counterpart in Proposition \ref{pro:LAtoLAres}. For the lone case left,
      \begin{compactitemize}
        \item Case $\int{\opint{\vec{X}}{\vec{x}}}[\alpha!]\int{\opint{\vec{Y}}{\vec{y}}}\chi$. Use again IH, now on $\com(\int{\opint{\vec{X}}{\vec{x}}}[\alpha!]\int{\opint{\vec{Y}}{\vec{y}}}\chi) > \com([\int{\opint{\vec{X}}{\vec{x}}}\alpha!]\int{\opint{\vec{X'}}{\vec{x'}}, \opint{\vec{Y}}{\vec{y}}}{\chi})$ (with $\opint{\vec{X'}}{\vec{x'}}$ as indicated above) to obtain $\vdashlofull [\int{\opint{\vec{X}}{\vec{x}}}\alpha!]\int{\opint{\vec{X'}}{\vec{x'}}, \opint{\vec{Y}}{\vec{y}}}{\chi} \leftrightarrow \trr([\int{\opint{\vec{X}}{\vec{x}}}\alpha!]\int{\opint{\vec{X'}}{\vec{x'}}, \opint{\vec{Y}}{\vec{y}}}{\chi})$. But $\vdashlofull \int{\opint{\vec{X}}{\vec{x}}}[\alpha!]\int{\opint{\vec{Y}}{\vec{y}}}\chi \,\leftrightarrow\, [\int{\opint{\vec{X}}{\vec{x}}}\alpha!]\int{\opint{\vec{X'}}{\vec{x'}}, \opint{\vec{Y}}{\vec{y}}}{\chi}$ (axiom \ax{{=}!{=}}; see Lemma \ref{lem:reka}). Then, by propositional reasoning, it follows that $\vdashlofull \int{\opint{\vec{X}}{\vec{x}}}[\alpha!]\int{\opint{\vec{Y}}{\vec{y}}}\chi \,\leftrightarrow\, \trr([\int{\opint{\vec{X}}{\vec{x}}}\alpha!]\int{\opint{\vec{X'}}{\vec{x'}}, \opint{\vec{Y}}{\vec{y}}}{\chi})$. This is enough, as the right-hand side of the latter is, by definition, $\trr(\int{\opint{\vec{X}}{\vec{x}}}[\alpha!]\int{\opint{\vec{Y}}{\vec{y}}}\chi)$.
      \end{compactitemize}
      \item By the previous item, $\vdashlofull \chi \leftrightarrow \trr(\chi)$. But \LOfull is sound for recursive causal models; therefore, $\models \chi \leftrightarrow \trr(\chi)$.
    \end{compactenumerate}
  \end{proof}
\end{proposition}

\myparagraphit{Completeness, step 2: from \LAfullres to \LAkcres.} The translation from \LAfullres to \LAkcres uses \ax{!_{=}}, \ax{!_\lnot}, \ax{!_\land} and \ax{!_K} for eliminating public announcements.

\begin{definition}[Translation $\trf$]\label{def:trf}
  Define $\trf:\LAfullres \to \LAkcres$ as
  \begingroup
    \footnotesize
    \[
      \renewcommand{\arraystretch}{1.4}
      \begin{array}{@{}c@{}c@{}}
        \begin{array}{@{}r@{\,:=\,}l@{}}
          \trf(\int{\opint{\vec{X}}{\vec{x}}}Y{=}y)                     & \int{\opint{\vec{X}}{\vec{x}}}Y{=}y \\
          \trf(\lnot\theta)                                             & \lnot \trf(\theta) \\
          \trf(\theta_1 \land \theta_2)                                 & \trf(\theta_1) \land \trf(\theta_2) \\
          \trf(K\theta)                                                 & K \trf(\theta) \\
        \end{array}
        &
        \begin{array}{@{}r@{\,:=\,}l@{}}
          \trf([\alpha!]\int{\opint{\vec{X}}{\vec{x}}}Y{=}y)                        & \trf(\alpha \rightarrow \int{\opint{\vec{X}}{\vec{x}}}Y{=}y) \\
          \trf([\alpha!]\lnot \theta)                & \trf(\alpha \rightarrow \lnot [\alpha!]\theta) \\
          \trf([\alpha!](\theta_1 \land \theta_2))   & \trf([\alpha!]\theta_1 \land [\alpha!]\theta_2) \\
          \trf([\alpha!]K\theta)                     & \trf(\alpha \rightarrow K(\alpha \rightarrow [\alpha!]\theta)) \\
          \trf([\alpha_1!][\alpha_2!]\theta) & \trf([\alpha_1 \land [\alpha_1!]\alpha_2!]\theta) \\
        \end{array}
      \end{array}
    \]
  \endgroup
\end{definition}

\begin{proposition}\label{pro:LAkarestoLAkres}
  For every $\theta \in \LAfullres$,
  \begin{multicols}{3}
    \begin{enumerate}
      \item\label{pro:LAkarestoLAkres:trans} $\trf(\theta) \in \LAkcres$,
      \item\label{pro:LAkarestoLAkres:syn} $\vdashlofull \theta \leftrightarrow \trf(\theta)$,
      \item\label{pro:LAkarestoLAkres:sem} $\models \theta \leftrightarrow \trf(\theta)$.
    \end{enumerate}
  \end{multicols}
  \begin{proof}
    By induction $\com:\LAfullres \to \Nat\setminus\set{0}$, this time defined as
    \begin{ctabular}{c@{\qquad}c}
      \begin{tabular}[t]{r@{\;:=\;}l}
        $\com(\int{\opint{\vec{X}}{\vec{x}}}Y{=}y)$    & $1$ \\
        $\com(\lnot \theta)$                           & $1 + \com(\theta)$ \\
        $\com(\theta_1 \land \theta_2)$                & $1 + \max\set{\com(\theta_1), \com(\theta_2)}$ \\
      \end{tabular}
      &
      \begin{tabular}[t]{r@{\;:=\;}l}
        $\com(K\theta)$                                & $1 + \com(\theta)$ \\
        $\com([\alpha!]\theta)$                        & $(7+\com(\alpha))\com(\theta)$ \\
      \end{tabular}
    \end{ctabular}
    Thus, for every assignment $\opint{\vec{X}}{\vec{x}}$ and all formulas $\theta, \theta_1, \theta_2 \in \LAfullres$,
    \begingroup
      \small
      \[
        \begin{array}{c}
          \toprule
          \begin{array}{@{}r@{\;}c@{\;}l@{}}
            \com(\theta) & \geqslant & \com(\psi) \;\text{for every}\; \psi \in \sub(\theta), \\
          \end{array}
          \\
          \midrule
          \begin{array}{cc}
            \begin{array}{@{}r@{\;}c@{\;}l@{}}
              \com([\alpha!]\int{\opint{\vec{X}}{\vec{x}}}Y{=}y) & > & \com(\alpha \rightarrow \int{\opint{\vec{X}}{\vec{x}}}Y{=}y) \\
              \com([\alpha!]\lnot \theta) & > & \com(\alpha \rightarrow \lnot [\alpha!]\theta) \\
              \com([\alpha!](\theta_1 \land \theta_2)) & > & \com([\alpha!]\theta_1 \land [\alpha!]\theta_2) \\
            \end{array}
            &
            \begin{array}{@{}r@{\;}c@{\;}l@{}}
              \com([\alpha!]K\theta) & > & \com(\alpha \rightarrow K(\alpha \rightarrow [\alpha!]\theta)) \\
              \com([\alpha_1!][\alpha_2!]\theta) & > & \com([\alpha_1 \land [\alpha_1!]\alpha_2!]\theta) \\ \\
            \end{array}
          \end{array}
          \\
          \bottomrule
        \end{array}
      \]
    \endgroup
    The first item is proved by structural induction; the rest are small variations of the well-known cases from public announcement logic \citep[Section 7.4]{vanDitmarschEtAl2007}.

    The first two proofs use induction on $\com(\theta)$. The base cases are for formulas $\theta$ with $\bs{\com(\theta)=1}$ (i.e., $\int{\opint{\vec{X}}{\vec{x}}}Y{=}y$); the inductive cases are for formulas $\theta$ with $\bs{\com(\theta) > 1}$ (i.e., $\lnot \theta$,\; $\theta_1 \land \theta_2$,\; $K \theta$,\; $[\alpha!]\int{\opint{\vec{X}}{\vec{x}}}Y{=}y$,\; $[\alpha!]\lnot \theta$,\; $[\alpha!](\theta_1 \land \theta_2)$,\; $[\alpha!]K\theta$ and $[\alpha_1!][\alpha_2!]\theta$).
    \begin{compactenumerate}
      \item Analogous to that in Proposition \ref{pro:LAktoLAkres}. The base case is straightforward. For the inductive part, use the properties of $\com$ to generate appropriate IHs, and then use the definition of $\trf$.

      \item Analogous to that in Proposition \ref{pro:LAktoLAkres}. The base case is by propositional reasoning. The inductive cases rely on the properties of $\com$ to generate appropriate IHs, and then use the definition of $\trf$ with propositional reasoning (cases $\lnot \theta$, $\theta_1 \land \theta_2$) plus \ax{RE_K} (case $K \theta$), axiom \ax{!_{=}} (case $[\alpha!]\int{\opint{\vec{X}}{\vec{x}}}Y{=}y$), axiom \ax{!_\lnot} (case $[\alpha!]\lnot \theta$), axiom \ax{!_\land} (case $[\alpha!](\theta_1 \land \theta_2)$,\; axiom \ax{!_K} (case $[\alpha!]K\theta$) and axiom \ax{!_!} (case $[\alpha_1!][\alpha_2!]\theta$).\footnote{As an example, consider $[\alpha!]\int{\opint{\vec{X}}{\vec{x}}}Y{=}y$. Since $\com([\alpha!]\int{\opint{\vec{X}}{\vec{x}}}Y{=}y) > \com(\alpha \rightarrow \int{\opint{\vec{X}}{\vec{x}}}Y{=}y)$, from IH it follows that $\vdashlofull (\alpha \rightarrow \int{\opint{\vec{X}}{\vec{x}}}Y{=}y) \leftrightarrow \trf(\alpha \rightarrow \int{\opint{\vec{X}}{\vec{x}}}Y{=}y)$. But \ax{!_=} yields $\vdashlofull [\alpha!]\int{\opint{\vec{X}}{\vec{x}}}Y{=}y \leftrightarrow (\alpha \rightarrow \int{\opint{\vec{X}}{\vec{x}}}Y{=}y)$, so $\vdashlofull [\alpha!]\int{\opint{\vec{X}}{\vec{x}}}Y{=}y \leftrightarrow \trf(\alpha \rightarrow \int{\opint{\vec{X}}{\vec{x}}}Y{=}y)$. Then, from the definition of $\trf$, it follows that $\vdashlofull [\alpha!]\int{\opint{\vec{X}}{\vec{x}}}Y{=}y \leftrightarrow \trf([\alpha!]\int{\opint{\vec{X}}{\vec{x}}}Y{=}y)$. As another example, consider $[\alpha_1!][\alpha_2!]\theta$. Since $\com([\alpha_1!][\alpha_2!]\theta) > \com([\alpha_1 \land [\alpha_1!]\alpha_2!]\theta)$, from IH it follows that $\vdashlofull [\alpha_1 \land [\alpha_1!]\alpha_2!]\theta \leftrightarrow \trf([\alpha_1 \land [\alpha_1!]\alpha_2!]\theta)$. But \ax{!_!} yields $\vdashlofull [\alpha_1!][\alpha_2!]\theta \leftrightarrow [\alpha_1 \land [\alpha_1!]\alpha_2!]\theta$, so $\vdashlofull [\alpha_1!][\alpha_2!]\theta \leftrightarrow \trf([\alpha_1 \land [\alpha_1!]\alpha_2!]\theta)$. Then, from the definition of $\trf$, it follows that $\vdashlofull [\alpha_1!][\alpha_2!]\theta \leftrightarrow \trf([\alpha_1!][\alpha_2!]\theta)$.}

      \item By the previous item, $\vdashlofull \theta \leftrightarrow \trf(\theta)$. But \LOfull is sound within recursive causal models; therefore, $\models \theta \leftrightarrow \trf(\theta)$.
    \end{compactenumerate}
  \end{proof}
\end{proposition}

\noindent Then, for strong completeness of \LOfull for formulas in \LAfull, define $\tr:\LAfull \to \LAkcres$ as $\tr(\chi') := \trf(\trr(\chi'))$. Note: $\tr$ is properly defined because $\trr(\chi') \in \LAfullres$ (Proposition \ref{pro:LAkatoLAkares}.\ref{pro:LAkatoLAkares:trans}). Moreover, for every $\chi' \in \LAfull$,
\begin{itemize}
  \item $\tr(\chi') \in \LAkcres$ (from Proposition \ref{pro:LAkarestoLAkres}.\ref{pro:LAkarestoLAkres:trans}).

  \item $\vdashlofull \chi' \leftrightarrow \tr(\chi')$ (from Proposition \ref{pro:LAkatoLAkares}.\ref{pro:LAkatoLAkares:syn}, Proposition \ref{pro:LAkatoLAkares}.\ref{pro:LAkatoLAkares:trans} and Proposition \ref{pro:LAkarestoLAkres}.\ref{pro:LAkarestoLAkres:syn}).

  \item $\models \chi' \leftrightarrow \tr(\chi')$ (from Proposition \ref{pro:LAkatoLAkares}.\ref{pro:LAkatoLAkares:sem}, Proposition \ref{pro:LAkatoLAkares}.\ref{pro:LAkatoLAkares:trans} and Proposition \ref{pro:LAkarestoLAkres}.\ref{pro:LAkarestoLAkres:sem}).
\end{itemize}

With these three properties, the argument for strong completeness of \LOfull for formulas in \LAfull is as that for Theorem \ref{thm:rcm-LA} (Page \pageref{proof:thm:rcm-LA}).

\end{appendix:short}

\subsection{Proof of Theorem \ref{thm: soundness completeness LOfullo}}\label{app:LOfullo}

\begin{appendix:full}
Our goal is now to show the soundness and completeness of \LOfullo over epistemic causal models with observables.

\myparagraph{Soundness.} The soundness of most axioms and rules over epistemic causal models with observables is straightforward. We will only explicitly consider the axioms 
from Table \ref{tbl:rcm-LA} and \ax{PD}. 
Since the axioms of Table \ref{tbl:rcm-LA} are valid over causal models, they are taken care of by the following lemma:

\begin{lemma}
Let $\phi$ be an \LAfull formula without occurrences of $K$ or announcement operators. Let $\ec= \langle\Sig,\F,\T\rangle$ be an epistemic causal model (with observables), and $\A\in\T$. Then:
\[
(\ec,\A)\models^{O}\phi  \qiffq \langle\A,\F\rangle \models\phi
\]
\end{lemma}

\begin{proof}
A straightforward induction on $\phi$.
\end{proof}

Let us now take care of axiom \ax{PD}.

\begin{lemma}\label{lemma_xxoo}
 Let $\ec=\tuple{\Sig,\F,\T}$ be an epistemic causal model with observables; take $\A\in\T$. Let $\vec o = \A^\F_{\vec X=\vec x}(\O)$. Write $\delta$ for $[\vec X=\vec x]\O=\vec o$. Then, for all $\B\in\T^\delta$,
 \[
(\ec^\delta)^\B_{\vec X=\vec x}=(\ec^\delta)^\A_{\vec X=\vec x}= \ec^\A_{\vec X=\vec x} = \ec^\B_{\vec X=\vec x}.
 \]
\end{lemma}

\begin{proof}
Since $\B\in\T^\delta$, we have $(\ec^\delta,\B)\models [\vec X=\vec x]\O=\vec o$; then it follows that $((\ec^\delta)^\B_{\vec X=\vec x},\B^\F_{\vec X=\vec x})\models \O=\vec o$, which implies $\B^\F_{\vec X=\vec x}(\O)=\vec o=\A^\F_{\vec X=\vec x}(\O)$. This entails the first equality. The third equality is proved in the same way, using $\ec$ instead of $\ec^\delta$.

For the second equality, observe that any element of $\T^{\F,\A}_{\vec X=\vec x}$ is of the form $\C^\F_{\vec X=\vec x}$ for some $\C\in \T$. However, by the definition of epistemic model with observables, $\C^\F_{\vec X=\vec x}(\O)=\vec o$; so $(\ec^\A_{\vec X=\vec x},\C^\F_{\vec X=\vec x}) \models \O=\vec o$. Therefore, $(\ec,\C) \models [\vec X=\vec x]\O=\vec o$, i.e., $\C\in \T^\delta$. Since $\C^\F_{\vec X=\vec x}(\O)=\vec o$, then, $\C^\F_{\vec X=\vec x}\in (\T^\delta)^{\F,\A}_{\vec X=\vec x}$.
\end{proof}




\begin{theorem}
  Axiom \ax{PD} is sound on epistemic causal models w/ observables.
\end{theorem}

\begin{proof}
If case $\vec X$ is the empty set, axiom \ax{PD} reduces to:
\[
K\gamma \leftrightarrow \bigvee_{\vec o\in \Ran(O)}( \O = \vec o \land [\O = \vec o!]K\gamma).
\]
($\bs{\Rightarrow}$) Assume $(\ec,\A)\models K\gamma$. Let $\vec o=\A(\O)$. Then $(\ec,\A)\models \O = \vec o$. By definition of epistemic causal models with observables, then, $(\ec,\B)\models \O=\vec o$ for all $\B\in\T$. So $\T^{ \O=\vec o}=\T$, therefore $(\ec^{ \O=\vec o},\A)\models K\gamma$. Thus $(\ec,\A)\models [\O=\vec o!]K\gamma$. ($\bs{\Leftarrow}$) Assume $(\ec,\A)\models \bigvee_{\vec o\in \mathcal O}( \mathcal O = \vec o \land [\O = \vec o!]K\gamma)$. Then for some $\vec o \in \Ran(\O)$, $(\ec,\A)\models \O = \vec o$ and $(\ec,\A)\models [\O=\vec o!]K\gamma$. From the latter we obtain $(\ec^{ \O=\vec o},\A)\models K\gamma$, and from the former, as before, $\ec=\ec^{\O=\vec o}$; thus $(\ec,\A)\models K\gamma$.

Now, suppose now $\vec X$ is non-empty.

\noindent($\Rightarrow$) Let $\ec=\tuple{\Sig,\T,\F}$ and assume $(\ec,\A)\models [\vec X = \vec x]K\gamma$. We have to show that the right-hand side of axiom \ax{PD} holds for some $\vec o\in \Ran(\O)$; we will show it for $\vec o:= \A_{\vec X = \vec x}(\O)$. From the assumption we obtain:
\[
(\star): (\ec^\A_{\vec X = \vec x},\A^\F_{\vec X = \vec x})\models K\gamma.
\]
i.e., for every $\B\in \T^{\F,\A}_{\vec X = \vec x}$ we have $(\ec^\A_{\vec X = \vec x},\B)\models \gamma$. 
Notice that $(\ec^\A_{\vec X = \vec x},\A^\F_{\vec X = \vec x})\models \O = \vec o$; thus $(\ec,\A)\models [\vec X = \vec x]\O = \vec o$. This is the first conjunct we needed to prove.

Now write $\delta$ for $[\vec X = \vec x]\O = \vec o$. Let $\C\in \T^\delta$; we want to verify that $(\ec^\delta,\C)\models [\vec X=\vec x]\gamma$. 
By Lemma \ref{lemma_xxoo} we have $(\ec^\delta)^\C_{\vec X = \vec x} = 
\ec^\A_{\vec X = \vec x}$.
So $(\ec^\delta,\C)\models [\vec X=\vec x]\gamma \iff ((\ec^\delta)^\C_{\vec X = \vec x},\C^\F_{\vec X = \vec x})\models \gamma \iff (\ec^\A_{\vec X = \vec x},\C^\F_{\vec X = \vec x})\models \gamma$, which is true by ($\star$).
Since $(\ec^\delta,\C)\models [\vec X=\vec x]\gamma$ for each $\C\in \ec^\delta$, we have  $(\ec^\delta,\A)\models K[\vec X=\vec x]\gamma$. We then conclude $(\ec,\A)\models [\delta!]K[\vec X=\vec x]\gamma$, as needed.




\noindent($\Leftarrow$) Suppose there is an $\vec o\in\O$ such that $(\ec,\A)\models [\vec X = \vec x]\O = \vec o$ and $(\ec,\A)\models [\delta!]K[\vec X = \vec x]\gamma$.
For any $\C\in\T_{\vec X=\vec x}^{\F,\A}$ , $\C=\B^{\F}_{\vec{X}=\vec{x}}$ for some $\B\in\T$. Since $\B^{\F}_{\vec{X}=\vec{x}}(\O)=\vec{o}$, $\B\in \T^\delta$. So by Lemma \ref{lemma_xxoo}, $(\ec^\delta)_{\vec X=\vec x}^\B= \ec_{\vec X=\vec x}^\A$. By assumption, $(\ec^{\delta},\A)\models K[\vec{X}=\vec{x}]\gamma$, so $((\ec^\delta)^\B_{\vec{X}=\vec{x}},\B^\F_{\vec X=\vec x})\models\gamma$. Therefore $(\ec_{\vec X=\vec x}^\A,\B^\F_{\vec X=\vec x})\models\gamma$. So $(\ec_{\vec X=\vec x}^\A,\C)\models\gamma$. Since $\C$ is an arbitrary assignment in $\T_{\vec X=\vec x}^{\F,\A}$, we have $(\ec^\A_{\vec{X}=\vec{x}},\A^\F_{\vec{X}=\vec{x}})\models K\gamma$, namely $(\ec,\A)\models [\vec{X}=\vec{x}]K\gamma$.
\end{proof}

\myparagraph{Completeness via reduction to the case without observables} We want to prove the completeness of $\LOfullo$ over epistemic causal models with observables. The first step is the elimination of the announcement operators, which can be done using axioms \ax{!_=}, \ax{!_\lnot}, \ax{!_\land} and \ax{!_K} and \ax{PD}, together with replacement of equivalents. Regarding replacement, we can summarize all the replacement rules proposed before by the following:
\begin{center}
  $\ax{RE_{full}}:\; \text{ if } \vdash \chi_1 \leftrightarrow \chi_2 \text{ then } \vdash \phi \leftrightarrow \phi[\chi_2/\chi_1]$
\end{center}
where $\phi[\chi_2/\chi_1]$ is a formula obtained by replacing, in $\phi$, some occurrences of $\chi_1$ with $\chi_2$.


\begin{proposition}
The rule \ax{RE_{full}} is admissible in \LOfullo.
\end{proposition}

\begin{proof} (Sketch)
First observe that the rules  \ax{RE_K}, \ax{RE_=}, \ax{RE_!}, \ax{_!RE} are provable exactly in the same way as in \LOfull. The main claim can then be proved by induction on $\phi$. The cases for $\phi = K\psi$ (resp. $[\vec X=\vec x]\psi$, $[\alpha!]\psi$) are then covered by \ax{RE_K} (resp. \ax{RE_=}, \ax{_!RE}+\ax{RE_!}). The boolean cases are proved using classical logic (\ax{P}+\ax{MP}) and the base case is trivial.
\end{proof}

\begin{proposition}\label{prop_reduction}
  \begin{enumerate}
  \item Every formula $\phi \in \LAfull$ is logically equivalent to a formula $\xi_\phi \in \LAfullres$. Moreover, $\phi \leftrightarrow \xi_\phi$ is derivable in \LOfullo.
  \item Every formula $\xi \in \LAfullres$ is logically equivalent to a formula $\chi_\xi \in \LAkcres$. Moreover, $\xi \leftrightarrow \chi_\xi$ is derivable in \LOfullo.
  \end{enumerate}
\end{proposition}

\begin{proof}
(i): modify the translation ${\trr}$ (Definition \ref{def:trr}) by:
\[
\trr([\vec X=\vec x]K\gamma):= \bigvee_{\vec o\in \mathcal O}\Big( [\vec X = \vec x] \mathcal O = \vec o \land \big[[\vec X = \vec x] \mathcal O = \vec o ! \big]K[\vec X = \vec x]{\trr}(\gamma)\Big)
\]
The correctness of this clause is justified using axiom \ax{PD} and rule \ax{RE}. This clause allows removing instances of $K$ from the consequents of counterfactuals.

(ii) is proved by the same translation as in the case without observables (Definition \ref{def:trf}); see the proof of Proposition \ref{pro:LAkarestoLAkres}.
\end{proof}

In order to show the completeness of \LOfullo, by Proposition \ref{prop_reduction} it suffices to show that $\LOkco:= \LOfullo \setminus \{\ax{A_{[]}}, \ax{A_\lnot},\ax{A_\land},\ax{A_{[][]}}, \ax{!_=}, \ax{!_\lnot}, \ax{!_\land}, \ax{!_K}, \ax{PD}
\}$
is complete for the language \LAkcres (in the semantics with observables). Notice that $\LOkco = \LOkc\cup \{\ax{OC}\}$; therefore we have:

\begin{proposition}\label{prop: reduction to LOfull}
Let $\Gamma\subseteq \LAkcres$. Then
\[
\Gamma \text{ is \LOkco-consistent } \qiffq \Gamma \cup \{\ax{OC}\} \text{ is \LOkc-consistent.}
\]
\end{proposition}

Thus, in order to find a model for a \LOkco -consistent set of \LAkcres formulas $\Gamma$, we use the fact that the completeness theorem for the case without observables provides a pointed model for $\Gamma\cup \{\ax{OC}\}$, i.e. an epistemic causal model $\ec$ together with an assignment $\A$ such that $(\ec,\A)\models^W \Gamma\cup \{\ax{OC}\}$. If we manage to prove that $(\ec^*,\A)\models^O \Gamma\cup \{\ax{OC}\}$ (where $\ec^*$ differs from $\ec$ only in that its signature has a set of observables), we are done. But this is provided by the following result. We write here $\ax{OC}_{\Sig^*}$ for the axiom scheme \ax{OC} specialized to the signature $\Sig^*$.

\begin{proposition}\label{prop: OC models have observables}
Let $\Sig=\tuple{\XV,\NV,\Ran}$ be a signature and  $\ec = \tuple{\Sig,\T,\F}$ an epistemic causal model. Let $\O$ be a subset of $\XV\cup\NV$ and $\Sig^*=\tuple{\XV,\NV, \O, \Ran}$ be the corresponding signature with observables.

Now suppose that, for some $\A\in\T$, $(\ec,\A)\models^W \ax{OC}_{\Sig^*}$. Then:
\begin{enumerate}
\item The tuple $\O$ takes constant value in $\T$;
 therefore $\ec^*= \tuple{\Sig,\T,\F}$ is an epistemic causal model with observables $\O$. 
\item For all $\B\in\T$ and all $\varphi\in$\LAfull: $(\ec,\B)\models^W \varphi \iff (\ec^*,\B)\models^O \varphi$
\end{enumerate}
\end{proposition}


\begin{proof}
1) The constancy of $\O$ in $\T$ follows immediately from the fact that $(\ec,\A)\models \bigvee_{\vec o\in \mathcal O}K\O=\vec o$, the special instance of \ax{OC}$_{\Sig^*}$ for $\vec X=\emptyset$.

2) By induction on $\varphi$. The only nontrivial case is $\varphi= [\vec X=\vec x]\gamma$. We have $(\ec,\B)\models^W [\vec X=\vec x]\gamma$ iff  $(\ec_{\vec X=\vec x},\B^\F_{\vec X=\vec x})\models^W \gamma$. Now since $(\ec,\A)\models^W [\vec X=\vec x]\bigvee_{\vec o\in \O}K\O=\vec o$, we have $(\ec_{\vec X=\vec x},\A^\F_{\vec X= \vec x})\models^W K\O=\vec o$ for some value $\vec o\in\Ran(O)$; therefore  $(\ec_{\vec X=\vec x})^*$, the epistemic causal model with observables that differs from $\ec_{\vec X=\vec x}$ only in that it has signature $\Sig^*$ instead of signature $\Sig$, is an epistemic causal model with observables $\O$. So we can apply the inductive hypothesis to obtain (*): $((\ec_{\vec X=\vec x})^*,\B^\F_{\vec X=\vec x})\models^O \gamma$. But now observe that, since $(\ec_{\vec X=\vec x},\A^\F_{\vec X= \vec x})\models^W K\O=\vec o$, also $(\ec_{\vec X=\vec x},\B^\F_{\vec X=\vec x})\models^W K\O=\vec o$; thus $(\ec_{\vec X=\vec x})^* = \ec_{\vec X=\vec x} = \ec^\B_{\vec X=\vec x}=(\ec^*)^\B_{\vec X=\vec x}$. Thus (*) is equivalent to $((\ec^*)^\B_{\vec X=\vec x},\B^\F_{\vec X=\vec x})\models^O \gamma$, and then to $(\ec^*,\B)\models^O [\vec X=\vec x]\gamma$. 
\end{proof}

\begin{theorem}[Completeness with observables]\label{theorem: completeness with observables}
Let $\Gamma\cup\{\phi\}$ be a set of \LAfull formulas. Then:
\[
\Gamma \models^O \phi  \iff \Gamma\vdash_{\LOfullo} \phi.  \]
\end{theorem}

\begin{proof}
By Proposition \ref{prop_reduction}, we can assume that $\Gamma$ and $\phi$ are in \LAkcres.
By the usual arguments, the statement is then equivalent to the assertion that every maximally \LOkco-consistent set of formulas $\Gamma$ is true in some pair $(\ec,\A)$, where $\ec$ is an epistemic causal model with observables. Now, if $\Gamma$ is \LOkco-consistent, then (Proposition \ref{prop: reduction to LOfull}) it is \LOkc-consistent; thus, by the completeness theorem for the case without observables (Theorem \ref{thm:rcm-LAka}) there is an $(\ec,\A)$ that satisfies $\Gamma$. Since $\Gamma$ is maximally \LOkco-consistent, in particular $\ax{OC} \in \Gamma$. Thus, by Proposition \ref{prop: OC models have observables}, there is an $\ec^*$ with observables such that $(\ec^*,\A)\models \Gamma$, as needed.
\end{proof}

\myparagraph{Completeness by explicit construction of the models} In the previous section, we gave an indirect proof of the completeness of the proof system \LOfullo by reducing this problem to the completeness result for \LOfull. We now present a direct proof, which goes through a variant of the canonical model construction. That is, we explicitly show how to build a model (with observables) for any consistent set of $\LAfull$ formulas. By the translation given in Proposition \ref{prop_reduction}, it suffices to do this for sets of \LAkcres formulas.

\myparagraphit{Building causal models}
Given an \LOkco-consistent $\Gamma\subseteq \LAkcres$, we can construct a causal model $\langle\Sig,\F^\Gamma,\A^\Gamma\rangle$ for it 
by using the same the definitions as given in Appendix \ref{app:rcm-LAk} but relying on the set $\mcssko:=\{\text{maximal \LOkco -consistent theories}\}$ instead of $\mcssk$. Observing that $\mcssko\subseteq \mcssk$, we can reuse all the results on causal models from Appendix \ref{app:rcm-LA}, most significantly Proposition \ref{pro:LAres:truth-lemma:atoms}, which may be reformulated as follows:

\begin{proposition}\label{pro:truth-lemma-atoms}
  Let $\Gamma \in \mathbb C^O$ be a maximally \LOkco-consistent set of \LAkcres-formulas. Let $\opint{\vec{X}}{\vec{x}}$ be an assignment, for $\vec{X}$ a tuple of variables in $\AV$; take $Y \in \AV$ and $y \in \Ran(Y)$. Then,
  \[
    \int{\opint{\vec{X}}{\vec{x}}}Y{=}y \in \Gamma
    \qquad\text{if and only if}\qquad
    \tuple{\Sig, \F^\Gamma, \A^\Gamma} \models \int{\opint{\vec{X}}{\vec{x}}}Y{=}y
  \]
\end{proposition}

\myparagraphit{Building the epistemic causal model with observables} We have seen how to associate a causal model $\tuple{\Sig,\F^\Gamma,\A^\Gamma}$ to each maximally \LOkco-consistent set of formulas $\Gamma$. Now we also want to associate an appropriate team to it. As a first step, let $\rmcss^\Gamma := \set{ \Gamma' \in \mcssko \mid \F^{\Gamma'} = \F^\Gamma}$. \\
 Let $R^\Gamma\subseteq \mcssko\times\mcssko$ be defined as $(\Gamma',\Gamma'')\in R^\Gamma$ if and only if
 \begin{enumerate}
 \item  for all   $\chi\in\LAkcres,(K\chi\in \Gamma' \Rightarrow \chi\in\Gamma'')$, and
 \item  there is an $\vec o\in\O \text{ s.t. } \O=\vec o \in \Gamma'\cap\Gamma''$.
 \end{enumerate}
Obviously, $R^\Gamma$ is a refinement of the canonical relation given in Appendix \ref{app:rcm-LAk}. We show that also  $R^\Gamma$ is an equivalence relation. 

  For transitivity: suppose $\Gamma_1 R^{\Gamma}\Gamma_2$ and $\Gamma_2 R^{\Gamma}\Gamma_3$, let $\O=\vec{o}\in\Gamma_1\cap\Gamma_2$, and let $\O=\vec{o'}\in\Gamma_2\cap\Gamma_3$. In order to verify $\Gamma_1 R^{\Gamma}\Gamma_3$, we need to check conditions (i) and (ii). $\Gamma_2$ is maximal consistent; thus, by axiom \ax{A_1}, for each $O\in \O$ there is exactly one value $o\in\R(O)$ such that $O=o\in \Gamma_2$. Therefore there is a (unique) tuple of value $\vec{o_2}$ for $\O$ such that $\O=\vec{o_2}\in \Gamma_2$. Since furthermore  $\O=\vec{o},\O=\vec{o'}\in\Gamma_2$, we conclude $\vec{o}=\vec{o}_2=\vec{o'}$. Now we have $\O=\vec{o_2}\in \Gamma_1\cap\Gamma_3$, so condition (ii) is satisfied. For (i), notice that, by axiom \ax{4}, 
  the assumption $K\chi\in\Gamma_3$ implies that $KK\chi\in\Gamma_3$; thus $K\chi\in\Gamma_2$ and $\chi\in\Gamma_1$. So, condition (i) is also satisfied.

 For reflexivity, we need to check conditions (i) and (ii) to obtain $\Gamma_1 R^{\Gamma}\Gamma_1$. since $\Gamma_1$ is maximal consistent, by the same reason as before, there is a 
 formula $\O=\vec{o}\in \Gamma_1\cap\Gamma_1$, so condition (ii) is satisfied. For condition (i), by axiom \ax{T}, 
 $K\chi\in\Gamma_1$ implies $\chi\in\Gamma_1$, so condition (i) is also satisfied.

 In order to verify $R^{\Gamma}$ is symmetric, suppose $\Gamma_1 R^{\Gamma} \Gamma_2$; we will show condition (i) and condition (ii) for $\Gamma_2 R^{\Gamma}\Gamma_1$. Condition (ii) is trivially satisfied as $\O=\vec{o}\in\Gamma_1\cap\Gamma_2$ iff $\O=\vec{o}\in\Gamma_2\cap\Gamma_1$. For condition (i), for any $\chi\in \changeFV{\LAkc}{\LAkcres}$, suppose $K\chi\in\Gamma_2$ while $\chi\not\in\Gamma_1$. Then $K\chi\not\in\Gamma_1$ (by axiom \ax{T}). Thus, $\neg K\chi\in\Gamma_1$ by maximality,  and by axiom 
 \ax{5}, 
 $K\neg K\chi\in\Gamma_1$. Then by $\Gamma_1 R^{\Gamma}\Gamma_2$, $\neg K\chi\in\Gamma_2$, contradiction. So condition (ii) is also satisfied.

 So $R^{\Gamma}$ is an equivalence relation.

Then, we can define the team associated to $\Gamma$ as  $\T^\Gamma:=\{\A^{\Gamma'} \mid (\Gamma,\Gamma')\in R^\Gamma\}$ (which is a smaller set than the one introduced in Appendix \ref{app:rcm-LAk}). We have:

  \begin{proposition}\label{pro:ecmo}
  Take $\Gamma \in \mcssko$. The tuple $\tuple{\Sig, \F^\Gamma, \T^\Gamma}$ is an epistemic causal model with observables, that is:

  (i) every valuation in $\T^\Gamma$ complies with $\F^\Gamma$.

  (ii) the value of $\O$ is constant in $\T^\Gamma$.
\end{proposition}

\begin{proof}
(i) is proved as before.

(ii): if $\B\in \T^\Gamma$, then $\B=\A^{\Gamma'}$ for some $\Gamma'$ such that $(\Gamma,\Gamma')\in R^\Gamma$. So there is an $\vec o\in\Ran(\O)$ such that $\O=\vec o \in \Gamma\cap\Gamma'$. Furthermore, $\F^\Gamma = \F^{\Gamma'}$. So by Proposition \ref{pro:truth-lemma-atoms} we have $\langle\Sig,\F^\Gamma,\A^\Gamma\rangle\models \O=\vec o$ and $\langle\Sig,\F^{\Gamma'},\A^{\Gamma'}\rangle\models \O=\vec o$. So $\B(\O)=\A^\Gamma(\O)$ for each $\B\in \T^\Gamma$.
\end{proof}

The tuple $\ec^\Gamma := \tuple{\Sig, \F^\Gamma, \T^\Gamma}$ will act as a sort of canonical model, although indexed by the set of sentences $\Gamma$. As proved below, it indeed satisfies the existence and truth lemmas.




\begin{lemma}[Existence Lemma w/ observables]\label{lemma: existence with observables}
Let $\Gamma\in\mcssko$. Then, for any $\Gamma' \in \rmcss^\Gamma$,
\begin{center}
$\neg K\neg \chi \in \Gamma'$ iff there is a $\Gamma'' \in\rmcss^\Gamma$ such that $\Gamma' R \, \Gamma''$ and $\chi\in\Gamma''$.
\end{center}
\end{lemma}

\begin{proof} By the same argument as in modal logic, we can show that $\{\chi\} \cup \{\psi \mid K\psi\in\Gamma'\}$ is a consistent set (see e.g the proof of Lemma 4.20 in \citealp{BlackburnRijkeVenema2001}). Let $\Gamma''$ be a $\LOfull-MCS$ extended from $\{\chi\} \cup \{\psi \mid K\psi\in\Gamma'\}$.

We show that $\LOfull$ guarantees that $\F^{\Gamma''}=\F^{\Gamma'} (=\F^\Gamma)$, so that $\Gamma''\in\rmcss^\Gamma$ as well. Suppose not; then $\F^{\Gamma''}\neq \F^{\Gamma'}$. 
Then there is a variable $Y\in\NV$ and an $\vec x \in \Ran(\vec X)$ (where $\vec{X}= (\XV\cup \NV)\setminus \{Y\}$) 
such that $\F^{\Gamma''}(\vec{x})\neq \F^{\Gamma'}(\vec{x})$. Let $\F^{\Gamma'}(\vec{x})=y'$ and $\F^{\Gamma''}(\vec{x})=y''$ where $y'$ and $y''$ are distinct values in $\Ran(Y)$. By the definition of $\F^{\Gamma'}$, $\F^{\Gamma'}(\vec{x})=y'$ iff $[\vec{X}=\vec{x}]Y=y'\in \Gamma'$. By axiom \ax{KL}, $K[\vec{X}=\vec{x}]Y=y'\in \Gamma'$. By the definition of $\Gamma''$, $[\vec{X}=\vec{x}]Y=y'\in \Gamma''$; but on the other hand, by the definition of $\F^{\Gamma'}$ we also have $[\vec{X}=\vec{x}]Y=y'\in \Gamma''$, and thus by axiom \ax{A_1} $\neg[\vec{X}=\vec{x}]Y=y''\in \Gamma''$; thus $\Gamma''$ is inconsistent, and we obtain a contradiction.

It remains to show that there is a tuple $ \vec o\in\O \text{ s.t. } \O=\vec o \in \Gamma'\cap\Gamma''$; take $\vec o$ to be $\A^{\Gamma'}(\O)$.
First, by the definition of $\A^{\Gamma'}(\O)$ we have $\O=\vec o \in \Gamma'$. Secondly, using
axiom \ax{OC} and classical logic, we can show that $\LOfullo\vdash \O=\vec o \rightarrow K(\O=\vec o)$.
Since $\O=\vec o \in \Gamma'$, then, since $\Gamma'$ is closed under MP we have $K(\O=\vec o)\in \Gamma'$. Thus, by definition of $\Gamma''$, $\O=\vec o \in \Gamma''$.
\end{proof}

\begin{lemma}[Truth Lemma with observables]\label{Truthlemma1}
Let $\Gamma\in\mathbb{C}^O$. For any $\phi \in \LAkcres$ and any $\Gamma'\in \T^{\Gamma}$, 
$(\ec^{\Gamma},\A^{\Gamma'})\models^O \phi$ iff $\phi\in\Gamma$.
\end{lemma}

\begin{proof}
First, notice that  if $\Gamma' \in\T^\Gamma$, then $\ec^\Gamma = \ec^{\Gamma'}$. Thus, it suffices to prove the simpler statement that $(\ec^{\Gamma},\A^{\Gamma})\models^O \phi$ iff $\phi\in\Gamma$. 

Secondly, we remark that it suffices to prove the statement for $\Gamma,\phi$ in \LAkcres. Indeed, by Proposition \ref{prop_reduction} we have that, for any $\phi\in \LAfull$, there is $\xi_\phi\in\LAkcres$ such that $\phi\leftrightarrow \xi_\phi$ is derivable in $\LOfullo$. So for any formula $\phi\in\LAfull$, by the soundness of $\LOfullo$ we have:\\
\[
(\ec^\Gamma,\A^\Gamma)\models^O \phi \iff
(\ec^\Gamma,\A^\Gamma)\models^O \xi_\phi;
\]
and on the other hand, since $\phi\leftrightarrow \xi_\phi$ is derivable in $\LOfullo$, and $\Gamma$ is $\LOfullo$ maximal consistent (thus closed under classical logic), we have:
\[
\xi_\phi \in \Gamma 
\iff  \phi\in\Gamma.
\]

The proof of the statement $(\ec^{\Gamma},\A^{\Gamma})\models^O \phi$ iff $\phi\in\Gamma$, for $\Gamma,\phi$ in \LAkcres, is by induction on $\phi$ and is identical to the proof of the Truth Lemma for the semantics without observables (Lemma \ref{lem:LAkres:truth-lemma}) with the exception of the case for $\phi = K\chi$. 

If $\phi$ is of the form $K\chi$, notice that
 \[
 \begin{array}{ll}
 (\ec^\Gamma,\A^\Gamma)\models^O K\chi &  \Leftrightarrow \text{ for all  } \Gamma' \text{ with } \Gamma R \Gamma', 
 (\ec^\Gamma,\A^{\Gamma'})\models^O \chi\\
 & \Leftrightarrow \text{ for all } \Gamma' \text{ with } \Gamma R\Gamma', \chi\in \Gamma \text{ (by IH)}\\
&  \Leftrightarrow \  K\chi\in \Gamma \text{ by the Existence Lemma \ref{lemma: existence with observables}}.
 \end{array}
 \]
\end{proof}



Thus, for any $\Gamma\in\mathbb{C}^O$, $(\ec^\Gamma,\A^\Gamma)$ is a model (with observables) for it; we then have another proof of the completeness Theorem \ref{theorem: completeness with observables}.

\end{appendix:full}

\begin{appendix:short}
\myparagraph{Soundness.} The soundness of most axioms and rules over epistemic causal models with observables is straightforward. We will only explicitly consider the axioms from Table \ref{tbl:rcm-LA} and \ax{PD}. The first are taken care of by the following lemma, which can be proved by a straightforward structural induction.

\begin{lemma}
  Let $\phi$ be an \LAfull formula without occurrences of $K$ or announcement operators. Let $\ec= \langle\Sig,\F,\T\rangle$ be epistemic causal model with observables; take $\A\in\T$. Then:
  \[ (\ec,\A)\models^{O}\phi  \qiffq \langle\A,\F\rangle \models\phi. \]
\end{lemma}

\noindent For axiom \ax{PD}, we have the following.

\begin{lemma}\label{lemma_xxoo}
  Let $\ec=\tuple{\Sig,\F,\T}$ be an epistemic causal model w/ observables and $\A\in\T$. Let $\vec o = \A^\F_{\vec X=\vec x}(\O)$; write $\delta$ for $[\vec X=\vec x]\O=\vec o$. Then, $\B\in\T^\delta$ implies
  \begin{center}
    $(\ec^\delta)^\B_{\vec X=\vec x}=(\ec^\delta)^\A_{\vec X=\vec x}= \ec^\A_{\vec X=\vec x} = \ec^\B_{\vec X=\vec x}$.
  \end{center}
  \begin{proof}
    From $\B\in\T^\delta$ it follows that $(\ec^\delta,\B)\models [\vec X=\vec x]\O=\vec o$; then we get $((\ec^\delta)^\B_{\vec X=\vec x},\B^\F_{\vec X=\vec x})\models \O=\vec o$, which implies $\B^\F_{\vec X=\vec x}(\O)=\vec o=\A^\F_{\vec X=\vec x}(\O)$. This yields the first equality; the third is proved analogously, using $\ec$ instead of $\ec^\delta$.

    For the second equality, observe that any element of $\T^{\F,\A}_{\vec X=\vec x}$ is of the form $\C^\F_{\vec X=\vec x}$ for some $\C\in \T$. However, by the definition of epistemic model with observables, $\C^\F_{\vec X=\vec x}(\O)=\vec o$; thus, $(\ec^\A_{\vec X=\vec x},\C^\F_{\vec X=\vec x}) \models \O=\vec o$. Therefore, $(\ec,\C) \models [\vec X=\vec x]\O=\vec o$, i.e., $\C\in \T^\delta$. Since $\C^\F_{\vec X=\vec x}(\O)=\vec o$, then, $\C^\F_{\vec X=\vec x}\in (\T^\delta)^{\F,\A}_{\vec X=\vec x}$.
  \end{proof}
\end{lemma}

\begin{theorem}
  Axiom \ax{PD} is sound on epistemic causal models w/ observables.
  \begin{proof}
    If case $\vec X$ is the empty set, axiom \ax{PD} reduces to:
    \[ K\gamma \leftrightarrow \bigvee_{\vec o\in \Ran(O)}( \O = \vec o \land [\O = \vec o!]K\gamma). \]
    Thus, from left to right, assume $(\ec,\A)\models K\gamma$. Let $\vec o=\A(\O)$. Then $(\ec,\A)\models \O = \vec o$. By definition of epistemic causal models with observables, $(\ec,\B)\models \O=\vec o$ for all $\B\in\T$. So $\T^{ \O=\vec o}=\T$ and therefore $(\ec^{ \O=\vec o},\A)\models K\gamma$. Thus, $(\ec,\A)\models [\O=\vec o!]K\gamma$. From right to left, assume $(\ec,\A)\models \bigvee_{\vec o\in \mathcal O}( \mathcal O = \vec o \land [\O = \vec o!]K\gamma)$. Then $(\ec,\A)\models \O = \vec o$ and $(\ec,\A)\models [\O=\vec o!]K\gamma$ for some $\vec o \in \Ran(\O)$. The latter implies $(\ec^{ \O=\vec o},\A)\models K\gamma$; the former implies $\ec=\ec^{\O=\vec o}$, as before. Thus, $(\ec,\A)\models K\gamma$.

    \smallskip

    Now, suppose $\vec X$ is non-empty; let $\ec=\tuple{\Sig,\T,\F}$. From left to right, assume $(\ec,\A)\models [\vec X = \vec x]K\gamma$. We have to show the right-hand side of the axiom for some $\vec o \in \Ran(\O)$; we will show it for $\vec o:= \A_{\vec X = \vec x}(\O)$. The assumption implies
    \[ (\star): (\ec^\A_{\vec X = \vec x},\A^\F_{\vec X = \vec x})\models K\gamma. \]
    so, for every $\B\in \T^{\F,\A}_{\vec X = \vec x}$, we have $(\ec^\A_{\vec X = \vec x},\B)\models \gamma$. Notice that $(\ec^\A_{\vec X = \vec x},\A^\F_{\vec X = \vec x})\models \O = \vec o$; thus, $(\ec,\A)\models [\vec X = \vec x]\O = \vec o$. This is the first conjunct we needed to prove. Now write $\delta$ for $[\vec X = \vec x]\O = \vec o$. Take $\C\in \T^\delta$; we want to verify that $(\ec^\delta,\C)\models [\vec X=\vec x]\gamma$. By Lemma \ref{lemma_xxoo}, we have $(\ec^\delta)^\C_{\vec X = \vec x} = \ec^\A_{\vec X = \vec x}$. Thus, $(\ec^\delta,\C)\models [\vec X=\vec x]\gamma$ iff $((\ec^\delta)^\C_{\vec X = \vec x},\C^\F_{\vec X = \vec x})\models \gamma$ iff $(\ec^\A_{\vec X = \vec x},\C^\F_{\vec X = \vec x})\models \gamma$, which is true by ($\star$). Since $(\ec^\delta,\C)\models [\vec X=\vec x]\gamma$ for each $\C\in \ec^\delta$, we have  $(\ec^\delta,\A)\models K[\vec X=\vec x]\gamma$. We then conclude $(\ec,\A)\models [\delta!]K[\vec X=\vec x]\gamma$, as needed.

    From right to left, assume there is $\vec o\in\O$ with $(\ec,\A)\models [\vec X = \vec x]\O = \vec o$ and $(\ec,\A)\models [\delta!]K[\vec X = \vec x]\gamma$. For any $\C\in\T_{\vec X=\vec x}^{\F,\A}$ we have $\C=\B^{\F}_{\vec{X}=\vec{x}}$ for some $\B\in\T$. Thus, $\B\in \T^\delta$ (as $\B^{\F}_{\vec{X}=\vec{x}}(\O)=\vec{o}$). So, by Lemma \ref{lemma_xxoo}, $(\ec^\delta)_{\vec X=\vec x}^\B= \ec_{\vec X=\vec x}^\A$. By assumption, $(\ec^{\delta},\A)\models K[\vec{X}=\vec{x}]\gamma$, so $((\ec^\delta)^\B_{\vec{X}=\vec{x}},\B^\F_{\vec X=\vec x})\models\gamma$. Therefore $(\ec_{\vec X=\vec x}^\A,\B^\F_{\vec X=\vec x})\models\gamma$ and then $(\ec_{\vec X=\vec x}^\A,\C)\models\gamma$. Since $\C$ is an arbitrary assignment in $\T_{\vec X=\vec x}^{\F,\A}$, we have $(\ec^\A_{\vec{X}=\vec{x}},\A^\F_{\vec{X}=\vec{x}})\models K\gamma$ and thus $(\ec,\A)\models [\vec{X}=\vec{x}]K\gamma$.
  \end{proof}
\end{theorem}

\myparagraph{Completeness via reduction to the case without observables} The first step consists in eliminating the announcement operators. This is done using axioms \ax{!_=}, \ax{!_\lnot}, \ax{!_\land} and \ax{!_K} and \ax{PD}, together with the following generalisation of the replacement rules proposed before:
\begin{center}
  $\ax{RE_{full}}:\;\; \text{ if } \vdash \chi_1 \leftrightarrow \chi_2 \text{ then } \vdash \phi \leftrightarrow \phi[\chi_2/\chi_1]$
\end{center}
where $\phi[\chi_2/\chi_1]$ obtained by replacing, in $\phi$, some occurrences of $\chi_1$ with $\chi_2$.

\begin{proposition}
  The rule \ax{RE_{full}} is admissible in \LOfullo.
  \begin{proof}[Sketch]
    The rules \ax{RE_K}, \ax{RE_=}, \ax{RE_!} and \ax{_!RE} are provable as in \LOfull. The main claim can then be proved by induction on $\phi$: the base case is trivial, the Boolean cases rely on using classical logic (\ax{P}+\ax{MP}), and the cases for $\phi = K\psi$ (resp. $[\vec X=\vec x]\psi$, $[\alpha!]\psi$) are covered by \ax{RE_K} (resp. \ax{RE_=}, \ax{_!RE}+\ax{RE_!}).
  \end{proof}
\end{proposition}

\begin{proposition}\label{prop_reduction}
  \begin{inlineenum} \item Every formula $\phi \in \LAfull$ is logically equivalent to a formula $\xi_\phi \in \LAfullres$. Moreover, $\phi \leftrightarrow \xi_\phi$ is derivable in \LOfullo. \item Every formula $\xi \in \LAfullres$ is logically equivalent to a formula $\chi_\xi \in \LAkcres$. Moreover, $\xi \leftrightarrow \chi_\xi$ is derivable in \LOfullo.\end{inlineenum}
  \begin{proof}
    For the first, modify the translation ${\trr}$ (Definition \ref{def:trr}) by taking
    \begin{center}
      $\displaystyle \trr([\vec X=\vec x]K\gamma):= \bigvee_{\vec o\in \mathcal O}\Big( [\vec X = \vec x] \mathcal O = \vec o \land \big[[\vec X = \vec x] \mathcal O = \vec o ! \big]K[\vec X = \vec x]{\trr}(\gamma)\Big)$.
    \end{center}
    The correctness of this clause is proved by axiom \ax{PD} and rule \ax{RE}. This clause allows removing instances of $K$ from the consequents of counterfactuals. For the second, use the same translation as in the case without observables (Definition \ref{def:trf}); see the proof of Proposition \ref{pro:LAkarestoLAkres}.
  \end{proof}
\end{proposition}

\noindent For the completeness of \LOfullo, by Proposition \ref{prop_reduction} it suffices to show that $\LOkco:= \LOfullo \setminus \{\ax{A_{[]}}, \ax{A_\lnot},\ax{A_\land},\ax{A_{[][]}}, \ax{!_=}, \ax{!_\lnot}, \ax{!_\land}, \ax{!_K}, \ax{PD}$ is complete for \LAkcres (in the semantics w/ observables). Since $\LOkco = \LOkc\cup \{\ax{OC}\}$, we have the following.

\begin{proposition}\label{prop: reduction to LOfull}
  Let $\Gamma\subseteq \LAkcres$. Then
  \begin{center}
    $\Gamma \text{ is \LOkco-consistent } \qiffq \Gamma \cup \{\ax{OC}\} \text{ is \LOkc-consistent.}$
  \end{center}
\end{proposition}

\noindent Thus, for finding a model for a \LOkco-consistent set of \LAkcres formulas $\Gamma$, use the completeness theorem for the case without observables, which provides a pointed model for $\Gamma\cup \{\ax{OC}\}$, i.e. an epistemic causal model $\ec$ together with an assignment $\A$ such that $(\ec,\A)\models^W \Gamma\cup \{\ax{OC}\}$. If we prove that $(\ec^*,\A)\models^O \Gamma\cup \{\ax{OC}\}$ (where $\ec^*$ differs from $\ec$ only in that its signature has a set of observables), we are done. But this is provided by the following result. We write $\ax{OC}_{\Sig^*}$ for the axiom scheme \ax{OC} specialized to the signature $\Sig^*$.

\begin{proposition}\label{prop: OC models have observables}
  Let $\Sig=\tuple{\XV,\NV,\Ran}$ be a signature and $\ec = \tuple{\Sig,\T,\F}$ epistemic causal model. Let $\O$ be a subset of $\XV\cup\NV$ and $\Sig^*=\tuple{\XV,\NV, \O, \Ran}$ be the corresponding signature with observables. Suppose that $(\ec,\A)\models^W \ax{OC}_{\Sig^*}$ for some $\A\in\T$. Then,
  \begin{enumerate}
    \item The tuple $\O$ takes constant value in $\T$; therefore $\ec^*= \tuple{\Sig,\T,\F}$ is an epistemic causal model with observables $\O$.
    \item For all $\B\in\T$ and all $\varphi\in$\LAfull: $(\ec,\B)\models^W \varphi \iff (\ec^*,\B)\models^O \varphi$
  \end{enumerate}
  \begin{proof}
    \begin{inlineenum} \item This follows from $(\ec,\A)\models \bigvee_{\vec o\in \mathcal O}K(\O=\vec o)$, the special instance of \ax{OC}$_{\Sig^*}$ for $\vec X=\emptyset$. \item Use induction on $\varphi$. For the only non-trivial case, $\varphi= [\vec X=\vec x]\gamma$, we have $(\ec,\B)\models^W [\vec X=\vec x]\gamma$ iff  $(\ec_{\vec X=\vec x},\B^\F_{\vec X=\vec x})\models^W \gamma$. But $(\ec,\A)\models^W [\vec X=\vec x]\bigvee_{\vec o\in \O}K\O=\vec o$, so $(\ec_{\vec X=\vec x},\A^\F_{\vec X= \vec x})\models^W K\O=\vec o$ for some value $\vec o\in\Ran(O)$. Then $(\ec_{\vec X=\vec x})^*$, the epistemic causal model with observables that differs from $\ec_{\vec X=\vec x}$ only in that it has signature $\Sig^*$ instead of signature $\Sig$, is an epistemic causal model with observables $\O$. Then we can use the inductive hypothesis to obtain (*): $((\ec_{\vec X=\vec x})^*,\B^\F_{\vec X=\vec x})\models^O \gamma$. But now observe that, since $(\ec_{\vec X=\vec x},\A^\F_{\vec X= \vec x})\models^W K\O=\vec o$, also $(\ec_{\vec X=\vec x},\B^\F_{\vec X=\vec x})\models^W K\O=\vec o$; thus $(\ec_{\vec X=\vec x})^* = \ec_{\vec X=\vec x} = \ec^\B_{\vec X=\vec x}=(\ec^*)^\B_{\vec X=\vec x}$. Hence, (*) is equivalent to $((\ec^*)^\B_{\vec X=\vec x},\B^\F_{\vec X=\vec x})\models^O \gamma$, and then to $(\ec^*,\B)\models^O [\vec X=\vec x]\gamma$.\end{inlineenum}
  \end{proof}
\end{proposition}

\begin{theorem}[Completeness with observables]\label{theorem: completeness with observables}
  Let $\Gamma\cup\{\phi\}$ be a set of \LAfull formulas. Then:
  \begin{center}
    $\Gamma \models^O \phi  \iff \Gamma\vdash_{\LOfullo} \phi$.
  \end{center}
  \begin{proof}
    By Proposition \ref{prop_reduction}, we can assume that $\Gamma$ and $\phi$ are in \LAkcres. By the usual arguments, the statement is then equivalent to the assertion that every maximally \LOkco-consistent set of formulas $\Gamma$ is true in some pair $(\ec,\A)$, where $\ec$ is an epistemic causal model with observables. Now, if $\Gamma$ is \LOkco-consistent, then (Proposition \ref{prop: reduction to LOfull}) it is \LOkc-consistent; thus, by completeness for the case without observables (Theorem \ref{thm:rcm-LAka}) there is an $(\ec,\A)$ that satisfies $\Gamma$. Since $\Gamma$ is maximally \LOkco-consistent, in particular $\ax{OC} \in \Gamma$. Thus, by Proposition \ref{prop: OC models have observables}, there is an $\ec^*$ with observables such that $(\ec^*,\A)\models \Gamma$, as needed.
  \end{proof}
\end{theorem}

\end{appendix:short}

\bibliographystyle{abbrvnat}
\bibliography{thesis}

\end{document}